\documentclass[11pt]{article}

\usepackage[final]{acl}

\usepackage{times}
\usepackage{latexsym}

\usepackage[T1]{fontenc}
\usepackage[utf8]{inputenc}

\usepackage{microtype}

\usepackage{inconsolata}

\usepackage{graphicx}
\usepackage{algorithm}
\usepackage{algorithmic}
\usepackage{multirow}
\usepackage{amsmath}
\usepackage{booktabs}
\usepackage[table]{xcolor}
\usepackage{arydshln}
\usepackage{subcaption}
\usepackage{enumitem}
\usepackage{adjustbox}
\usepackage{pdfpages}
\usepackage[normalem]{ulem}

\title{Evaluation and LLM-Guided Learning of ICD Coding Rationales}

\author{
  \textbf{Mingyang Li\textsuperscript{1}},
  \textbf{Viktor Schlegel\textsuperscript{1,2,3}},
  \textbf{Tingting Mu\textsuperscript{1}},
\\
  \textbf{Wuraola Oyewusi\textsuperscript{1}},
  \textbf{Kai Kang\textsuperscript{4}},
  \textbf{Goran Nenadic\textsuperscript{1}}
\vspace{0.5em}
\\
  \textsuperscript{1}University of Manchester, Department of Computer Science,
  \\
  \textsuperscript{2}Imperial Global Singapore, \textsuperscript{3}Imperial College London, Department of Bioengineering,
  \\
  \textsuperscript{4}Shanxi Medical University
\vspace{0.5em}
\\\small{\texttt{\{mingyang.li, Tingting.Mu, gnenadic\}@manchester.ac.uk, v.schlegel@imperial.ac.uk}}
\\ \small{\texttt{wuraola.oyewusi@postgrad.manchester.ac.uk, k.nikey0422@gmail.com}}
}

\begin{document}
\maketitle
\begin{abstract}
%ICD coding involves mapping unstructured text from Electronic Health Records (EHRs) to standardized code system International Classification of Diseases (ICD). 
%
%While recent advances in deep learning have significantly improved the accuracy and efficiency of ICD coding, the lack of explainability in these models remains a major limitation, undermining trust and transparency.
%
%Current explorations about explainability largely rely on attention-based rationales and qualitative assessments by physicians, yet lack systematic evaluation over diverse rationales using consistent criteria on high-quality rationale datasets for ICD coding task particularly, as well as dedicated approaches explicitly trained to generate plausible rationales.

ICD coding is the process of mapping unstructured text from Electronic Health Records (EHRs) to standardised codes defined by the International Classification of Diseases (ICD) system.
%Although recent advances in deep learning have markedly improved the accuracy and efficiency of automated ICD coding, the lack of explainability in these models remains a critical limitation, undermining trust and transparency.
In order to promote trust and transparency, existing explorations on the explainability of ICD coding models primarily rely on attention-based rationales and qualitative assessments conducted by physicians, yet lack a systematic evaluation across diverse types of rationales using consistent criteria and high-quality rationale-annotated datasets specifically designed for the ICD coding task. Moreover, dedicated methods explicitly trained to generate plausible rationales remain scarce.
%
%In this work, we present evaluations of the explainability of rationales in ICD coding, focusing on two fundamental dimensions: faithfulness, which quantifies the extent to which explanations reflect the model’s actual decision-making process, and plausibility, which assesses the degree of alignment between model-generated explanations and human expert judgment.
In this work, we present evaluations of the explainability of rationales in ICD coding, focusing on two fundamental dimensions: faithfulness and plausibility---in short how rationales influence model decisions and how convincing humans find them. %To enable the systematic evaluation of 
For plausibility, we construct a novel, multi-granular rationale-annotated ICD coding dataset, based on the MIMIC-IV database and the updated ICD-10 coding system. %This dataset features richer annotations with diverse levels of granularity. 
We conduct a comprehensive evaluation across three types of ICD coding rationales: entity-level mentions automatically constructed via entity linking, LLM-generated rationales, and rationales based on attention scores of ICD coding models.
Building upon the strong plausibility exhibited by LLM-generated rationales, we further leverage them as distant supervision signals to develop rationale learning methods. Additionally, by prompting the LLM with few-shot human-annotated examples from our dataset, we achieve notable improvements in the plausibility of rationale generation in both the teacher LLM and the student rationale learning models.
%
%Using this dataset, we compare three categories of explanation strategies: attention-based rationales, entity-level rationales, and generative rationales. We also demonstrate how few-shot prompting with our dataset improves rationale generation in large language models (LLMs), yielding outputs more aligned with expert annotations.
%
%We empirically find that LLM-generated rationales align most closely with those of human experts. Moreover, incorporating few-shot human-annotated examples not only further improves rationale generation but also enhances rationale-learning process.
%Our findings offer a unified evaluation method and new rationale learning approaches, contributing to the development of more trustworthy and interpretable clinical coding systems.
\end{abstract}
\section{Introduction}

%Clinical coding involves translating free-text descriptions in patients' Electronic Health Records (EHRs) into standardized codes, which plays a crucial role in facilitating billing, reimbursement, audit procedures, and decision support within healthcare systems \cite{blundell2023health}. %These notes, typically free-text descriptions in Electronic Health Records (EHRs), include patient medical history, symptom descriptions, lab test summaries, diagnostic rationale, and records of daily activities. Clinical codes are hierarchical alphanumeric identifiers that represent specific medical diagnoses, procedures, treatments, and healthcare services in a standardized format. In this work, we focus on ICD coding—the process of assigning codes from the International Classification of Diseases (ICD) system \cite{tzitzivacos2007international}, hierarchical alphanumeric identifiers that represent medical diagnoses, procedures. %This system includes: Diagnosis codes, which identify a patient's diseases, disorders, symptoms, and reasons for hospital visits; and procedural codes, which document surgical, medical, or diagnostic interventions.
Clinical coding is the process of translating free-text descriptions in patients' Electronic Health Records (EHRs) into standardized codes, serving a critical role in billing, reimbursement, auditing, and decision support within healthcare systems \cite{blundell2023health}. In this study, we focus on ICD coding---the document-level assignment of codes from the International Classification of Diseases (ICD) system \cite{tzitzivacos2007international}, which provides hierarchical alphanumeric identifiers representing medical diagnoses and procedures.

%Early ICD coding was performed manually by trained physicians or medical coders, a process that was often costly, labor-intensive, and prone to human error \cite{nguyen2018computer}. To address these challenges, researchers developed rule-based systems designed to replicate the decision-making process of human clinical coders \cite{pereira2006construction, crammer2007automatic}. Over time, machine learning methods emerged as a promising alternative, with Support Vector Machines (SVM) standing out as the state-of-the-art approach for a significant period \cite{lita2008large}.

%Following the rapid advancements in deep learning, the application of deep learning methods to clinical coding, such as Gated Recurrent Units (GRUs) \cite{catling2018towards} and Convolutional Neural Networks (CNNs) \cite{karimi2017automatic}, significantly improved the efficiency and accuracy of the coding process. Notably, CNNs achieved approximately 15\% higher precision compared to the best-performing traditional machine learning models. In recent years, even more advanced architectures, such as attention mechanisms \cite{shi2017towards, mullenbach2018explainable, falis2019ontological, vu2020label, feucht2021description, kim2021read, dong2021explainable, sun2021multitask, liu2021effective, van2022patient, yuan2022code} and transformer models \cite{chalkidis2020empirical, zhang2020bert, ji2021does, zhang2022automatic, huang2022plm, michalopoulos2022icdbigbird, yogarajan2022concatenating, yang2022knowledge}, have been adopted, consistently setting new state-of-the-art benchmarks in this field.

ICD coding relies on manual efforts by trained professionals, which is costly, labour-intensive, and error-prone \cite{nguyen2018computer}. To mitigate these challenges, rule-based systems were developed \cite{pereira2006construction, crammer2007automatic}, followed by machine learning approaches such as Support Vector Machines (SVM) \cite{lita2008large}. %, which became the state of the art for a time. 
With the rise of deep learning, approaches based on like Gated Recurrent Units (GRUs) \cite{catling2018towards} and Convolutional Neural Networks (CNNs) \cite{karimi2017automatic} substantially improved coding efficiency and accuracy. More recently, attention-based architectures \cite{liu2021effective, van2022patient, yuan2022code} such as transformers \cite{michalopoulos2022icdbigbird, yogarajan2022concatenating, yang2022knowledge} have been adopted, consistently achieving state-of-the-art results on clinical coding benchmarks.

%In addition to advancing state-of-the-art deep learning methods, researchers have explored enriching the representations of patients' notes and codes by incorporating external knowledge to enhance coding performance. These approaches include data augmentation \cite{wiegreffe2019clinical, sun2021multitask, falis2022horses, song2021generalized}, leveraging code descriptions or synonyms \cite{yuan2022code}, and utilizing the hierarchical structure or knowledge graph of ICD codes and knowledge graph of patients \cite{lu2020multi, cao2020hypercore, yuan2021graph, yang2022knowledge, michalopoulos2022icdbigbird}.

%Although machine learning and deep learning methods have demonstrated success in ICD coding, their lack of explainability - the nature of deep learning models - presents a significant challenge for understanding and interpreting model decisions. This limitation hampers trust and transparency, which are essential for enabling healthcare professionals and patients to comprehend and confidently rely on AI-driven outcomes \cite{amann2020explainability}. Consequently, researchers are increasingly focusing on developing methods that provide reliable explanations for clinical diagnoses based on electronic medical records (EMRs). They employ similar methods by extracting influential text snippets based on attention weights and evaluating the generated explanations through physician assessment.
Although these methods have achieved notable success in ICD coding, their inherent lack of explainability poses a major challenge for understanding and interpreting model decisions. This limitation may undermine trust and transparency, which are critical for enabling healthcare professionals and patients to rely on AI-driven recommendations~\cite{amann2020explainability}. %To address this issue, researchers are increasingly developing methods that provide reliable explanations, often by extracting short text snippets (\emph{rationales}) based on attention mechanisms, and sometimes validating these explanations through physician assessment. Regarding to the evaluation of these explanations, these works base physicians' rating, which are based on different evaluation metrics and criteria. Some works do the evaluation using the only existing rationale dataset MDACE, however, the codes are assigned independently during the annotation in MDACE, which suffering from label distribution discrepancy with ICD coding benchmark.
To address this issue, researchers have increasingly developed methods that provide reliable explanations, often by extracting short text snippets (\emph{rationales}) using attention mechanisms. For the evaluation of these explanations, some prior studies rely on physicians' assessments, which are based on various evaluation rubrics; while others use the only existing rationale-annotated resource, MDACE \citep{cheng2023mdace}, based on MIMIC-III and thus using the outdated ICD-9 system. Furthermore, MDACE was re-annotated with new codes, resulting in a significant label distribution shift from the original MIMIC-III labels, exacerbating the evaluation of training-based approaches.% coding benchmark.

%Despite the growing exploration of rationale generation in ICD coding, there remain two main gaps: \emph{(a)} a lack of systematic analysis of the explainability of these models using a high-quality rationale (evidence) dataset with a unified and aligned set of evaluation criteria which lead to \emph{(b)} a lack of approaches focussed on  rationale generation. 

%These issues reveal three main gaps in this task: (a) a lack of rationale (evidence) dataset designed for the ICD coding task specifically and constructed in accordance with the updating EHR database and coding system; (b) rationales extracted and evaluated in existing the works premirily reply on attention-based methods, yet a lack of diverse rationales evaluated and compared on consistent criteria; (c) a lack of exploration of rationale learning approaches and a lack of its supervision dataset.

These issues reveal three main gaps in this task: \emph{(a)} the absence of a rationale (evidence) dataset specifically designed for the ICD coding task and constructed based on modern resources; \emph{(b)} limited explorations of different types of rationales---as most existing studies focus exclusively on rationales derived from attention scores---and thus a lack of their consistent comparative evaluation; and \emph{(c)} the absence into \emph{rationale learning} approaches given  large-scale supervised datasets required for effective training do not exist.

%In this study, we address this gap by introducing a new rationale dataset and conducting a comprehensive analysis of the explainability of ICD coding models upon it, from two fundamental perspectives: faithfulness, which assesses the extent to which generated explanations accurately reflect the underlying decision-making processes of the models, and plausibility, which evaluates how convincing and interpretable these explanations are to human users. Additionally, we explore rationales learning approaches with the use of our rationale dataset. Our contributions include:
To address this gap, we comprehensively evaluate the quality of rationales in ICD coding from the angles of faithfulness and plausibility---in short, how rationales affect model classificatio ndecisions and how plausible human annotators find them---following~\citet{edin2024unsupervised}. Specifically, we systematically evaluate three types of rationales, which is enabled by a new rationale dataset we introduce in this paper. Finally, we explore two rationale learning approaches and examine the benefits of leveraging few-shot examples from our dataset for rationale generation and rationale learning. Our key contributions\footnote{The annotated rationale dataset, the code used in this study, and the Gemini-generated rationale dataset covering 122K MIMIC-IV ICD-10 documents are publicly available at: https://github.com/mingyangligithub/ICD-Coding-Explainability-Evaluation.} are summarised below:

\begin{itemize}[leftmargin=1em]
    %\item We systematically evaluate the explainability quality rationales produced by state-of-the-art ICD coding models equipped with label-wise attention mechanisms.
    %\item For faithfulness, we investigate models equipped with label-wise attention mechanisms to assess the extent to which their generated explanations accurately reflect the models’ true decision-making processes by perturbing or removing them.
    \item We construct a new rationale dataset for ICD coding task specifically based on the up-to-date MIMIC-IV benchmark with ICD-10 coding system for plausibility evaluation. This dataset provides richer rationales across multiple levels of granularity.
    %\item We conduct empirical comparisons of plausibility across three types of explanation strategies: (1) extracted rationales - unsupervised and (semi-)supervised approaches, (2) entity-level rationales, and (3) generated rationales - could and local LLMs.
    \item We conduct a comprehensive comparison of rationale plausibility across three types: (1) naive entity-level rationales derived from an entity linking dataset; (2) strong LLM-generated rationales, generated by both cloud-based and locally deployed LLMs; and (3) model-generated rationales, including both unsupervised and supervised approaches.
    \item We investigate two rationale learning approaches—multi-objective learning and a named entity recognition (NER) formulation—supervised by LLM-generated weak rationale labels.
    \item We demonstrate that leveraging few-shot human-annotated examples from our rationale dataset further improves both rationale generation and rationale learning.

\end{itemize}
%We show empirically that LLM-guided learning generates rationales more closely aligned with expert annotations, and the incorporation of human-annotated examples further improves rationale quality.
%
%We also observe a trade-off between the ICD coding classification performance and the rationale alignment with human annotations, informing future research that balancing both remains a challenging task.
\section{Related Work}

\paragraph{Rationale Snippets.} In one of the seminal studies on model explainability in clinical coding, \citet{mullenbach2018explainable} focus on extracting the most influential text snippets associated with predicted labels based on the importance values (attention weights) to $n$-grams in the discharge
summaries. %They also extract rationales with the highest cosine similarity to code descriptions and take them as an additional explanation. 
\citet{lovelace2020dynamically} follow the same idea, but apply attention mechanisms over multiple convolutional filters of different lengths, which allow them to consider variable spans of text. %\citet{yuan2021graph} also highlight the words based on the attention weights. 
\citet{dong2021explainable} build a Hierarchical Label-wise Attention Network (HLAN), which has label-wise word-level and sentence-level attention mechanisms, providing more comprehensive explanations for each label by highlighting key words and sentences. \citet{wang2022novel} visualise the attention distribution to provide explanations. Similarly, \citet{gao2024optimising} design heatmap visualisation to help coders better understand the inference logic from the notes. 

%\paragraph{Explainable Graph} \citet{chen2020towards} brings forward three variants of Bayesian
%Networks for disease inference that provide explanations. Each prediction explanation is generated with a template which contain the information of diagnosed disease and its connected symptom, vital sign and lab test in three networks, respectively. \citet{yuan2021graph} builds a patient-level explainable medical graph which consists of disease hierarchy and causal graph. \citet{wu2021counterfactual} models a hierarchical graph network that decomposes the EMR into sentences, clauses, and entities. The top upporting facts are evaluated by human annotators.

\paragraph{Evaluation Methods and Rationale Learning.} \citet{mullenbach2018explainable} evaluate the models' effectiveness in identifying highly informative rationales by a physician's assessment. \citet{kim2022can} assess explainability using human-grounded evaluation, where annotators rate each explanation rationale for a predicted code as highly informative, informative, or irrelevant. \citet{van2022patient} evaluate the explainability by faithfulness and conduct a manual analysis to judge whether highlighted tokens and prototypical patients aided decision‑making. %However, they only focus on three disease codes and compare the proposed model to post-hoc explanation methods, rather than other ICD coding baselines. 
%. 
%This work evaluates rationales using exact and position-independent token/span matching and further proposes a multi-objective supervision learning framework to jointly generate labels and rationales. 
%However, 
These studies primarily focus on evaluating the attention-based rationales. \citet{edin2024unsupervised} additionally assess the explainability of gradient-based and perturbation-based rationales. Furhtermore, both \citet{cheng2023mdace} and \citet{edin2024unsupervised} investigate whether rationales can be learned jointly with the main task of ICD coding (multi-objective), %supervised by the annotations in MDACE,
but their approaches are constrained by the small size of the dataset and the limited set of labels that it covers. 

In this work, we examine whether other types of rationales, specifically those produced by entity linking and large language models (LLM), can serve as effective explanations. We further propose a new rationale learning approach based on an NER formulation, which surpasses the multi-objective learning approach in generating plausible rationales. Moreover, we examine the effectiveness of LLM-generated labels in addressing the limitations of scarce and biased supervised data.
%However, the MDACE dataset suffers from code distribution discrepancies, mapping inaccuracy, sparse evidence annotations and outdated references.
%

\paragraph{Data Limitation.}
To facilitate automated evaluation of rationales \citet{cheng2023mdace} introduced MDACE, the only existing rationale-annotated ICD coding dataset other than the one we present in this paper.
%Being the only publicly available code rationale dataset, 
However, MDACE exhibits several limitations that limit its suitability to be the de-facto standard of rationale evaluation of ICD coding models:
%
% \begin{itemize}
    %\item 
    
    \textit{Distribution Shift:} The MDACE annotations are conducted independently of existing MIMIC-III labels, by coding each chart from scratch\footnote{coders could consult the original ICD-9 codes for reference}. Furthermore, coders used ICD-10 in line with their experience, which were subsequently mapped to ICD-9. This results in a significant code distribution shift from the standard MIMIC-III training set. Specifically, overlap for both Top-50 and all codesis only 37.00\% and 14.59\% on average per document, respectively. Notably, 40 new ICD-9 are introduced in MDACE, while 725 out of 1,281 codes in the Full setting are completely omitted. As a consequence,  the performance of PLM-ICD~\cite{huang2022plm} drops from 78.02\% Precision@8 on the official MIMIC-III to only 55.34\% on MDACE (Appendix D for more details); thus, MDACE significantly underestimates the performance of trained models.
    %The coding performance evaluated on the MDACE and the filtered MIMIC-III ICD-9 test set (sharing the same HADM IDs as MDACE) achieves 55.34\% and 78.02\% on the Precision@8 metric, respectively. This substantial inconsistency poses challenges in accurately evaluating the explainability of ICD coding models. Detailed statistics and analysis are provided in Appendix D.
%
%\textit{Inaccuracies Arising from Mapping:}
%The mapping between ICD-9 and ICD-10 codes is not one-to-one, resulting in inaccuracies during conversion. In instances where multiple ICD-9 codes correspond to a single ICD-10 code, and none of the mapped codes appear in the original MIMIC-III ICD-9 dataset, the selection is guided by the textual similarity between ICD-9 and ICD-10 code descriptions.
%
%\item 

\textit{Sparse Rationale Annotations:}
%Upon examining the quality of the annotations, we observe that most ICD codes are associated with very few supporting rationale annotations—often only a single instance. 
Most ICD codes have very sparse rationale annotations, often limited to a single instance. Such limited coverage may omit valid generated rationales, which identify other statements that support the classification, such as repeated mentions of the same finding or drug that supports a diagnosis.  %insufficient to represent the full spectrum of clinical rationale required for meaningful coding explanations.
%
%\item 

\textit{Outdated Relevance of MIMIC-III:}
%Moreover, MIMIC-III has become increasingly outdated in the context of ICD coding. Contemporary research predominantly adopts the MIMIC-IV dataset, which employs the ICD-10 coding system, offering improved alignment with current clinical documentation standards and coding practices.
With the obsoletion of ICD-9, the relevance of MIMIC-III vanes for ICD coding research---recent work uses the more recent MIMIC-IV dataset with ICD-10 annotation.
%\end{itemize}

Taken together, these observations mandate a strong need to develop a modern rationale annotation resource specifically for ICD coding. We heed the call in this paper by introducing  a rationale-annotated dataset, based on MIMIC-IV and its ICD-10 code annotations. We demonstrate further utility of this resource by using instances as few-shot examples to prompt LLMs, which benefits both rationale generation and rationale learning.

%However, the MDACE dataset suffers from 1) \textbf{code distribution discrepancies} compared to the standard MIMIC‑III ICD‑9 dataset, caused by the mapping between ICD‑10 and ICD‑9 coding systems, with code overlaps of only 37.00\% and 14.59\% for the Full and Top‑50 settings, respectively, and the omission of many frequently used codes; 2) \textbf{inaccurate} arising from the non‑one‑to‑one correspondence between ICD‑9 and ICD‑10 codes; 3) \textbf{sparse evidence annotations}, often limited to a single instance per code;  and 4) \textbf{reliance on the outdated MIMIC‑III} dataset reduces the relevance of these annotations, as contemporary research increasingly favors MIMIC‑IV with its ICD‑10 coding and closer alignment to current clinical practices.

%In this work, we present a framework for evaluating the explainability of ICD coding models from the perspectives of faithfulness and plausibility. We also develop a high-quality evidence annotation dataset with accurate code distributions, built on the up-to-date MIMIC-IV dataset with ICD-10 codes.
%\vspace{-3mm}
\begin{figure}[t]
\centering
\includegraphics[width=0.48\textwidth]{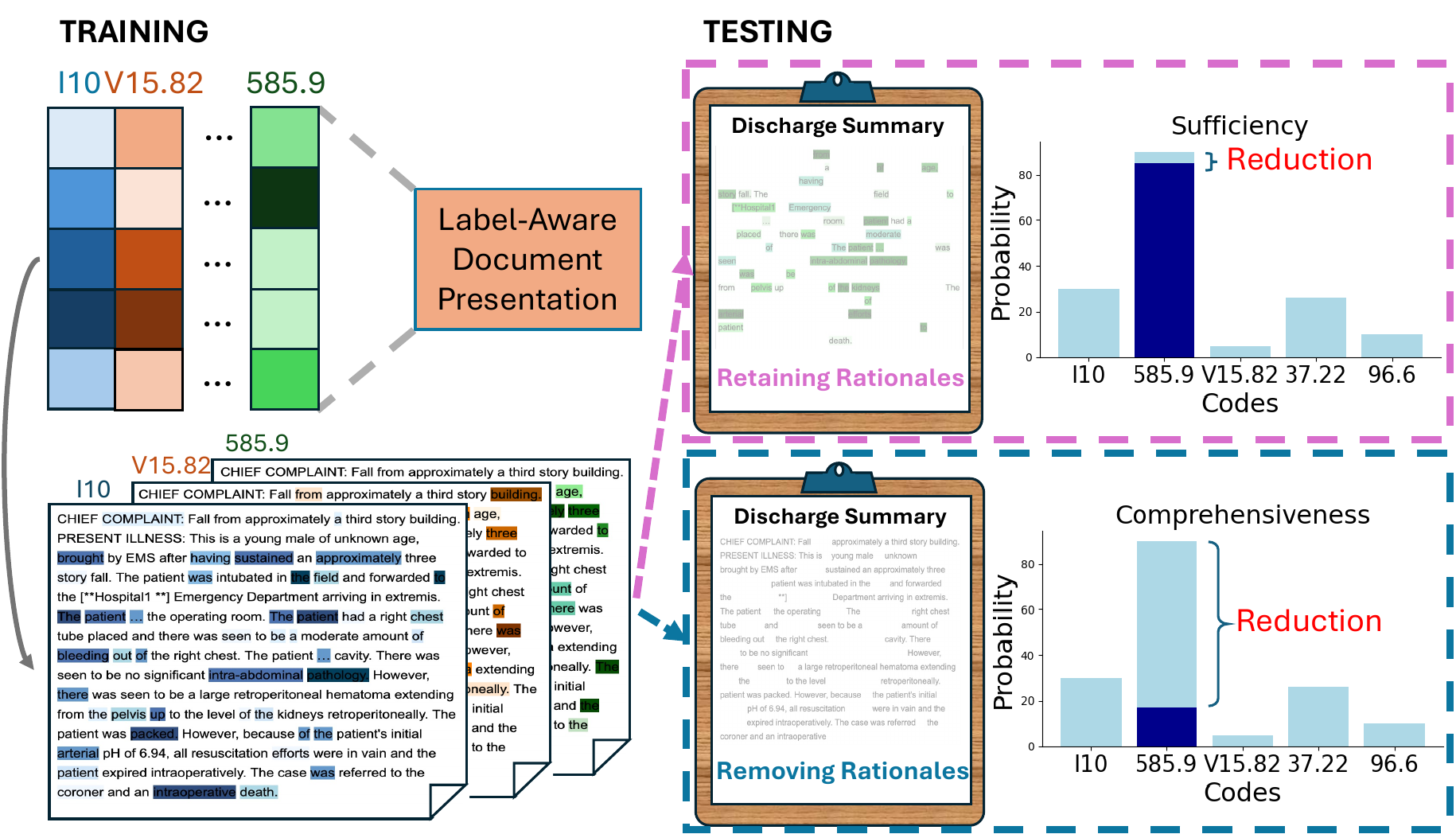}
    \caption{Faithfulness testing workflow. Sufficiency and comprehensiveness are evaluated by retaining or removing rationales from the original documents and using the modified texts as inputs to the trained ICD coding models.}
\label{fig:faithfulness}
\end{figure}
%\vspace{-8mm}
\section{Preliminary on Explainable ICD Coding}
ICD coding is treated as a multi-label classification task. 
It assigns  ICD codes based on a patient's clinical record,  describing their diseases or procedures. 
We consider clinical documents, each of which is a discharge summary denoted as $\mathbf{x}_i = \{\mathbf{t}_{i,1},\mathbf{t}_{i,2},...,\mathbf{t}_{i,N_t}\}$, consisting of $N_t$ tokens. 
The ICD coding model computes a label distribution over $N_l$ labels for the input $\mathbf{x}_i$, i.e., $f(\mathbf{x}_i) = \mathbf{p}_i = [p_{i,1},p_{i,2},\ldots,p_{i,N_l}]$. 
The final   codes are selected by thresholding the predicted probabilities with $0<\tau<1$, i.e., 
$\hat{y}_{i,l} =
\begin{cases}
1, & \text{if } p_{i,l} > \tau\\
0, & \text{otherwise}
\end{cases}$,  for $l=1,2,\ldots, N_l$.

To enable explainable ICD coding, the prediction models are expected to  provide rationale explaining the decision making.
The family of attention-driven ICD coding models  achieves this by flagging key text rationales influential to the prediction using attention weights.
We select three attention-driven  ICD coding models: CAML, LAAT and PLM-ICD.
They compute an attention weight $\tilde{a}_{i,j,l}$ for each token $\mathbf{t}_{i,j}$ and for each label $l$.
Specifically, CAML   \cite{mullenbach2018explainable} employs a single-filter CNN to encode the input text and computes the attention weight    by $\tilde{a}_{i,j,l} = \text{softmax}({\mathbf{u}_l^T\mathbf{t}_j})$.
LAAT  \cite{vu2020label} shares the same underlying framework as CAML, but leverages a BiLSTM to represent the input text $\mathbf{x}_i$ and computes the attention weight by $\tilde{a}_{i,j,l} = \text{softmax}\left({\mathbf{u}_l^T\text{tanh}(\mathbf{W}_j \mathbf{t}_j})\right)$, introducing an additional weight matrix $\mathbf{W}_j$.
PLM-ICD  \cite{huang2022plm} utilizes a transformer pre-trained on biomedical and clinical texts to encode the input text, and employs the same attention layer as LAAT. 
The attention weights are then used to compute a  label-aware document representation by $\tilde{\mathbf{h}}_{i,l} = \sum_{j=1}^{N_t} \tilde{a}_{i,j,l} \,\mathbf{t}_j$.
Together with the label representation vector $\mathbf{z}_l$, $\tilde{\mathbf{h}}_{i,l}$ is used to estimate the label probability by $p_{i,l} = \sigma\!\bigl(\mathbf{z}_l^T \tilde{\mathbf{h}}_{i,l}\bigr)$ through the sigmoid function $\sigma$.
These models are trained by minimizing a binary cross‐entropy loss:

{\small
\begin{equation}
\mathcal{L}_{\text{coding}} = -\frac{1}{N_D} \sum_{i=1}^{N_D} \sum_{l=1}^{N_l} 
\left[ y_{i,l} \log \hat{y}_{i,l}   + (1 - y_{i,l}) \log (1 - \hat{y}_{i,l}) \right],
\end{equation}
}
where $N_D$ denotes the document number.
Rationales are then extracted by  selecting  \textit{top $p\%$ tokens} or \textit{top $N$ tokens} with the highest attention weights. 
% - to ensure a fairer comparison. This choice is motivated by our observation that the distribution of attention weights varies significantly across different methods. A fixed threshold does not ensure a consistent level of importance across models—it may result in selecting most or all tokens in some cases, while selecting very few in others.
%\textbf{\textit{Top $p \%$ tokens}}: We retain the top $p\%$ of tokens, which maintains a consistent token proportion across documents.
%Rather than selecting rationales based on fixed thresholds, we adopt two alternative
\begin{table}[t]
\centering
\captionof{table}{Statistics of RD-IV-10 and MDACE. The statistics are computed based on annotations from documents that appear in the ICD coding datasets. A / B: A represents the statistics of the rationale datasets, and B represents the statistics of MIMIC-IV ICD-10 (compared with RD-IV-10) and MIMIC-III ICD-10 (compared with MDACE).}
%\vspace{-2mm}
\resizebox{\columnwidth}{!}{%
\begin{tabular}{l c c}
    \hline
    \textbf{Statistics} & \textbf{RD-IV-10} & \textbf{MDACE} \\
    \hline
    No. documents & 150 & 354 \\
    Tokens / doc & 1690.63 & 1837.27 \\
    Codes / doc & 14.82 / 16.15 & 11.57 / 17.54 \\
    Code overlap (Full / Top-50) & 93.15\% / 83.88\% & 37.00\% / 14.59\% \\
    No. codes & 2223 / 2422 & 4096 / 6208 \\
    No. distinct codes & 989 / 1044 & 1195 / 1381 \\
    No. annotations & 5391 & 4992 \\
    Annotations / doc & 35.94 & 14.10 \\
    Tokens / annotation & 5.44 & 2.13 \\
    \hline
\end{tabular}
}
%\vspace{-7mm}
\end{table}
\begin{figure}[t]
\centering
%\vspace{-3mm}
\includegraphics[width=0.5\textwidth]{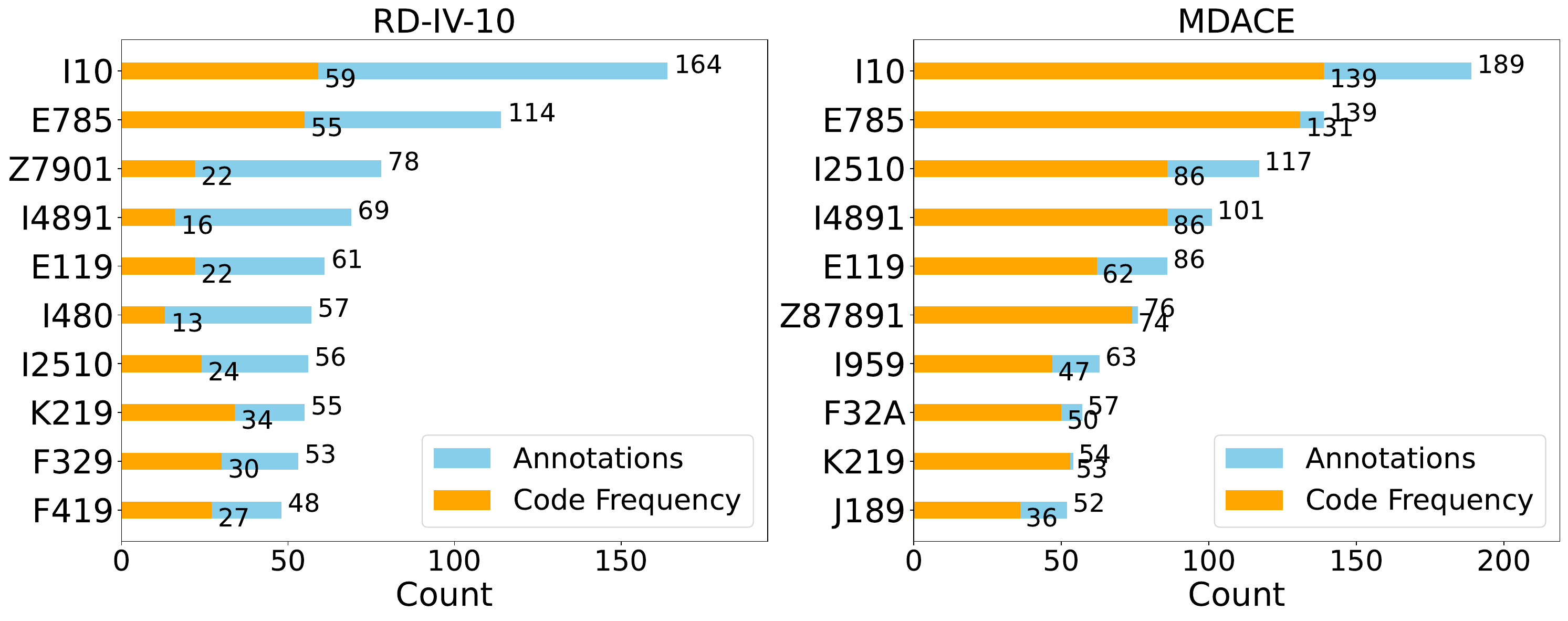}
\caption{Statistics of code and annotation frequencies for the top 10 codes in RD-IV-10 and MDACE. RD-IV-10 provides richer annotations for each label than MDACE.}
\label{top10_statistics}
\end{figure}
%\vspace{-12mm}
\section{An Empirical Analysis of ICD Rationales }
\subsection{Evaluation Metrics}
Explainability is typically evaluated from two complementary perspectives \cite{mendez2024outputs}: a) Model-centric  faithfulness, assessing how accurate  the extracted rationales reflect the internal reasoning of the prediction models. b) Human-centric   plausibility, measuring how convincing the rationales appear to people.  

\paragraph{Faithfulness.} %An explanation is considered faithful if modifying or removing the highlighted rationale significantly changes the model’s prediction.

We denote the extracted rationale for explaining why input $\mathbf{x}_i$ is predicted to label $l$ by $\mathbf{\hat{r}}_{i,l}$.
Its faithfulness can be assessed by two metrics, including \textbf{sufficiency}  and \textbf{comprehensiveness}   \cite{deyoung2019eraser},  quantifying the effect of only retaining or removing rationales, respectively. 
A rationale $\mathbf{\hat{r}}_{i,l}$ is considered sufficient if it enables a prediction that closely approximates the one produced by using the full input $\mathbf{x}_i $. 
This motivates the sufficiency metric of  $\text{Suff} = P(f(\mathbf{x}_i)) - P(f(\mathbf{\hat{r}}_{i,l}))$, where $P(\cdot)$ denotes a used performance measure for ICD coding, e.g., classification accuracy, precision  and recall, etc.
Conversely, a highly comprehensive explanation should significantly impact the model prediction when the rationale $\mathbf{\hat{r}}_{i,l}$ is removed. 
This motivates the comprehensiveness metric  that measures the prediction change   when the rationale $\mathbf{r}_{i,l}$ is excluded from the input $\mathbf{x}_i$, computed as  $\text{Comp} = P(f(\mathbf{x}_i)) - P(f(\mathbf{x}_i \backslash \mathbf{\hat{r}}_{i,l}))$.
 In general, lower  sufficiency  and  higher  comprehensiveness  indicate better faithfulness.

%For a given input $\mathbf{x}_i$ and a predicting label $l$,  %its rationale $\mathbf{\hat{r}}_{i,l}$ is a set of tokens that significantly contribute to the input representation in prediction by receiving higher attention scores. 
%we represent it as a binary mask $\mathbf{\hat{M}}_{i,l}$ over the input that indicates whether each token belongs to the rationale set or not (1 to include, 0 to exclude). %$\mathbf{\hat{r}}_i$ is a set of tokens $\{\hat{r}_1, \hat{r}_2, \dots, \hat{r}_{N_{\hat{r}}} \}$ that significantly contribute to the input representation in prediction by receiving higher attention scores. Here, $N_r$ denotes the number of rationale tokens.

%A rationale $\mathbf{r}_i$ is sufficient if it contains enough information to allow the model to make a prediction close to what it would make with full information $\mathbf{x}_i$. Specifically, sufficiency (suff) measures the change in $f(\mathbf{\cdot})$ when only $\mathbf{r}_i$ is kept in the input:
\paragraph{Plausibility.}

A rationale is considered plausible when it highlights text rationales that are perceived by humans (e.g., domain experts) as relevant and appropriate to support   model prediction.
We assess plausibility by comparing the three types of rationales with human-annotated rationales, using the same matching metrics as in MDACE—exact / position-independent (PI) token matches (TM) and span matches (SM)—as in MDACE.
To facilitate this, we construct an rationale dataset as below.
\subsection{Rationale Dataset Construction}

We introduce a new \textbf{r}ationale \textbf{d}ataset derived from MIMIC-\textbf{IV} and aligned with the ICD-\textbf{10} coding system - \textbf{RD-IV-10}, annotated by medical professionals. It includes detailed annotations capturing richer rationales supporting each code assignment, such as direct and indirect mentions, medications, and other pertinent clinical factors.
Details of dataset construction, including Data Selection, Annotator, Annotation Guidelines, Annotation Platform, Inter-Annotator Agreement and Details of Data Processing are provided in Appendix C.

%As shown in Table 1, the annotations in RD-IV-10 cover a significantly larger proportion of the original ICD labels compared to MDACE—an average of 14.82 out of 16.15 labels per document, versus 11.57 out of 17.54 in MDACE.
As shown in Table 1, the annotations in RD-IV-10 much closely match the original distribution of the ICD coding dataset compared to MDACE—93.15\% and 83.88\% in the Full and Top-50 settings, respectively, versus 37.00\% and 14.59\% in MDACE. We also report the specific labels for which no supporting rationale could be identified in Appendix C, offering further insight into the annotation quality of the MIMIC-IV ICD-10 dataset. Furthermore, our dataset provides richer rationale: it includes an average of 35.94 rationale spans for 14.82 labels per document, whereas MDACE contains only 14.10 spans for 11.57 labels. Figure~\ref{top10_statistics} further presents code‑level statistics, showing that the number of annotations far exceeds the code frequency in the RD‑IV‑10 dataset. This confirms our earlier observation that MDACE typically offers only a single piece of supporting rationale per code. In addition, our dataset features more comprehensive annotation formats across multiple levels of granularity, including words, phrases, and both complete and partial sentences. The average length of each rationale span is also greater—5.44 tokens compared to 2.13 tokens in MDACE. A detailed case study comparing the annotation quality of the two datasets is presented in Appendix E.
%Additionally, some codes consistently have the highest number of annotations across both datasets, including I10, E785, E119, I2510, and others.
\begin{figure*}[t]
\centering
\includegraphics[width=1\textwidth]{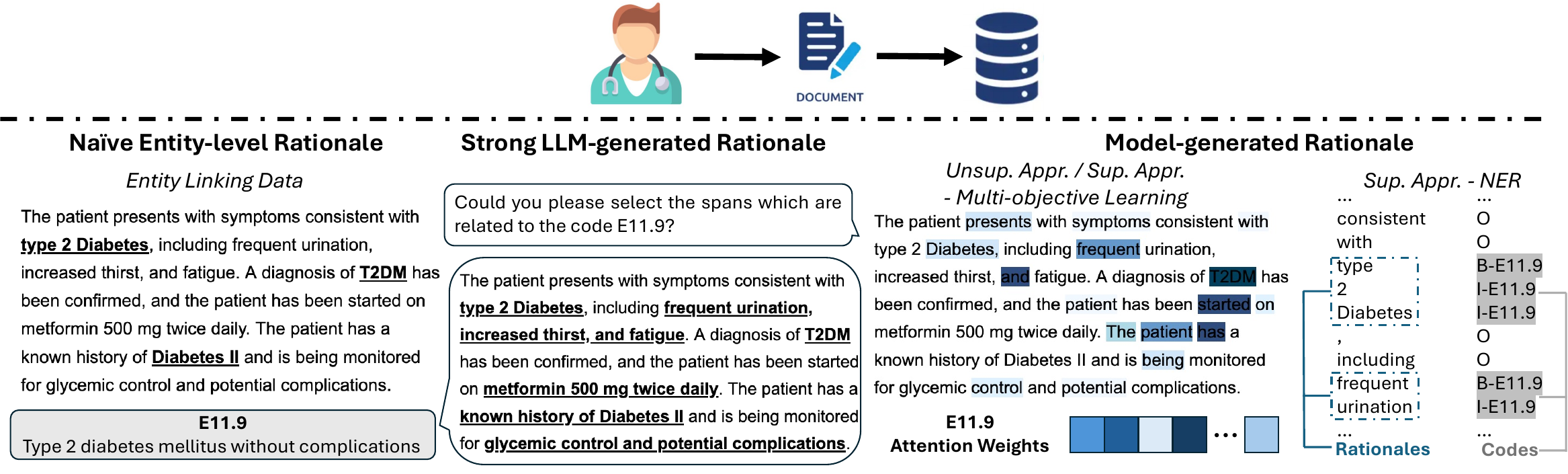} % Reduce the figure size so that it is slightly narrower than the column.
\caption{Examples of three types of rationales evaluated for plausibility. \textit{Unsup. / Sup.} denote Unsupervised and Supervised, separately. \textit{Appr.} indicates Approach.}
\label{plausibility}
\end{figure*}
%\vspace{-8mm}
\subsection{Explainability Evaluation}
We focus on examining the faithfulness of rationales extracted based on attention weights for CAML, LAAT and PLM--ICD. These are \textcolor{green!30!black}{\textbf{model-generated rationales}} produced with only supervision from ICD coding labels.
The overall architecture for  testing faithfulness is illustrated in Figure~\ref{fig:faithfulness}.
%Mingyang: Add one or two sentences to explain the architecture. A bit abrupt here.
%
In evaluating plausibility, we also analyze two additional types of rationales.
One is \textcolor{orange}{\textbf{naive entity-level rationales}} derived from an existing entity linking dataset -- SNOMED CT Entity Linking Challenge dataset \cite{hardman2023snomed}.
This dataset links ICD codes to direct named-entities appeared in text, e.g.,  entities `\textit{type 2 diabetes}', `\textit{T2DM}', and `\textit{Diabetes II}' are  linked to ICD-10 code `\textit{E11.9 -- Type 2 diabetes mellitus without complications}'. 
%When such entities appear in the input text $\mathbf{x}_i$, they can directly serve as the rationale $\mathbf{\hat{r}}_{i,l}$ of the associated code $l$.
%
The other is \textcolor{blue!60!black}{\textbf{strong LLM-generated rationales}}.
We prompt LLMs to extract rationale spans from patient notes that support specific ICD code assignments, and have observed that 2-Flash performs the best.
Given that LLMs occasionally fail to produce spans that exactly match the original text, we design algorithms to align the generated spans back to the original document.
Details of our prompt design and span alignment algorithm are provided in Appendix B.
%A thorough analysis of the comparison result is reported in the experiment section.

%Encouraged by the promising plausibility of LLM-generated rationales, we propose to incorporate supervision from rationale labels generated by best-performing Gemini 2-Flash, and explain the methodology in the next section. Figure~\ref{plausibility} presents an example of each type of rationale.
Motivated by the strong plausibility of LLM-generated rationales, we incorporate supervision from rationale labels produced by the best-performing Gemini 2-Flash model, with the methodology detailed in the following section. Figure~\ref{plausibility} illustrates representative examples of each rationale type.

%
%Mingyang: You dont have to say this. When you propose a new method, of course you evaluate it...
%and evaluate another type of \textbf{model-generated rationales} under this supervision. 
\section{LLM-Guided Rationale Learning}

LLMs have demonstrated strong performance across a variety of tasks in the clinical domain.
Also, we have observed from the previous rationale analysis that the LLM-generated rationales align quite well with  human-annotated rationales. 
This motivates us to take advantage of LLMs, i.e., being able to quickly identify rationale with reasonable quality, and  design LLM-guided rationale learning approaches.
We propose  to use  rationales produced by prompting LLMs as distant supervision signal, aiming at maximizing simultaneously classification accuracy for ICD coding and  plausibility of the model-generated rationales.
%
%The first is generated based on entity linking that is the task of mapping textual mentions within a document to standardized medical concepts. As illustrated in Figure 2, expressions such as `\textit{type 2 diabetes}', `\textit{T2DM}', and `\textit{Diabetes II}' are all linked to the ICD-10 code `\textit{E11.9 – Type 2 diabetes mellitus without complications}'. These direct mentions can serve as evidence for the associated codes. To investigate the plausibility of such entity-level rationales, we incorporate a baseline derived from the SNOMED CT Entity Linking Challenge dataset \cite{hardman2023snomed}.
%
\begin{figure*}[htbp]
    \centering
    
    % -------- Row 1 Title --------
    Sufficiency of Models on Four MIMIC Datasets (Precision@5/8) $\downarrow$\\[0.5em]
    
    \begin{subfigure}{0.24\textwidth}
        \includegraphics[width=\linewidth]{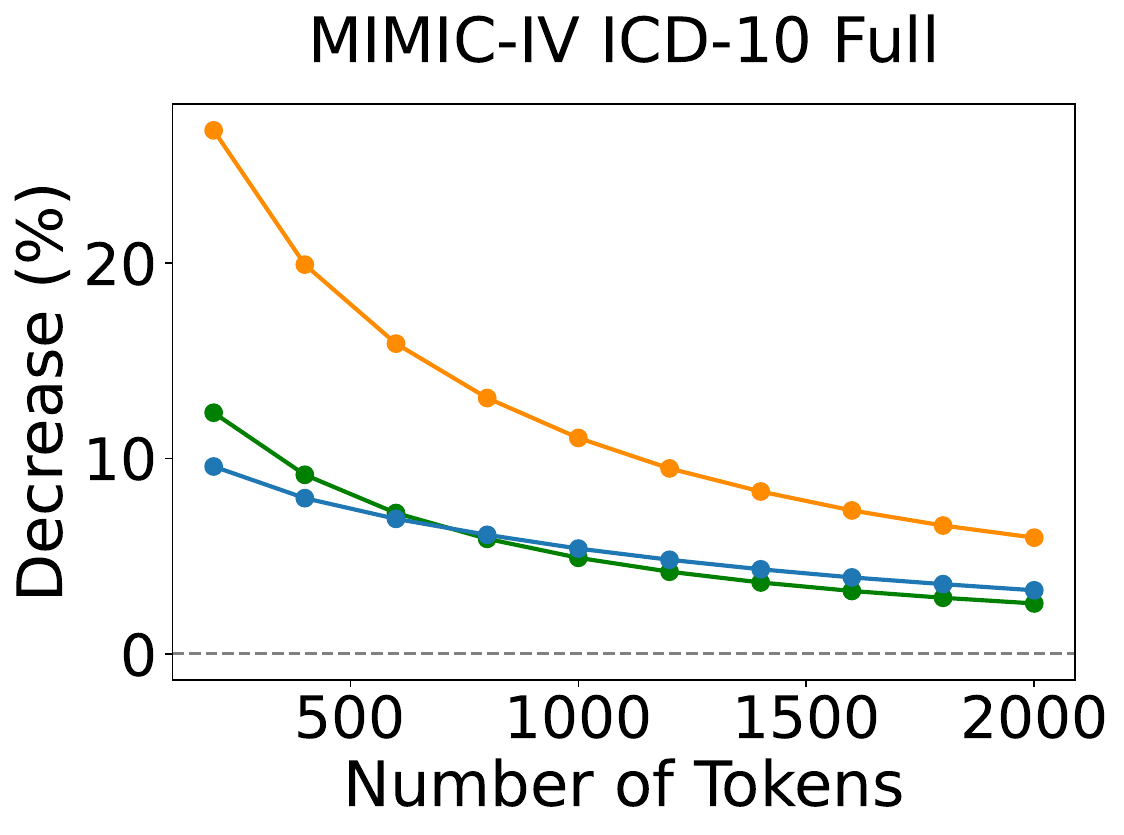}
    \end{subfigure}
    \begin{subfigure}{0.24\textwidth}
        \includegraphics[width=\linewidth]{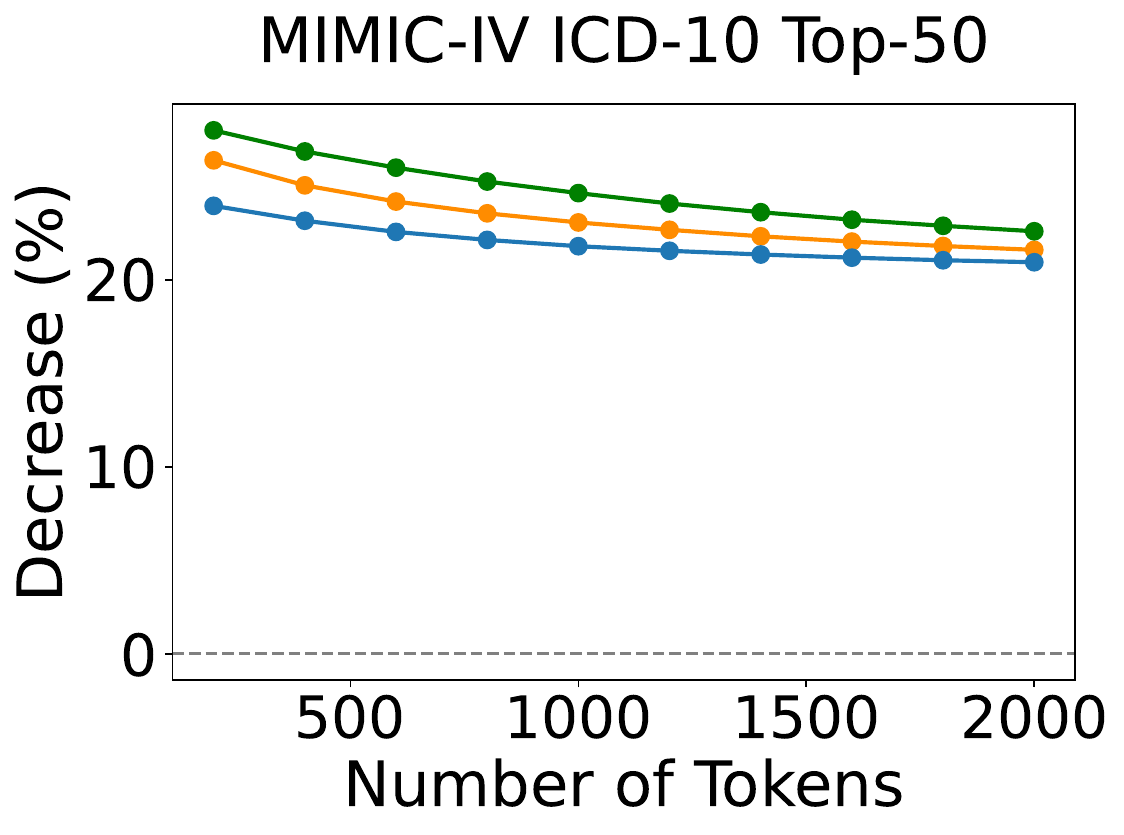}
    \end{subfigure}
    \begin{subfigure}{0.24\textwidth}
        \includegraphics[width=\linewidth]{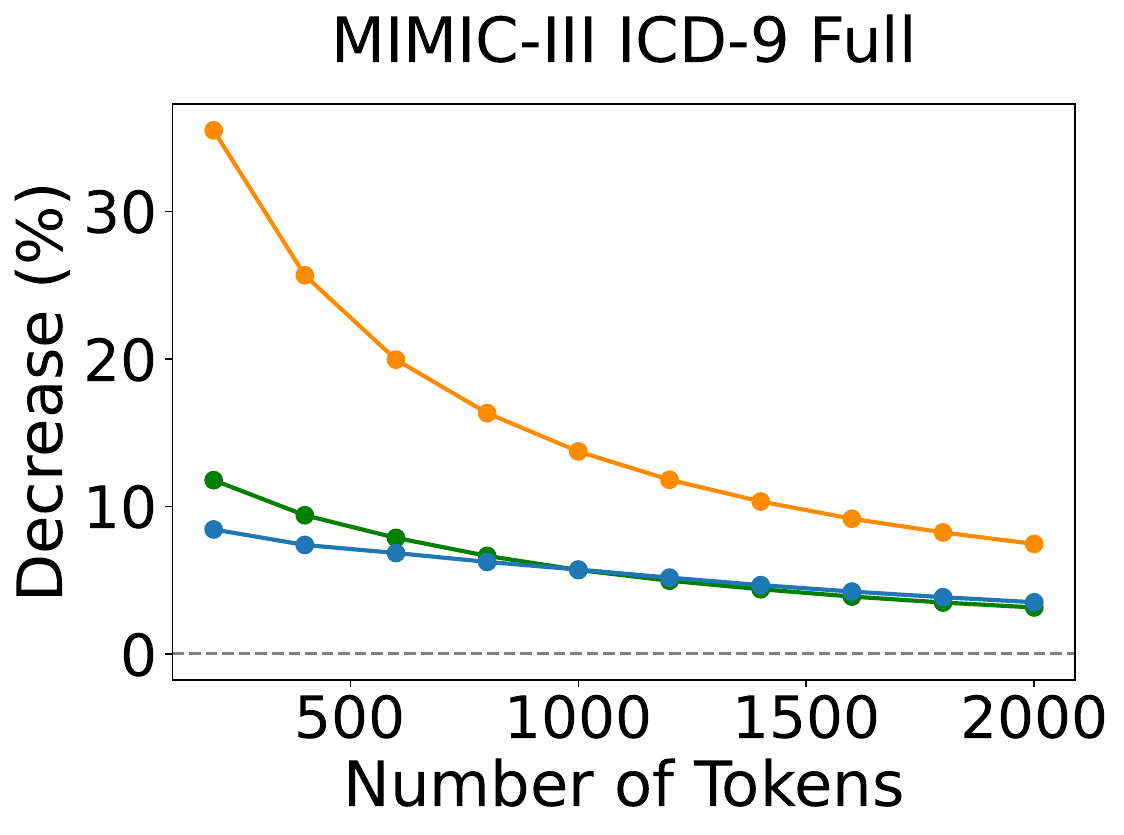}
    \end{subfigure}
    \begin{subfigure}{0.24\textwidth}
        \includegraphics[width=\linewidth]{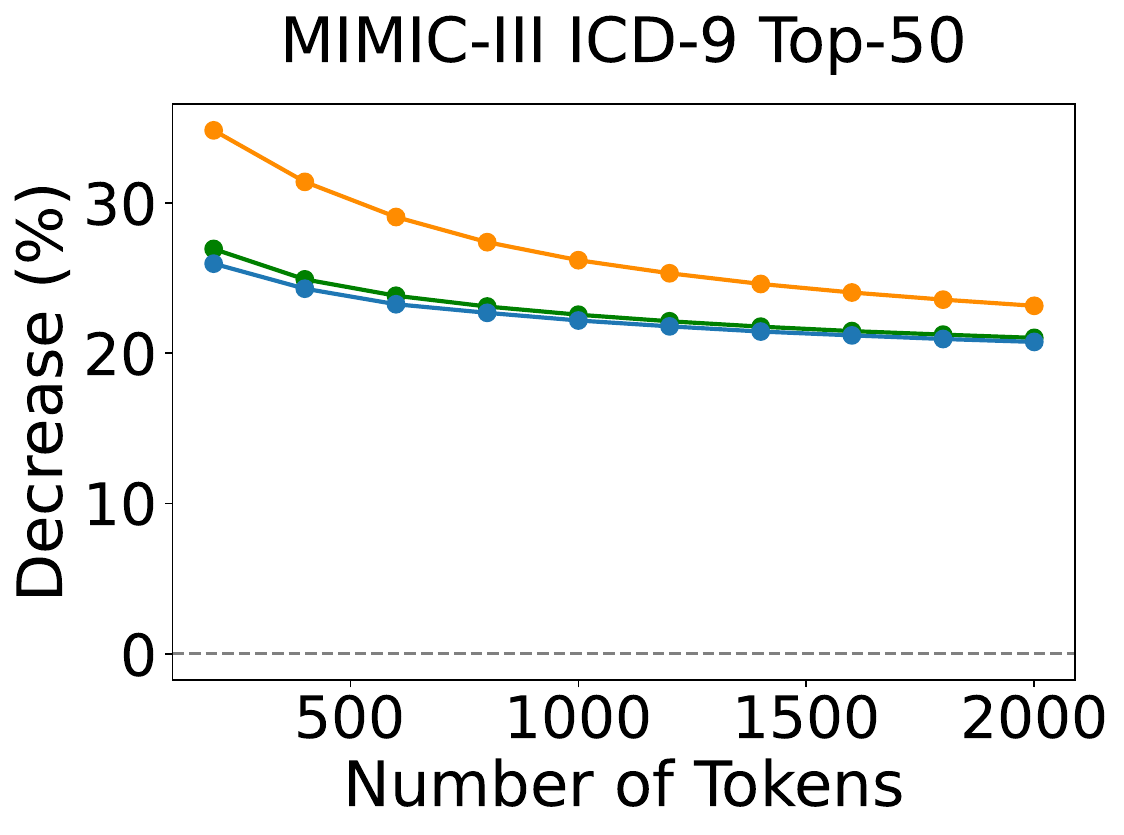}
    \end{subfigure}
    
    % -------- Row 2 Title --------
    Comprehensiveness of Models on Four MIMIC Datasets (Precision@5/8) $\uparrow$ \\[0.5em]
    
    \begin{subfigure}{0.24\textwidth}
        \includegraphics[width=\linewidth]{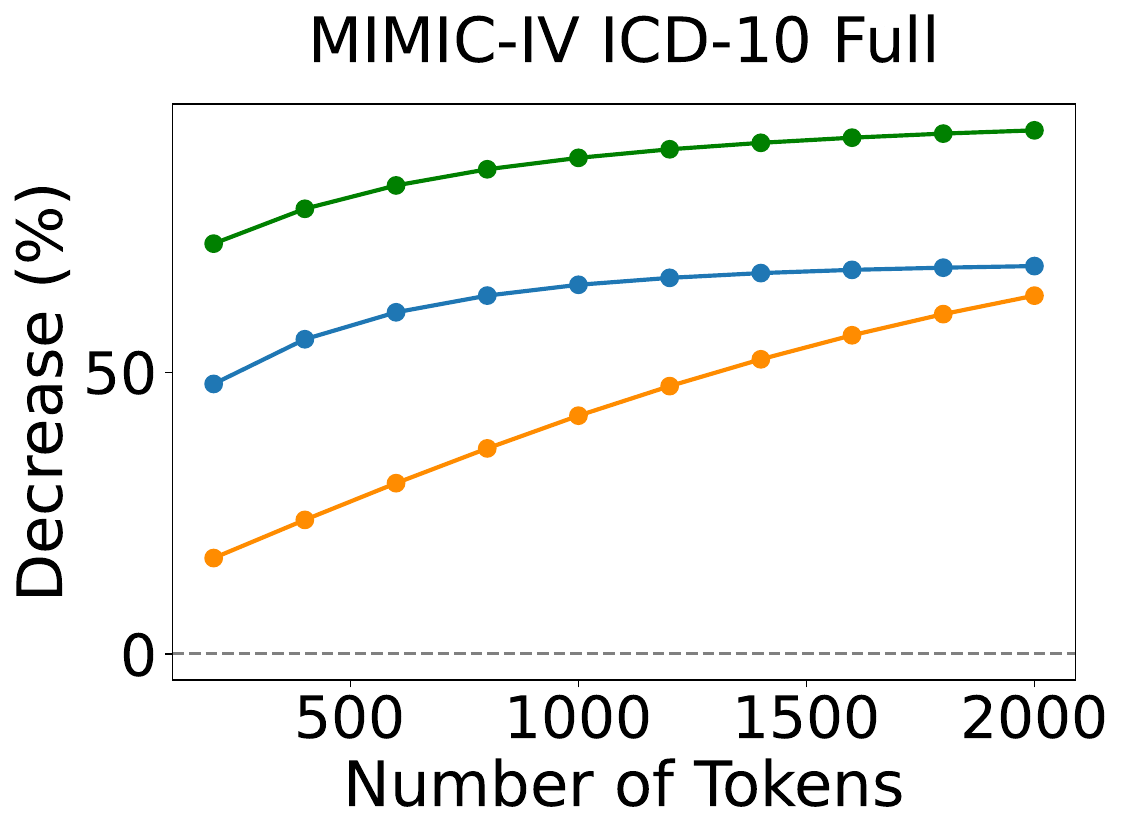}
    \end{subfigure}
    \begin{subfigure}{0.24\textwidth}
        \includegraphics[width=\linewidth]{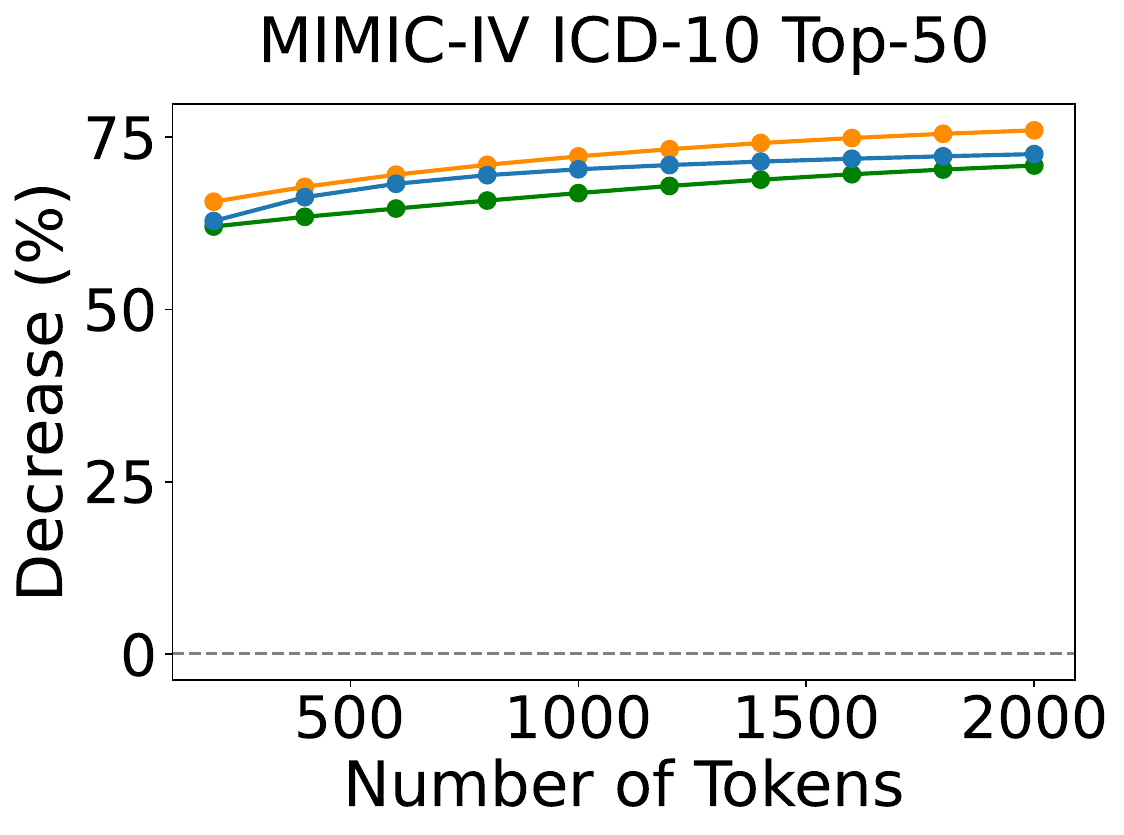}
    \end{subfigure}
    \begin{subfigure}{0.24\textwidth}
        \includegraphics[width=\linewidth]{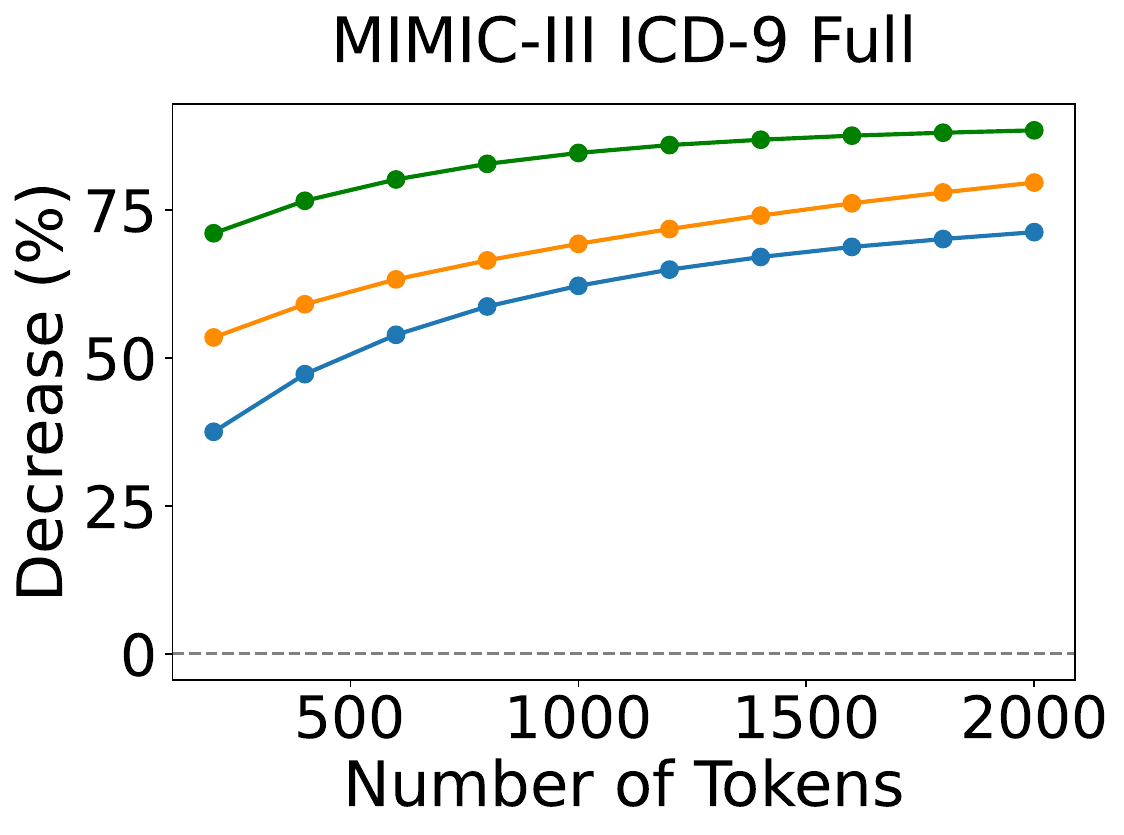}
    \end{subfigure}
    \begin{subfigure}{0.24\textwidth}
        \includegraphics[width=\linewidth]{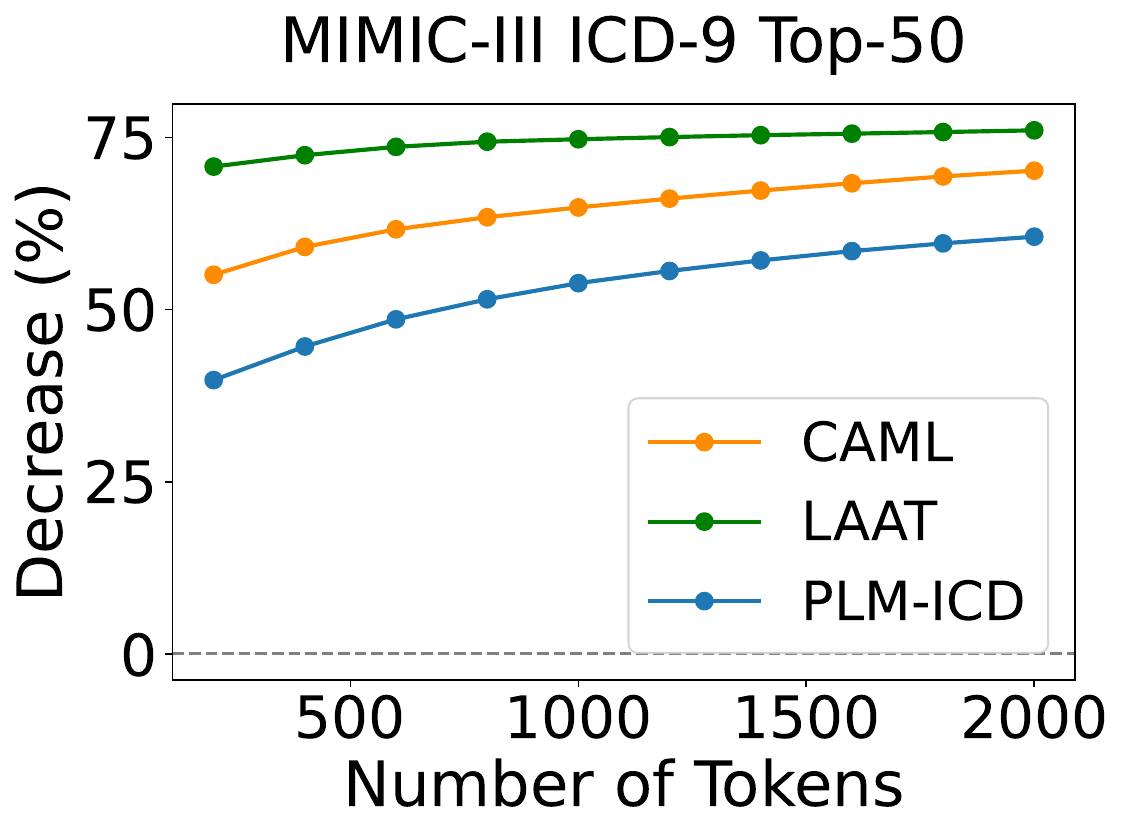}
    \end{subfigure}
    \vspace{-3mm}
    \caption{Faithfulness results of ICD coding models on four MIMIC datasets. The y-axis represents the decrease ratio, computed based on Precision@N scores (N = 8 for the Full set and N = 5 for the Top-50 set) as $(\mathrm{S}_{\mathrm{orig}} - \mathrm{S}_{\mathrm{com/suff}})/\mathrm{S}_{\mathrm{orig}} \times 100\%$
, where $\mathrm{S}_{\mathrm{orig}}$ denotes the performance obtained with the original input, and, $\mathrm{S}_{\mathrm{com/suff}}$  denotes the performance obtained when the input is modified by removing or retaining the rationales. The x-axis denotes the number of most-attended tokens selected. $\uparrow$ denotes higher is better; $\downarrow$ denotes lower is better.}
    \label{fig:combined}
\end{figure*}
\paragraph{Multi-objective Learning}  

One way to embed rational learning into ICD coding is to incorporate  another learning objective alongside the primary classification objective of the ICD coding model, minimizing the discrepancy between the model-generated rationales (controlled by attention weights) and the provided rationale labels by LLMs.
We define the rationale labels associated with a code label \( l \) as \( \mathbf{r}_{il} \). 
To represent these rationales, we construct a binary mask matrix \( \mathbf{M}_{i,l} \), 
where a value of 1 indicates that the corresponding token is part of the rationale, 
and 0 otherwise. 
We design the following rationale generation loss by applying
binary cross-entropy to the attention weights and  rationale masks:
%
%ratio of parameters
% add an example 
%
\begin{equation}
\begin{split}
\small
\mathcal{L}_{\text{rationale}} = -\frac{1}{ N_D } \sum_{i=1}^{ N_D  } \sum_{l=1}^{  N_l  } \sum_{j=1}^{  N_t  } 
\left[ M_{i,l,j} \log \tilde{a}_{i,j,l} \right. \\
\left. + (1 - M_{i,l,j}) \log (1 - \tilde{a}_{i,j,l}) \right].
\end{split}
\end{equation}
The final ICD coding model is trained by minimizing the combined loss of $\mathcal{L}_{\text{coding}}+\mathcal{L}_{\text{rationale}}$.

\paragraph{Learning by NER Formulation}  
An alternative approach  to enable both rationale and ICD code learning is to leverage the  rationale labels provided by LLMs to train a NER model.
Specifically, each rationale is treated as an entity with its corresponding ICD code assigned as its class label, while the span that is not identified as a rationale is assigned a null class. 
The success of entity recognition contributes to  both ICD code classification and rationale extraction.
As a result, the ICD coding and rationale learning tasks are 
neatly converted to one single NER task, solved by following the standard NER training.

\paragraph{Enhanced Supervision by Few-shot Prompting}

Manual rationale annotation is time‑consuming and costly, limiting scalability to large datasets.
Although LLMs provide a promising alternative for automatic annotation, they are susceptible to hallucinations and inaccuracies \cite{li2023halueval, ji2023survey}, particularly in expert domains such as healthcare \cite{nagar2024umedsum}. 
Prior studies have shown that few‑shot prompting, where models are provided with few examples, can substantially improve generation quality \cite{sivarajkumar2024empirical}.
Motivated by this, we further incorporate a small amount of example annotations provided by our constructed rationale dataset into the prompts of Gemini 2-Flash.
Details of the prompt design are provided in Appendix B.
The rationales generated are then used to supervise the rationale learning.

\section{Experiments and Result Analysis}

We conduct evaluation using both the MIMIC-III dataset with ICD-9 codes and the more recent MIMIC-IV benchmark with ICD-10 codes. %, following the latest state-of-the-art work \cite{edin2023automated, grundmann2024data, falis2024can}.
To assess the ICD coding performance, we use F1, AUC, and Precision@N, following the standard practice.
To conduct experiments on plausibility, we use the  SNOMED CT Entity Linking Challenge dataset \cite{hardman2023snomed} to implement the naive rationale extraction, i.e., entity-level rationales.
We compare both cloud-based and locally deployed LLMs, including Gemini 2-Flash,  Gemini 1.5-Pro and LLaMA-3.3, to implement the strong rationale LLM-generated rationales.
To ensure a fair comparison between the three models of CAML, LAAT and PLM-ICD, we follow the experimental setup of a reproducibility framework \citet{edin2023automated}. We implement NER models under the default hyperparameter settings in \citet{yang2020clinical}. More details on datasets and implementations are provided in Appendices A and B.

 % along with an example illustrating the computation of overlap scores for multiple candidate matches of a generated span.
\subsection{Comparison of ICD Rationales}

\subsubsection{Faithfulness of CAML, LAAT and PLM-ICD: \textit{How Explainable Are Rationales to Machines?}}

Figure~\ref{fig:combined} compares the faithfulness of the three  ICD coding models with \textit{Top N tokens} rationale selection strategy. Results are reported in terms of precision@5 for the Top-50 datasets and precision@8 for the Full datasets. More results across more metrics with both  \textit{Top N tokens} and \textit{Top $p$\% tokens} are reported in Appendix F.

For sufficiency, \textit{PLM-ICD performs comparably to LAAT, and both models outperform CAML} on the MIMIC-IV ICD-10 Full dataset as well as the two MIMIC-III ICD-9 datasets. %, since a lower performance decrease ratio indicates better sufficiency. 
%Notably, PLM-ICD demonstrates the highest sufficiency when input only 200 tokens, resulting in approximately a 10\% and 25\% performance reduction on the Top-50 and Full datasets, respectively. The 200 tokens constitute 13.21\%, 16.28\%, 11.42\%, and 12.92\% of the entire input text across the four datasets (in the same order as presented in the figures). This indicates the presence of substantial noisy content in the inputs and underscores PLM-ICD’s ability to effectively focus on the most influential rationales associated with the target labels.
PLM-ICD achieves the highest sufficiency using only 200 tokens, with a 10\% and 25\% performance drop on the Top-50 and Full datasets, respectively. These tokens represent just 11–16\% of the input text. %this highlights substantial noise in the data and PLM-ICD’s ability to focus on the most influential rationales. 
\textit{For comprehensiveness, removing rationales causes the greatest performance drop in LAAT}, particularly on the MIMIC-IV ICD-10 Full and two MIMIC-III ICD-9 datasets. In contrast, PLM-ICD shows smaller declines, even less than CAML on the ICD-9 datasets, suggesting that its pre-training enhances robustness by effectively extracting relevant information from residual inputs.

\begin{table}[t]
\centering
    \captionof{table}{Document-level and code-level plausibility results (F1) of \textcolor{orange}{\uline{naive entity-level rationales}}, \textcolor{blue!60!black}{\dashuline{strong LLM-generated rationales}}, \textcolor{green!30!black}{\uuline{model-generated rationales}}. 200 denotes selected tokens, “w/o” and “w/” indicate the absence or inclusion of few-shot examples in prompts. All results are reported as percentages.}
\resizebox{\columnwidth}{!}{%
\begin{tabular}{lccccc}
\hline
\textbf{Model / Code} & \textbf{Settings} & \textbf{Exact SM} & \textbf{PI SM} & \textbf{Exact TM} & \textbf{PI TM} \\
\hline
\multicolumn{6}{c}{\textbf{Document-level Evaluation}} \\ 
\hdashline
\textcolor{orange}{\uline{Entity-Linking}}      & -- & 10.3 & 9.2  & 6.3  & 6.1  \\
\textcolor{blue!60!black}{\dashuline{Gemini 2-Flash}} & -- & \textbf{21.6} & \textbf{24.1} & \textbf{30.1} & \textbf{37.3} \\
\textcolor{blue!60!black}{\dashuline{Gemini 1.5-Pro}} & -- & 13.5 & 14.6 & 20.6 & 26.0 \\
\textcolor{blue!60!black}{\dashuline{LLaMA-3.3 Ins}}     & -- & 18.6 & 21.5 & 27.8 & 35.0 \\
\textcolor{blue!60!black}{\dashuline{LLaMA-3.3 AWQ}}     & -- & 17.5 & 20.1 & 27.2 & 34.1 \\
\textcolor{green!30!black}{\uuline{CAML}}         & 200 & 0.1  & 0.2  & 3.1  & 6.5  \\
\textcolor{green!30!black}{\uuline{LAAT}}         & 200 & 0.7  & 0.8  & 5.0  & 7.2  \\
\textcolor{green!30!black}{\uuline{PLM-ICD}}      & 200 & 0.5  & 0.7  & 4.3  & 8.8  \\
\hline
\multicolumn{6}{c}{\textbf{Code-level Evaluation} (\textcolor{blue!60!black}{\dashuline{Gemini 2-Flash})}} \\ 
\hdashline
\multirow{2}{*}{I10}    & w/o & 50.3 & 63.1 & 27.0 & 32.2 \\
       & w/  & \textbf{60.5} & \textbf{78.2} & \textbf{53.8} & \textbf{66.4} \\
\multirow{2}{*}{E785}   & w/o & 62.5 & 76.6 & 50.2 & 59.7 \\
       & w/  & \textbf{73.2} & \textbf{89.5} & \textbf{66.7} & \textbf{78.2} \\
\multirow{2}{*}{Z7901}  & w/o & 9.0  & 9.7  & 35.8 & 43.3 \\
       & w/  & \textbf{13.0} & \textbf{16.8} & \textbf{45.8} & \textbf{54.8} \\
\multirow{2}{*}{I4891}  & w/o & 11.8 & 16.7 & 28.0 & 34.8 \\
       & w/  & \textbf{24.7} & \textbf{36.9} & \textbf{47.9} & \textbf{56.2} \\
\multirow{2}{*}{E119}   & w/o & 40.8 & 48.8 & 36.2 & 37.4 \\
       & w/  & \textbf{44.2} & \textbf{52.9} & \textbf{43.6} & \textbf{44.8} \\
\hline
\label{tab:plausibility_all}
\end{tabular}
}
\end{table}
%\vspace{-8mm}
\subsubsection{Plausibility of Naive, Strong and Model-generated Rationales: \textit{How Explainable Are Rationales to Experts?}}

Table \ref{tab:plausibility_all} summarises the plausibility results across different types of rationales. The results show that model-generated rationales yield very low metric scores, which indicates that they do not align with human explanations. Additional results across a wider range of thresholds are provided in Appendix F. Entity-level rationales rank second, while LLM-generated rationales perform the best, with Gemini 2-Flash achieving the highest scores. An additional case study comparing these rationales with human annotations is provided in Appendix H.
%Notably, LLaMA-3.3-instruct delivers performance very close to that of the better Gemini model. A case study comparing three types of rationales with human annotations is provided in Appendix G.

%\subsubsection{Gold Standard}
%The gold standards used across different ICD coding models vary slightly due to differences in their data preprocessing. Mapping the selected rationales from the processed input back to their exact positions in the original documents is infeasible because of modifications such as character removal and whitespace trimming during preprocessing. To address this, we processed the human annotations following each model’s specific preprocessing pipeline and conducted separate evaluation. Details of the data processing procedures are provided in Appendix~\ref{dataset_construction}. For all other models, comparisons were made against the original human annotations. 
\paragraph{Naive Rationales: \textit{Are Linked Entities Sufficient to Serve as Rationales?}}
The performance of directly linked entities ranks in the middle among the three types of rationales. However, its actual matching quality is underestimated, as the entity linking and MIMIC-IV ICD-10 dataset use different coding schemes despite referring to the same clinical mentions. For instance, in sample HADM ID: 24813967, all occurrences of `\textit{fall}' are assigned `\textit{R29.6 – Repeated falls}', whereas MIMIC-IV ICD-10 labels the case with `\textit{W01.0XXA – Fall on same level from slipping, tripping and stumbling without subsequent striking against object, initial encounter}'. \textit{In conclusion, directly linked entities can serve as rationales to a certain extent.}
\paragraph{Strong Rationales: \textit{Can LLMs Generate High-Quality Rationales?}}
\textit{The Gemini 2-Flash model achieves the highest performance among all comparisons.} However, both more cost‑effective local LLaMA-3.3 models deliver competitive results, with only a minor drop observed in the quantized variant AWQ. Importantly, the AWQ requires significantly fewer computational resources-approximately 40 GB of memory, compared to the Instruct model.
%In the code-level evaluation, we conduct experiments for the five most frequent codes in the dataset: I10, E785, Z7901, I4891, and E119. We analyze the rationales generated by Gemini 2-Flash with and without few-shot examples in the prompts to further investigate whether incorporating example annotations can further enhance Gemini’s performance. The prompt design is detailed in Appendix C. In the incorporating few-shot examples experiments, ror each code, five samples containing that code are randomly selected, and their human‑annotated rationales are used as examples within the prompt.
\begin{table}[h!]
        \centering
        \captionof{table}{ICD coding performance of LLM-guided supervised approaches. The experiments are conducted on the Top-50 code settings. All results are reported as percentages.}
\resizebox{\columnwidth}{!}{%
\begin{tabular}{ccccccc}
\hline
\textbf{Model} & \textbf{F1-Mac} & \textbf{F1-Mic} & \textbf{P-Mac} & \textbf{P-Mic} & \textbf{R-Mac} & \textbf{R-Mic} \\
\hline
PLM-ICD & 68.18 & 73.40 & 68.71 & 73.48 & 69.38 & 73.33\\
Multi-objective & 67.93 & 73.26 & 67.60 & 72.66 & 69.53 & 73.87 \\
\hline
PLM-ICD & 61.09 & 68.18 & 60.06 & 67.62 & 63.90 & 68.76 \\
NER & 53.46 & 67.75 & 49.21 & 60.93 & 61.79 & 76.30 \\
\hline
\label{supervised_coding}
\end{tabular}
}
\end{table}
\subsubsection{Few-shot Prompting: \textit{Does It Improve Plausibility of LLM-Generated Rationales?}}
In the code-level evaluation, we conduct experiments on the five most frequent codes in the dataset: I10 (essential (primary) hypertension), E785 (hyperlipidemia, unspecified), Z7901 (long term (current) use of anticoagulants), I4891 (unspecified atrial fibrillation (AFib)), and E119 (type 2 diabetes mellitus without complications). We analyze rationales generated by Gemini 2-Flash with and without few-shot examples in the prompts to examine whether incorporating human-annotated examples can further enhance Gemini’s performance. In the few-shot experiments, examples from five test instances per code are included in the prompts, and Gemini then regenerates rationales for the remaining samples, which are then re-evaluated using the same plausibility metrics. \textit{Incorporating examples yields substantial improvements in F1 scores} across all five codes shown in Table \ref{tab:plausibility_all}, with average gains of 39.90\%, 48.67\%, 50.31\%, and 49.01\% in Exact Span Match, PI Span Match, Exact Token Match, and PI Token Match, respectively. These results demonstrate that examples from our rationale dataset guide Gemini in generating more plausible rationales. More results for this experiment are provided in Appendix G.

%\vspace{-7mm}
\begin{table}[h!]
        \centering
        \captionof{table}{Document-level and code-level plausibility results (F1) of LLM-guided supervised approaches. The experiments for multi-objective learning and NER are conducted on different datasets using the Top-50 code settings. 50 denotes selected tokens, “w/o” and “w/” indicate the absence or inclusion of few-shot examples in prompts. \textbf{when constructing the training datasets}. All results are reported as percentages.}
\resizebox{\columnwidth}{!}{%
\begin{tabular}{lccccc}
\hline
\textbf{Model} & \textbf{Settings} & \textbf{Exact SM} & \textbf{PI SM} & \textbf{Exact TM} & \textbf{PI TM} \\
\hline
\multicolumn{6}{c}{\textbf{Document-level Evaluation}} \\
\hdashline
PLM-ICD          & 50 & 2.7  & 3.0  & 9.5  & 12.2 \\
Multi-objective  & 50 & \textbf{3.9}  & \textbf{4.1}  & \textbf{10.6} & \textbf{13.2} \\
\hdashline
PLM-ICD          & 50 & 4.1  & 4.3  & 8.1  & 12.0 \\
Gemini 2-Flash   & -- & 18.2 & 23.0 & \textbf{29.4} & \textbf{31.4} \\
NER              & -- & \textbf{26.5} & \textbf{30.6} & 21.8 & 27.0 \\
\hline
\multicolumn{6}{c}{\textbf{Code-level Evaluation} (NER)} \\
\hdashline
\multirow{2}{*}{I10}    & w/o & 55.4 & 76.9 & 55.8 & 75.3 \\
                        & w/  & \textbf{62.5} & \textbf{85.2} & \textbf{64.9} & \textbf{86.2} \\
\multirow{2}{*}{E785}   & w/o & 67.9 & 85.5 & 61.6 & 75.2 \\
                        & w/  & \textbf{70.3} & \textbf{87.0} & \textbf{63.0} & \textbf{76.3} \\
\multirow{2}{*}{Z7901}  & w/o & 10.3 & \textbf{13.8} & \textbf{25.5} & 38.6 \\
                        & w/  & \textbf{10.8} & 12.0 & 22.5 & \textbf{40.7} \\
\multirow{2}{*}{I4891}  & w/o & 22.2 & 37.7 & 38.5 & 48.8 \\
                        & w/  & \textbf{26.1} & \textbf{43.3} & \textbf{40.7} & \textbf{54.5} \\
\multirow{2}{*}{E119}   & w/o & \textbf{49.4} & \textbf{62.0} & \textbf{53.2} & \textbf{62.4} \\
                        & w/  & 45.8 & 58.5 & 49.7 & 60.2 \\
\hline
\label{llm-guided_plausibility}
\end{tabular}
}
\end{table}
%\vspace{-5mm}
\subsection{Results for LLM-Guided Rationale Learning}
\subsubsection{Standard Prompting: \textit{Does It Improve ICD Coding Performance and Rationale Plausibility?}}
%Supervised by weak rationale labels generated by Gemini 2-Flash, multi-objective learning approach improves the plausibility (Table \ref{llm-guided_plausibility}), while still keep the ICD coding performance(\ref{supervised_coding}). While NER approach which is a separete learning formulation from PLM-ICD, it achieves the highest span-level plausibility, even surpassing that of its `teacher model' - Gemini 2-Flash. However, the NER model exhibits substantially lower overall coding performance than PLM-ICD since it's not a specific ICD coding model, it's for recognising the rationales for a set of ICD codes.
Supervised by weak rationale labels generated by Gemini 2-Flash, the multi-objective learning approach--which introduces the additional rationale-targeted objective--does not degrade ICD coding performance (Table \ref{supervised_coding}); instead, it improves plausibility by approximately 1\% (F1) across all four metrics (Table \ref{llm-guided_plausibility}). %improves rationale plausibility (Table \ref{llm-guided_plausibility}) while maintaining ICD coding performance (Table \ref{supervised_coding}). 

For the NER-based approach, which adopts a learning formulation distinct from PLM-ICD, there exists a clear trade-off between prediction accuracy and rationale plausibility, e.g., 12.49\% lower in coding performance (F1-macro) but on average 363\% higher plausibility than PLM-ICD. Notably it achieves the highest span-level plausibility, even surpassing its teacher model, Gemini 2-Flash. This trade-off arises because the NER model is specifically designed for entity recognition. It is trained on rationale–code pairs, focusing on highlighting tokens and assigning labels for each token, rather than to maximise classification accuracy, whereas PLM-ICD is trained on full documents, focusing on optimising document-level classification accuracy. Its competing objectives can reduce its raw predictive performance. However, the NER model benefits from learning entity patterns across all training samples, where these LLM generated training samples can be unstable and may miss some spans. This enables NER to consistently recognize complete instances of relevant entities, and this is why the NER model surpasses its teacher model, Gemini 2-Flash, in plausibility. Moreover, it generates stable outputs for all labels simultaneously, whereas Gemini is prompted with one code at a time; providing the entire code set leads to incomplete recognition. However, since the NER model is designed to identify rationales for a set of ICD codes rather than to perform coding directly, it exhibits substantially lower overall coding performance compared to PLM-ICD. Additionally, the NER-based approach offers an alternative, cost-free method for generating rationales. The results presented correspond to the \textit{top 50 tokens}. Additional results under a wider range of experimental settings are provided in Appendix G. Further results for the NER-based approach across different training data sizes are presented in Appendix I.

\subsubsection{Few-shot Prompting: \textit{Does It Improve LLM-Guided  Rationale Learning?}}
%In the previous experiments and analysis we have concluded that including few-shot examples in the prompts enhances the rationale generation. In this experiment, we aim to investigate whether the enhanced rationales help train better NER models for the rationale recognition.
%
%A separate NER model is trained using Gemini-generated rationales, both with and without few-shot examples in the prompts, for each of the top five codes individually. 
%Building on our previous experiments and analyses, which demonstrated that incorporating few-shot examples in prompts enhances rationale generation, 
This experiment further investigates whether the enhanced rationales generated through few-shot prompting can facilitate the rationale learning process. Specifically, we train separate NER models using Gemini-generated rationales, both with and without few-shot examples in the prompts, for each of the five most frequent codes.
%The rationale plausibility results are reported in the bottom half of Table \ref{llm-guided_plausibility}.

Table \ref{llm-guided_plausibility} shows that the \textit{NER formulation is effective for rationale recognition for specific single code}. It significantly outperforms the teacher models (w/o results in Table \ref{tab:plausibility_all}) across all codes and metrics, achieving average improvements of 28.49\%, 45.71\%, 37.01\%, and 51.21\% on the four metrics, respectively. %Among all codes, E785 and I10 achieve higher performance, which is attributed to their higher frequency, thereby benefiting model training with richer supervision data. Furthermore, \uline{the model trained on data generated with few-shot examples in the prompts consistently outperforms the one trained without such examples in most cases}, with average improvements of 6.30\%, 1.74\%, 1.19\%, 5.91\%. This further demonstrates that the enhanced rationales generated through few-shot prompting using the examples from out dataset further enhance the training of rationale learning models.
Among all codes, E785 and I10 achieve higher performance, which is attributed to their higher frequencies, providing the model with richer supervision signals. Furthermore, \textit{a model trained on data generated with few-shot examples in the prompts consistently outperforms one trained without such examples in most cases}, with average improvements of 6.30\%, 1.74\%, 1.19\%, and 5.91\%. These results further demonstrate that the enhanced rationales generated through few-shot prompting using examples from our dataset further enhance the training of rationale learning models. See Appendix G for complete results across all metrics.

\section{Conclusion}

In this study, we introduce a rationale dataset specifically designed for ICD coding. We evaluate the faithfulness of coding models and the plausibility of three types of rationales, among which Gemini 2-Flash achieves the best performance. We examine LLM-guided rationale learning approaches, where the NER formulation demonstrates strong potential, as both coding and rationale extraction tasks can be unified under a single NER framework. This approach achieves the highest span-level plausibility. Moreover, incorporating human-annotated examples from our dataset into prompts enhances both rationale generation and rationale learning process.
%We introduce a high-quality rationale dataset to evaluate ICD coding model faithfulness and the plausibility of three rationale types. We also propose LLM-guided rationale-learning approaches, showing that LLM-generated rationales achieve the highest plausibility and that few-shot prompting enhances both rationale generation and learning.

\section{Limitations}
While our study contributes to advancing both rationale evaluation and rationale learning in ICD coding area by providing benchmark resource and empirical analyses, it also has several limitations that suggest directions for future research. 

First, annotation is a resource-intensive process that demands domain expertise as well as significant time and cost investment. Our dataset currently comprises 150 samples, constrained by budget limitations. With additional funding or institutional support, the scale of annotations could be expanded in future work. With a sufficient number of human-annotated rationale labels, supervision using these labels becomes feasible. This enables a direct comparison between models trained with real labels and those trained with weak labels. Additionally, these 150 samples cover only 989 distinct codes, whereas MIMIC-IV contains 7,942 codes. Increasing the number of samples would improve the diversity of codes.

Second, we evaluate three classic attention-based ICD coding models—CAML, LAAT, and PLM-ICD. There are also many other models that incorporate label-wise attention layers, such as the recent CoRelation \cite{luo2024corelation} and MSAM \cite{gomes2024accurate}. However, for plausibility evaluation, the rationales generated by these attention-based models tend to have low plausibility.

Third, the evaluation of naïve entity-level rationales could be further refined through a more precise alignment of coding schemes.

Finally, our NER models were trained on relatively small datasets of 5,000 randomly selected samples. Future work could extend these experiments to the full datasets.

\section*{Acknowledgement}
This research is part of the IN-CYPHER programme and is supported by the National Research Foundation, Prime Minister’s Office, Sin-
gapore under its Campus for Research Excellence and Technological Enterprise (CREATE) programme. We are grateful for the support
provided by Research IT in the form of access to the Computational Shared Facility at The University of Manchester. We also thank the anony-
mous ARR reviewers for their feedback that helped us improve the paper further.

\bibliography{custom}

@inproceedings{nguyen2018computer,
  title={Computer-assisted diagnostic coding: effectiveness of an NLP-based approach using SNOMED CT to ICD-10 mappings},
  author={Nguyen, Anthony N and Truran, Donna and Kemp, Madonna and Koopman, Bevan and Conlan, David and O’Dwyer, John and Zhang, Ming and Karimi, Sarvnaz and Hassanzadeh, Hamed and Lawley, Michael J and others},
  booktitle={AMIA Annual Symposium Proceedings},
  volume={2018},
  pages={807},
  year={2018},
  organization={American Medical Informatics Association}
}

@article{amann2020explainability,
  title={Explainability for artificial intelligence in healthcare: a multidisciplinary perspective},
  author={Amann, Julia and Blasimme, Alessandro and Vayena, Effy and Frey, Dietmar and Madai, Vince I and Precise4Q Consortium},
  journal={BMC medical informatics and decision making},
  volume={20},
  pages={1--9},
  year={2020},
  publisher={Springer}
}

@article{mullenbach2018explainable,
  title={Explainable prediction of medical codes from clinical text},
  author={Mullenbach, James and Wiegreffe, Sarah and Duke, Jon and Sun, Jimeng and Eisenstein, Jacob},
  journal={arXiv preprint arXiv:1802.05695},
  year={2018}
}

@article{dong2021explainable,
  title={Explainable automated coding of clinical notes using hierarchical label-wise attention networks and label embedding initialisation},
  author={Dong, Hang and Su{\'a}rez-Paniagua, V{\'\i}ctor and Whiteley, William and Wu, Honghan},
  journal={Journal of biomedical informatics},
  volume={116},
  pages={103728},
  year={2021},
  publisher={Elsevier}
}

@inproceedings{lovelace2020dynamically,
  title={Dynamically extracting outcome-specific problem lists from clinical notes with guided multi-headed attention},
  author={Lovelace, Justin and Hurley, Nathan C and Haimovich, Adrian D and Mortazavi, Bobak J},
  booktitle={Machine Learning for Healthcare Conference},
  pages={245--270},
  year={2020},
  organization={PMLR}
}

@inproceedings{wang2022novel,
  title={A novel framework based on medical concept driven attention for explainable medical code prediction via external knowledge},
  author={Wang, Tao and Zhang, Linhai and Ye, Chenchen and Liu, Junxi and Zhou, Deyu},
  booktitle={Findings of the Association for Computational Linguistics: ACL 2022},
  pages={1407--1416},
  year={2022}
}

@article{van2022patient,
  title={This patient looks like that patient: Prototypical networks for interpretable diagnosis prediction from clinical text},
  author={Van Aken, Betty and Papaioannou, Jens-Michalis and Naik, Marcel G and Eleftheriadis, Georgios and Nejdl, Wolfgang and Gers, Felix A and L{\"o}ser, Alexander},
  journal={arXiv preprint arXiv:2210.08500},
  year={2022}
}

@article{kim2022can,
  title={Can current explainability help provide references in clinical notes to support humans annotate medical codes?},
  author={Kim, Byung-Hak and Deng, Zhongfen and Yu, Philip S and Ganapathi, Varun},
  journal={arXiv preprint arXiv:2210.15882},
  year={2022}
}

@article{cheng2023mdace,
  title={MDACE: MIMIC Documents Annotated with Code Evidence},
  author={Cheng, Hua and Jafari, Rana and Russell, April and Klopfer, Russell and Lu, Edmond and Striner, Benjamin and Gormley, Matthew R},
  journal={arXiv preprint arXiv:2307.03859},
  year={2023}
}

@article{edin2024unsupervised,
  title={An Unsupervised Approach to Achieve Supervised-Level Explainability in Healthcare Records},
  author={Edin, Joakim and Maistro, Maria and Maal{\o}e, Lars and Borgholt, Lasse and Havtorn, Jakob D and Ruotsalo, Tuukka},
  journal={arXiv preprint arXiv:2406.08958},
  year={2024}
}

@article{gao2024optimising,
  title={Optimising the paradigms of human AI collaborative clinical coding},
  author={Gao, Yue and Chen, Yuepeng and Wang, Minghao and Wu, Jinge and Kim, Yunsoo and Zhou, Kaiyin and Li, Miao and Liu, Xien and Fu, Xiangling and Wu, Ji and others},
  journal={npj Digital Medicine},
  volume={7},
  number={1},
  pages={368},
  year={2024},
  publisher={Nature Publishing Group UK London}
}

@article{mendez2024outputs,
  title={From outputs to insights: a survey of rationalization approaches for explainable text classification},
  author={Mendez Guzman, Erick and Schlegel, Viktor and Batista-Navarro, Riza},
  journal={Frontiers in Artificial Intelligence},
  volume={7},
  pages={1363531},
  year={2024},
  publisher={Frontiers Media SA}
}

@article{deyoung2019eraser,
  title={ERASER: A benchmark to evaluate rationalized NLP models},
  author={DeYoung, Jay and Jain, Sarthak and Rajani, Nazneen Fatema and Lehman, Eric and Xiong, Caiming and Socher, Richard and Wallace, Byron C},
  journal={arXiv preprint arXiv:1911.03429},
  year={2019}
}

@article{vu2020label,
  title={A label attention model for ICD coding from clinical text},
  author={Vu, Thanh and Nguyen, Dat Quoc and Nguyen, Anthony},
  journal={arXiv preprint arXiv:2007.06351},
  year={2020}
}

@article{huang2022plm,
  title={PLM-ICD: Automatic ICD coding with pretrained language models},
  author={Huang, Chao-Wei and Tsai, Shang-Chi and Chen, Yun-Nung},
  journal={arXiv preprint arXiv:2207.05289},
  year={2022}
}

@inproceedings{edin2023automated,
  title={Automated medical coding on MIMIC-III and MIMIC-IV: a critical review and replicability study},
  author={Edin, Joakim and Junge, Alexander and Havtorn, Jakob D and Borgholt, Lasse and Maistro, Maria and Ruotsalo, Tuukka and Maal{\o}e, Lars},
  booktitle={Proceedings of the 46th international ACM SIGIR conference on research and development in information retrieval},
  pages={2572--2582},
  year={2023}
}

@article{johnson2020mimic,
  title={Mimic-iv},
  author={Johnson, Alistair and Bulgarelli, Lucas and Pollard, Tom and Horng, Steven and Celi, Leo Anthony and Mark, Roger},
  journal={PhysioNet. Available online at: https://physionet. org/content/mimiciv/1.0/(accessed August 23, 2021)},
  pages={49--55},
  year={2020}
}

@article{blundell2023health,
  title={Health information and the importance of clinical coding},
  author={Blundell, James},
  journal={Anaesthesia \& Intensive Care Medicine},
  volume={24},
  number={2},
  pages={96--98},
  year={2023},
  publisher={Elsevier}
}

@article{tzitzivacos2007international,
  title={International classification of diseases 10th edition (icd-10)},
  author={Tzitzivacos, Dimitri},
  journal={CME: Your SA Journal of CPD},
  volume={25},
  number={1},
  pages={8--10},
  year={2007},
  publisher={Health and Medical Publishing Group (HMPG)}
}

@inproceedings{pereira2006construction,
  title={Construction of a semi-automated ICD-10 coding help system to optimize medical and economic coding.},
  author={Pereira, Suzanne and N{\'e}v{\'e}ol, Aur{\'e}lie and Massari, Philippe and Joubert, Michel and Darmoni, Stefan},
  booktitle={MIE},
  pages={845--850},
  year={2006}
}

@inproceedings{crammer2007automatic,
  title={Automatic code assignment to medical text},
  author={Crammer, Koby and Dredze, Mark and Ganchev, Kuzman and Talukdar, Partha and Carroll, Steven},
  booktitle={Biological, translational, and clinical language processing},
  pages={129--136},
  year={2007}
}

@inproceedings{lita2008large,
  title={Large scale diagnostic code classification for medical patient records},
  author={Lita, Lucian Vlad and Yu, Shipeng and Niculescu, Stefan and Bi, Jinbo},
  booktitle={Proceedings of the Third International Joint Conference on Natural Language Processing: Volume-II},
  year={2008}
}

@article{catling2018towards,
  title={Towards automated clinical coding},
  author={Catling, Finneas and Spithourakis, Georgios P and Riedel, Sebastian},
  journal={International journal of medical informatics},
  volume={120},
  pages={50--61},
  year={2018},
  publisher={Elsevier}
}

@inproceedings{karimi2017automatic,
  title={Automatic diagnosis coding of radiology reports: a comparison of deep learning and conventional classification methods},
  author={Karimi, Sarvnaz and Dai, Xiang and Hassanzadeh, Hamed and Nguyen, Anthony},
  booktitle={BioNLP 2017},
  pages={328--332},
  year={2017}
}

@inproceedings{liu2021effective,
  title={Effective convolutional attention network for multi-label clinical document classification},
  author={Liu, Yang and Cheng, Hua and Klopfer, Russell and Gormley, Matthew R and Schaaf, Thomas},
  booktitle={Proceedings of the 2021 Conference on Empirical Methods in Natural Language Processing},
  pages={5941--5953},
  year={2021}
}

@article{yuan2022code,
  title={Code synonyms do matter: Multiple synonyms matching network for automatic ICD coding},
  author={Yuan, Zheng and Tan, Chuanqi and Huang, Songfang},
  journal={arXiv preprint arXiv:2203.01515},
  year={2022}
}

@article{michalopoulos2022icdbigbird,
  title={ICDBigBird: a contextual embedding model for ICD code classification},
  author={Michalopoulos, George and Malyska, Michal and Sahar, Nicola and Wong, Alexander and Chen, Helen},
  journal={arXiv preprint arXiv:2204.10408},
  year={2022}
}

@inproceedings{yogarajan2022concatenating,
  title={Concatenating BioMed-Transformers to Tackle Long Medical Documents and to Improve the Prediction of Tail-End Labels},
  author={Yogarajan, Vithya and Pfahringer, Bernhard and Smith, Tony and Montiel, Jacob},
  booktitle={International Conference on Artificial Neural Networks},
  pages={209--221},
  year={2022},
  organization={Springer}
}

@inproceedings{yang2022knowledge,
  title={Knowledge injected prompt based fine-tuning for multi-label few-shot icd coding},
  author={Yang, Zhichao and Wang, Shufan and Rawat, Bhanu Pratap Singh and Mitra, Avijit and Yu, Hong},
  booktitle={Proceedings of the conference on empirical methods in natural language processing. Conference on empirical methods in natural language processing},
  volume={2022},
  pages={1767},
  year={2022}
}

@article{hardman2023snomed,
  title={SNOMED CT Entity Linking Challenge},
  author={Hardman, Will and Banks, Mark and Davidson, Rory and Truran, Donna and Ayuningtyas, Nindya Widita and Ngo, Hoa and Johnson, Alistair and Pollard, Tom},
  journal={PhysioNet. Version},
  volume={1},
  number={0},
  year={2023}
}

@article{yang2020clinical,
  title={Clinical concept extraction using transformers},
  author={Yang, Xi and Bian, Jiang and Hogan, William R and Wu, Yonghui},
  journal={Journal of the American Medical Informatics Association},
  volume={27},
  number={12},
  pages={1935--1942},
  year={2020},
  publisher={Oxford University Press}
}

@misc{snomed2024us,
  author       = {{SNOMED-CT}},
  title        = {SNOMED CT U.S. Edition Release, September 1, 2024},
  howpublished = {\url{https://www.nlm.nih.gov/healthit/snomedct/us_edition.html}},
  year         = {2024}
}

@article{nagar2024umedsum,
  title={umedsum: A unified framework for advancing medical abstractive summarization},
  author={Nagar, Aishik and Liu, Yutong and Liu, Andy T and Schlegel, Viktor and Dwivedi, Vijay Prakash and Kaliya-Perumal, Arun-Kumar and Kalanchiam, Guna Pratheep and Tang, Yili and Tan, Robby T},
  journal={arXiv preprint arXiv:2408.12095},
  year={2024}
}

@article{li2023halueval,
  title={Halueval: A large-scale hallucination evaluation benchmark for large language models},
  author={Li, Junyi and Cheng, Xiaoxue and Zhao, Wayne Xin and Nie, Jian-Yun and Wen, Ji-Rong},
  journal={arXiv preprint arXiv:2305.11747},
  year={2023}
}

@article{ji2023survey,
  title={Survey of hallucination in natural language generation},
  author={Ji, Ziwei and Lee, Nayeon and Frieske, Rita and Yu, Tiezheng and Su, Dan and Xu, Yan and Ishii, Etsuko and Bang, Ye Jin and Madotto, Andrea and Fung, Pascale},
  journal={ACM computing surveys},
  volume={55},
  number={12},
  pages={1--38},
  year={2023},
  publisher={ACM New York, NY}
}

@article{sivarajkumar2024empirical,
  title={An empirical evaluation of prompting strategies for large language models in zero-shot clinical natural language processing: algorithm development and validation study},
  author={Sivarajkumar, Sonish and Kelley, Mark and Samolyk-Mazzanti, Alyssa and Visweswaran, Shyam and Wang, Yanshan},
  journal={JMIR Medical Informatics},
  volume={12},
  pages={e55318},
  year={2024},
  publisher={JMIR Publications Toronto, Canada}
}

@article{gomes2024accurate,
  title={Accurate and well-calibrated ICD code assignment through attention over diverse label embeddings},
  author={Gomes, Gon{\c{c}}alo and Coutinho, Isabel and Martins, Bruno},
  journal={arXiv preprint arXiv:2402.03172},
  year={2024}
}

@article{luo2024corelation,
  title={CoRelation: Boosting automatic ICD coding through contextualized code relation learning},
  author={Luo, Junyu and Wang, Xiaochen and Wang, Jiaqi and Chang, Aofei and Wang, Yaqing and Ma, Fenglong},
  journal={arXiv preprint arXiv:2402.15700},
  year={2024}
}
\clearpage
\appendix
\section{Additional Information on Datasets and Implementation}

\subsection{Datasets}

\paragraph{MIMIC Datasets} The Medical Information Mart for Intensive Care (MIMIC) dataset is a large-scale, de-identified database comprising health records of patients admitted to the emergency department or intensive care units at the Beth Israel Deaconess Medical Center \cite{johnson2020mimic}. The MIMIC-IV dataset covers over 65,000 ICU admissions and more than 200,000 emergency department visits between 2008 and 2019, coded using ICD-10. The MIMIC-III dataset covers admissions between 2001 and 2012, including 52,723 discharge summaries from 41,126 patients, coded using ICD-9. Table \ref{split} presents the statistics of dataset splits for all datasets used in model training, where ``Ra'' refers to the subset of data comprising documents with rationale labels generated by Gemini 2‑Flash and it is used for training the multi‑objective learning model. Top-50 subsets include only the 50 most frequent codes from the respective Full datasets.

\begin{table}[h!]
\centering
\caption{Dataset Split.}
\resizebox{1\columnwidth}{!}{%
\begin{tabular}{lccc}
\toprule
\textbf{Dataset} & \textbf{Train} & \textbf{Test} & \textbf{Dev}\\
\midrule
MIMIC-IV ICD-10 Full & 88988 & 19931 & 13360 \\
MIMIC-IV ICD-10 Top-50 & 83890 & 18776 & 12590\\
MIMIC-III ICD-9 Full & 47719 & 3372 & 1631\\
MIMIC-III ICD-9 Top-50 & 8066 & 1729 & 1573\\
MIMIC-IV ICD-10 Top-50 Ra & 83465 & 18665 & 12517\\
\bottomrule
\end{tabular}
}
\label{split}
\end{table}
\paragraph{Entity Linking Dataset} The SNOMED CT Entity Linking Challenge dataset \cite{hardman2023snomed}  comprises 272 discharge summaries from MIMIC-IV-Note, annotated with 6,624 unique SNOMED-CT concepts. Among these, 64 documents overlap with the MIMIC-IV ICD-10 dataset. To enable comparison, we align SNOMED CT concepts with ICD-10 codes using established mapping resource \cite{snomed2024us}.

Access to MIMIC datasets and entity linking dataset is granted upon completion of human-subjects research training (e.g., CITI program), registration on the PhysioNet platform, acceptance of the dataset’s Data Use Agreement, and subsequent retrieval of the data via PhysioNet’s web interface or command-line tools.

\subsection{Implementation Details}
\paragraph{On CAML, LAAT and PLM-ICD} 
To ensure a fair comparison between models, we follow the experimental setup of \citet{edin2023automated}, which provides a reproducibility framework for state-of-the-art ICD coding models. Details of the key parameter configurations for the three ICD coding models are summarized in Table \ref{config_ICD}. All models are trained for 20 epochs, although CAML and LAAT typically converge within 10 epochs on MIMIC-IV ICD-10 and both MIMIC-III datasets. The random seed is set to 1337 for all model training.

\setcounter{table}{4}
\begin{table*}[h!]
\centering
\caption{Configurations of CAML, LAAT and PLM-ICD.}
\resizebox{2\columnwidth}{!}{%
\begin{tabular}{lcccc}
\toprule
\textbf{Parameter} & \textbf{MIMIC-IV ICD-10 Full} & \textbf{MIMIC-IV ICD-10 Top-50} & \textbf{MIMIC-III ICD-9 Full} & \textbf{MIMIC-III ICD-9 Top-50}\\
\midrule
\multicolumn{5}{c}{\textbf{CAML}}\\
\hdashline
batch size & 8 & 8 & 8 & 8\\
learning rate & 5 × $10^{-3}$ & 5 × $10^{-3}$ & $10^{-4}$ & $10^{-4}$\\
weight decay & $10^{-3}$ & $10^{-3}$ & - & - \\
\hline
\multicolumn{5}{c}{\textbf{LAAT}}\\
\hdashline
batch size & 8 & 8 & 8 & 8\\
learning rate & $10^{-3}$ & $10^{-3}$ & $10^{-3}$ & $10^{-3}$\\
weight decay & $10^{-3}$ & $10^{-3}$ & - & - \\
\hline
\multicolumn{5}{c}{\textbf{PLM-ICD}}\\
\hdashline
batch size & 16 & 16 & 8 & 8\\
learning rate & 5×$10^{-5}$ & 5×$10^{-5}$ & 5×$10^{-5}$ & 5×$10^{-5}$\\
weight decay & 0 & 0 & - & - \\
\bottomrule
\end{tabular}
}
\label{config_ICD}
\end{table*}

\paragraph{On Evaluation} During the faithfulness testing, if the whole document contains fewer than the threshold $N$ tokens, we include all tokens. To evaluate comprehensiveness, if all tokens are removed, the single token with the lowest attention weight is retained.

%The ICD coding performance metrics are reported on both the Full and Top-50 subsets of the MIMIC datasets, where the Top-50 subsets include only the 50 most frequent codes from the respective full datasets.
%
\paragraph{On NER}
We implement the NER models using the default hyperparameter settings provided by \citet{yang2020clinical}, with the parameter configuration summarized in Table \ref{config_NER}. All NER models are trained on 5,000 randomly selected samples from the MIMIC-IV Top-50 dataset. Document-level plausibility evaluation is conducted on 139 annotated documents, filtered from an initial set of 150 samples to retain only those associated with the Top-50 codes. For code-level plausibility evaluation, the test set consists of the remaining samples, excluding those 5 samples used as examples in few-shot prompting. The statistics of the original and test sets are shown in Table \ref{statistics_code_level_test}.

\begin{table}[h!]
\centering
\caption{Configurations of NER models.}
\begin{tabular}{lc}
\toprule
\textbf{Parameter} & \textbf{Value}\\
\midrule
batch size & 8 \\
learning rate & $10^{-5}$\\
warmup ratio & 0.01\\
truncation & 256\\
gradient accumulation steps & 1\\
epoch & 20\\
random seed & 13\\
\bottomrule
\end{tabular}
\label{config_NER}
\end{table}

\begin{table}[h!]
\centering
\caption{Statistics of the test sets used for evaluating code‑level NER models.}
\begin{tabular}{lc}
\toprule
\textbf{Code} & \textbf{Test Set / Full Set}\\
\midrule
I10 &  55 / 60 \\
E785 &  50 / 55\\
Z7901 &  17 / 22\\
I4891 &  11 / 16\\
E119 &  17 / 22\\
\bottomrule
\end{tabular}
\label{statistics_code_level_test}
\end{table}

\section{On LLM-Generated Rationales}

\subsection{Prompting  Without Examples}

The used prompt without examples follows the  format below. In the following, the \textbf{\textit{text}}, \textbf{\textit{code}}, and \textbf{\textit{description of the code}} are  variables that change based on the input note and the target code.

Note Text: \textbf{\textit{text}} + Code: \textbf{\textit{code}}. Description: \textbf{\textit{description of the code}}. Could you please select the spans (rationales) which are related to the code \textbf{\textit{code}}? The spans can be words, phrases, or sentences. Only list the exact spans extracted from the `Note Text', without including their section names. List each span with a number in front. For example: `1. Span1 2. Span2'. Only keep the spans. Do not include any additional responses. Exclude any punctuations at the end of the spans. Keep the spans as what they are in `Note Text'. Keep the spans as what they are in `Note Text'. Keep the spans as what they are in the `Note Text'.

\subsection{Prompting With Few-Shot Examples}

The used prompt with  examples follows the   format below.  In the following,the variable \textbf{\textit{examples}} represents annotation examples corresponding to each \textbf{\textit{code}}. 

Note Text: \textbf{\textit{text}} + Code: \textbf{\textit{code}}. Description: \textbf{\textit{description of the code}}. Could you please select the spans (rationales) which are related to the code \textbf{\textit{code}}? The spans can be words, phrases, or sentences. Only list the exact spans extracted from the `Note Text', without including their section names. For example: \textbf{\textit{examples}}. List each span with a number in front. For example: `1. Span1 2. Span2'. Only keep the spans. Do not include any additional responses. Exclude any punctuations at the end of the spans. Keep the spans as what they are in `Note Text'. Keep the spans as what they are in `Note Text'. Keep the spans as what they are in the `Note Text'.

%Mingyang: Where does these examples appear  in the above prompt?
%examples are variable. It's just a list of vocabularies.

The examples are drawn from five randomly selected documents containing the given code in the annotation dataset. Table \ref{tab:icd_examples} summarizes the samples used for few-shot prompting and the corresponding \textbf{\textit{examples}} value for each code.

\begin{table*}[!htbp]
\centering
\caption{Samples used for few-shot prompting and their annotations.}
\renewcommand{\arraystretch}{1.3}
\resizebox{\textwidth}{!}{%
\begin{tabular}{p{1cm} | p{3cm} | p{2cm} | >{\raggedright\arraybackslash}p{9cm}}
\hline
\textbf{Code} & \textbf{Description} & \textbf{HADM IDs} & \textbf{Annotations} \\
\hline
I10 & Essential (primary) hypertension & 
21893270; 20961577; 20272030; 23048750; 29161744 & 
'HTN', 'Hypertension', 'HYPERTENSION - ESSENTIAL, UNSPEC', 'Essential hypertension', 'hypertension', 'HYPERTENSION' \\
\hline
E785 & Hyperlipidemia, unspecified & 
21893270; 26102343; 20272030; 24014389; 27049443 & 
'HLD', 'Dyslipidemia', 'Hyperlipidemia', 'Hypercholesteremia', 'Dyslipidemia', 'hyperlipidemia' \\
\hline
Z7901 & Long term (current) use of anticoagulants & 
27049443; 29964986; 27021287; 25097869; 29155448 & 
'aspirin', 'Plavix', 'Coumadin', 'Afib on coumadin', 
"and the decision was made to restart the patient's ASA and Plavix immediately postoperatively", 
'Clopidogrel 75 mg PO DAILY', 'Aspirin 325 mg PO DAILY', 'Warfarin', 'on Coumadi', 
'Coumadin who presents with dyspnea', 'on coumadin', 'Warfarin held for supratherapeutic INR (INR 3.4)', 
'afib on warfarin', 'On coumadin.', 'Warfarin 2 mg PO 1X/WEEK', 'Warfarin 5 mg PO 6X/WEEK', 
'Warfarin 2 mg PO 1X/WEEK', 'on high-dose warfarin due to resistance', 'warfarin (5 mg)', 
'heparin gtt', 'Aspirin 81 mg PO DAILY', 'Enoxaparin Sodium 140 mg', 'Aspirin 81 mg PO DAILY' \\
\hline
I4891 & Unspecified atrial fibrillation & 
27049443; 24257587; 29964986; 27466246; 29588477 & 
'atrial fibrillation', 'atrial fibrillation', 'Troponinemia', 'Atrial Fibrillation', 'Digoxin', 'afib', 'Afib on coumadin', 'afib s/p', 'Atrial fibrillation', 'Afib s/p ablation', 'Afib' \\
\hline
E119 & Type 2 diabetes mellitus without complications & 
21893270; 27904530; 24257587; 25097869; 22473872 & 
'diabetes', 'DM', 'Diabetes', 'DM2', 'BLOOD Glucose-113', 
'Additionally, Diabetes service was consulted for newly found hyperglycemia.', 
'insulin dependent DM', 'Diabetes: his hemoglobin A1c was 7.8.', 'type 2 diabetes mellitus' \\
\hline
\end{tabular}
}
\label{tab:icd_examples}
\end{table*}

\subsection{LLM Configurations}
As the PhysioNet Credentialed Data Use Agreement prohibits sharing MIMIC data with external services such as ChatGPT, we follow PhysioNet’s recommendation to use Google Gemini, which does not utilize user prompts or responses for model training. We employ two variants of Gemini 2-Flash and 1.5-Pro. For local deployment, we select LLaMA-3.3, one of the most capable open-weight LLMs available, examining both the 70B Instruct and its quantized variant AWQ. The local LLMs are executed on 4 × and 1 × NVIDIA A100 80GB GPUs, respectively. 

For the Gemini variants, we configured both models with a temperature of 0.1 to minimize randomness and encourage more deterministic outputs, as our task requires selecting spans that exactly match those appearing in the input documents. We also set top\_p to 0.99.

For the LLaMA variants, we adopted the same parameter settings as Gemini. Additionally, we set max\_tokens to 8,000, which exceeds the length of the longest documents in our dataset. This configuration, however, caused the computation for the larger LLaMA model to exceed the default memory capacity of two NVIDIA A100 80GB GPUs.

\subsection{Span Alignment for Rationale Generation }

%LLMs occasionally fail to produce spans that exactly match the original text, despite our explicit instruction—emphasized three times in the prompt—`\textit{Keep the spans as they appear in the ‘Note Text’}.' To address this, we design algorithms to align the generated spans back to the original document and compute overlap scores. Spans with scores greater than 1.7 are retained as the final rationale. %One limitation of this approach is that identical spans appearing at multiple positions in the text are mapped to the same location in the original document, leading to slightly underestimated exact match scores.

LLMs occasionally fail to consistently reproduce spans that exactly match those in the original patient notes, despite our explicit instruction emphasized in the prompt `\textit{Keep the spans as they appear in the ‘Note Text’}.' During the prompt engineering, we observe that repeating this instruction three times improve the consistency to some extent, but the generated spans are still not perfectly aligned. To address this issue and enable accurate evaluation against human annotations afterwords, we developed a post-processing method to map the LLM-generated spans back to the original text. Specifically, we first identify all candidate spans. We narrow the search window within the original text and extract all spans that share the same initial and terminal $n$ characters as the generated rationale. We then compute their overlap scores to select the best match.

\paragraph{Overlap Score Calculation} Let \( S_{\text{g}} \) and \( S_{\text{c}} \) be the generated rationale and candidate target span extracted from the document, respectively.  
\( T_{\text{g}} = \text{Tokenize}(S_{\text{g}}) \), and \( T_{\text{c}} = \text{Tokenize}(S_{\text{t}}) \), where \texttt{Tokenize} returns a set of tokens. The overlap score $S$ is calculated by the following equation:
\setcounter{equation}{2}
\begin{equation}
S = \frac{T_{\text{g}} \cap T_{\text{c}}}{|T_{\text{g}}|} + \frac{T_{\text{g}} \cap T_{\text{c}}}{|T_{\text{c}}|}
\end{equation}
We present in Table \ref{candidates}   an example of overlap score calculation for a set of candidate spans corresponding to the generated span `\textbf{\textit{diagnosis type 2 diabetes}}', which shares the same initial character `d' and terminal character `s' (with a character window size of $n = 1$). The overlap score consists of two components: (1) the overlap ratio with the generated rationale, which captures the degree of alignment with the generated rationale (e.g., `\textit{diagnosis type 2 diabetes}' (1) is preferred over `\textit{diagnosis diabetes}' (0.5)), and (2) the overlap ratio with the candidate span, which reflects the degree of alignment with the candidate span in the original text (e.g., `\textit{diagnosis diabetes}' (1) is preferred over `\textit{diagnosis tuberculosis}' (0.25)).

\setcounter{table}{4}
\begin{table}[h!]
\centering
\caption{Candidates targets of a generated rationale \textbf{\textcolor{cyan}{d}iagnosis type 2 diabete\textcolor{red}{s}}. The decomposition consists of the two components of the overlap score function.}
\resizebox{1\columnwidth}{!}{%
\begin{tabular}{lrr}
\toprule
\textbf{Candidate} & \textbf{Decomposition} & \textbf{Overlap Score} \\
\midrule
\textcolor{cyan}{d}iagnosi\textcolor{red}{s} & 0.25 + 1 & 1.25 \\
\textcolor{cyan}{d}iabete\textcolor{red}{s} & 0.25 + 1 & 1.25 \\
\textcolor{cyan}{d}iagnosi\textcolor{red}{s} tuberculosis & 0.25 + 0.5 & 0.75 \\
\textcolor{cyan}{d}yslipidemia\textcolor{red}{s} & 0 + 0 & 0 \\
\textcolor{cyan}{d}iagnosis diabete\textcolor{red}{s} & 0.5 + 1 & 1.5 \\
\textcolor{cyan}{d}iagnosis type 2 diabete\textcolor{red}{s} & 1 + 1 & 2 \\
\bottomrule
\end{tabular}
}
\label{candidates}
\end{table}

\paragraph{Rationale Selection} We repeat this process across a range of character window sizes and select the candidate with the highest overlap score as the final mapping result. This procedure is applied to all generated spans, and those with an overlap score exceeding 1.7 are retained as supporting rationale for the document. The detailed steps of this mapping process are provided in the accompanying pseudocodes. Statistics for the Gemini 2-Flash–generated dataset across different span alignments, covering 122,004 samples from MIMIC-IV, are provided in Appendix J. Additional results on the plausibility of Gemini 2-Flash–generated rationales across different span alignments are presented in Appendix K. The impact of label selection under different span alignments during rationale learning is presented in Appendix L.

\begin{algorithm*}[!htbp]
\caption{\texttt{OverlapScore}: Compute Overlap Score Between Two Spans}
\label{alg:overlapscore}
\textbf{Input}: Generated span $generated\_span$, Candidate span $candidate\_span$ \\
\textbf{Output}: Overlap score between the two spans
\begin{algorithmic}[1] %[1] enables line numbers
\STATE $T_g \gets \texttt{Tokenize}(generated\_span)$
\STATE $T_t \gets \texttt{Tokenize}(candidate\_span)$
\STATE $I \gets T_g \cap T_t$
\STATE $score \gets \frac{|I|}{|T_g|} + \frac{|I|}{|T_t|}$
\STATE \textbf{return} $score$
\STATE \textit{Notes: \texttt{Tokenize} function splits a text span into individual tokens.}
\end{algorithmic}
\end{algorithm*}

\begin{algorithm*}[!htbp]
\caption{\texttt{BestCandidate}: Find Best Matching Candidate Span}
\label{alg:best_candidate}
\textbf{Input}: Generated span $generated\_span$, Document $note$ \\
\textbf{Output}: Best candidate and its overlap score
\begin{algorithmic}[1]
\STATE $best\_candidate$ $\gets$ \texttt{''}
\STATE $max\_score$ $\gets 0$
\STATE $candidates$ $\gets$ [ ]
\FOR{$n = 7$ \textbf{to} $1$ \textbf{step} $-1$}
    \STATE $span\_start$ $\gets$ \texttt{Lower}($generated\_span[:$n$]$)
    \STATE $span\_end$ $\gets$ \texttt{Lower}($generated\_span[-$n$ :]$)
    \FOR{$i = 0$ \textbf{to} \texttt{Len}($note$) $- n$}
        \STATE $window\_start$ $\gets$ \texttt{Lower}($note[$i$:$i+n$]$)
        \IF{$window\_start$ == $span\_start$}
            \STATE $start\_index$ $\gets i$
            \FOR{$j = 0$ \textbf{to} \texttt{Len}($note$) $-$ $start\_index$ $-$ $n$}
                \STATE $end\_index$ $\gets$ \texttt{Len}($note$) $- j$
                \IF{$end\_index$ $- n < 0$}
                    \STATE \textbf{break}
                \ENDIF
                \STATE $window\_end$ $\gets$ \texttt{Lower}(n$ote[end\_index -n : end\_index]$)
                \IF{$window\_end$ == $span\_end$ \textbf{and} $end\_index$ $>$ $start\_index$}
                    \STATE $text\_candidate$ $\gets$ $note[start\_index:end\_index]$
                    \STATE $candidates$ $\gets$
                    $candidates$ $\cup$ $text\_candidate$
                \ENDIF
            \ENDFOR
        \ENDIF
    \ENDFOR
\ENDFOR
\FOR{$item$ \textbf{in} $candidates$}
    \STATE $score$ $\gets$ \texttt{OverlapScore}($generated\_span$, $item$)
    \IF{$score$ $>$ $max\_score$}
        \STATE $max\_score$ $\gets$ $score$
        \STATE $best\_candidate$ $\gets$ $item$
    \ENDIF
\ENDFOR
\STATE \textbf{return} $best\_candidate$, $max\_score$
\STATE \textit{Notes: \texttt{Lower} denotes the lowercase conversion function; \texttt{Len} function returns the length of a string.}
\end{algorithmic}
\end{algorithm*}

\section{More Details on Dataset Construction}

\paragraph{Data Selection}

To build a comprehensive comparison that includes Entity Linking data, we first identify the overlap between the Entity Linking and the MIMIC-IV ICD-10 datasets. There are 64 overlapping samples between the two datasets. We then randomly select an additional 86 samples from the MIMIC-IV ICD-10 test set, resulting in a final dataset of 150 samples.

\paragraph{Annotator}

Two annotators with medical background contributed to this annotation work.  Annotator 1 holds a bachelor’s degree in medicine, and Annotator 2 holds a master’s degree in medicine. Annotator 1 annotated all samples, while Annotator 2 performed a quality check by annotating a subset of codes in 13 samples.

\paragraph{Payment}

The two annotators were paid £10 per document and an additional £15 for setting up the annotation platform.

\paragraph{Annotation Guidelines}

In this subsection, we outline the guidelines developed for the annotators.

\textit{Title.} Identifying Rationales of ICD Codes in Discharge Summaries

\textit{Purpose.}
Explainability is especially critical in the clinical domain, where transparent and well-supported rationales enable healthcare providers and decision-makers to confidently utilize model predictions in patient care. The data you annotate will serve as a gold standard for evaluating the explainability of ICD coding models. By comparing rationales identified by these models with those provided by human annotators, we aim to assess how effectively the models present rationales, particularly in a human-understandable manner.

\textit{Task Description.}
In this task, you will be presented with a patient’s discharge summary and its assigned ICD (International Classification of Diseases) codes. Your goal is to identify and highlight text spans that support the given ICD codes as rationales. These highlighted spans may be words, phrases, or sentences (complete or incomplete).

\textit{Platform Setup.}
\begin{itemize}
    \item Install Docker following the official installation guide.
    \item Set up the annotation platform (Doccano).
\end{itemize}

\textit{Annotation Process.}
\begin{itemize}
    \item On the Doccano interface, you will see a patient’s discharge summary along with its assigned ICD-10 codes and their descriptions.
    \item As you review the summary, highlight all text spans that you believe support each label (ICD code).
    \item When you select a span with your mouse, a selection list will appear. Click the appropriate label to annotate the span with it. The same text span can be annotated with multiple labels.
\end{itemize}

\textit{Completing the Annotation.}
\begin{itemize}
    \item When you would like to save your current annotations, click the ‘X’ mark on the top left corner to change the status from ‘Not Checked’ to ‘Checked’.
    \item Once you have finished annotating all the labels for a summary, the sample will be marked as ‘Finished’.
\end{itemize}

\paragraph{Platform}

We conducted the annotation using Doccano, a free and open-source platform designed to facilitate the creation of labeled datasets for natural language processing tasks. Doccano supports various annotation types, including text classification, sequence labeling (e.g., named entity recognition), and sequence-to-sequence tasks (e.g., machine translation or summarization). For this study, we employ the sequence labeling functionality. In the setup, the `Allow overlapping spans' and `Share annotations across all users' options are enabled. Additionally, a separate project is created for each document, each with its own defined label set.

\paragraph{Inter-Annotator Agreement}

Inter-annotator agreement (IAA) refers to the degree of consistency or reliability among different human annotators who independently annotate the same dataset. It is a critical measure to assess the quality and objectivity of annotated data, especially in tasks involving subjective judgments. We evaluate the annotation quality through a secondary annotator, who annotate the rationales for a subset of codes across 13 samples. To assess agreement, we calculate both span-level and token-level matching scores, including Precision, Recall and F1.

Here we explain how these scores are computed. Let $A_2$ be the set of tokens annotated by Annotator 2 and $A_1$ be the set of tokens annotated by Annotator 1. The overlap between these sets is:
\begin{equation}
    \mathrm{Overlap} = A_2 \cap A_1.
\end{equation}
Precision, recall, and F1 scores are computed as follows:
\begin{equation}
\mathrm{Precision} = \frac{|A_2 \cap A_1|}{|A_1|},
\end{equation}
\begin{equation}
\mathrm{Recall} = \frac{|A_2 \cap A_1|}{|A_2|},
\end{equation}
\begin{equation}
\mathrm{F1} = 2 \times \frac{\mathrm{Precision} \times \mathrm{Recall}}{\mathrm{Precision} + \mathrm{Recall}}.
\end{equation}

Table~\ref{IAA} presents the results of the inter-annotator agreement analysis. The token-level F1-score (53.32) is substantially higher than the span-level F1-score (31.58), indicating that annotators exhibit greater consistency in identifying relevant content than in determining precise span boundaries. Table~\ref{annotations_IAA} provides a subset of detailed annotation matches between the two annotators.

\begin{table}[h!]
\centering
\caption{Inter-Annotator Agreement}
\begin{tabular}{lr}
\toprule
\textbf{Metric} & \textbf{Value (\%)} \\
\midrule
Span-Level Precision & 38.00 \\
Span-Level Recall & 30.95 \\
Span-Level F1 & 31.58 \\
Token-Level Precision & 79.84 \\
Token-Level Recall & 44.16 \\
Token-Level F1 & 53.32 \\
\bottomrule
\end{tabular}
\label{IAA}
\end{table}

\begin{table*}[h!]
\centering
\caption{Part of annotation details of two annotators.}
\resizebox{0.95\textwidth}{!}{
\begin{tabular}{lllll}
\toprule
\textbf{HADM ID} & \textbf{Code} &	\textbf{Annotator2} & \textbf{Annotator1} &	\textbf{Match}\\
\midrule
29531980	&	027034Z	&	`Left heart cath'	&	`nan'	&	False	\\
29531980	&	I10	&	`hypertension'	&	`hypertension'	&	True	\\
29531980	&	E1140	&	`type 2 diabetes'	&	`type 2 diabetes'	&	True	\\
29531980	&	N390	&	`MRSA UTI'	&	`MRSA UTI '	&	False	\\
29531980	&	E1140	&	`Diabetes'	&	`Diabetes'	&	True	\\
29531980	&	I10	&	`Hypertension'	&	`Hypertension'	&	True	\\
29531980	&	I10	&	`Hypertension'	&	`Hypertension'	&	True	\\
29531980	&	E1140	&	`Diabetes'	&	`Diabetes'	&	True	\\
29531980	&	I10	&	`hypertension'	&	`hypertension'	&	True	\\
29474957	&	W363XXA	&	`Trauma'	&	`nan'	&	False	\\
29474957	&	W363XXA	&	`patient reports that he was blown off 	&	`he was blown off 	&		\\

	&		&	by the lid of a 
highly pressurized natural 	&	by the lid of a 
highly pressurized natural 	&		\\

	&		&	gas tank in his pickup truck and 
sustained 	&	gas tank in his pickup truck and 
sustained 	&		\\
	&		&	multiple injuries from the blast'	&	multiple injuries from the blast'	&	False	\\

29474957	&	H05221	&	`Right 
periorbital soft tissue swelling '	&	`Right 
periorbital soft tissue swelling '	&	True	\\
29474957	&	S2241XA	&	`Right-sided rib fractures'	&	`rib 
fractures '	&	False	\\

29474957	&	S2241XA	&	`Right-sided rib fractures involving 	&	`Right-sided rib fractures involving 	&		\\
	&		&	the eighth and ninth ribs 
which appear 	&	the eighth and ninth ribs 
which appear 	&		\\
	&		&	comminuted displaced as well as fractured right 
	&	comminuted displaced as well as fractured right 
	&		\\
	&		&	tenth rib at the  costovertebral junction'	&	tenth rib at the  costovertebral junction. '	&	False	\\
\bottomrule
\end{tabular}
}
\label{annotations_IAA}
\end{table*}

\paragraph{ICD Codes with no supporting evidence}

During the annotation, we observe that not all codes have their supporting rationales in the text. It is mostly due to initial coding errors made by human annotators during the construction of the MIMIC benchmark. We list all codes with no rationale in Table \ref{tab:no_rationale}, which also serves the analysis of the quality of the MIMIC-IV ICD10 dataset.

\begin{table*}[!htbp]
\centering
\caption{Codes without supporting rationales in the document.}
\resizebox{0.95\textwidth}{!}{
\begin{tabular}{ll}
\hline
\textbf{HADM ID} & \textbf{ICD-10 Code}\\
\hline
21843396 & J45909;02HV33Z\\
29964986 & 30283B1;Z87891\\
29918504 & F17210\\
29677969 & Z96651;Z22322\\
28100046 & J9811;J95811;Z23\\
27638102 & F17290\\
27044834 & E860\\
26620438 & E8342\\
25912628 & Z87891\\
24823574 & Z87891\\
23856554 & Z87891\\
23702445 & F329\\
23355051 & Z87891\\
23295582 & Z87891;F419\\
29588477 & 02HV33Z\\
22893898 & E874;J984;E870;H40052;Y848;Y92230;I9581;05UL0KZ;0NS60ZZ;08N1XZZ;02HV33Z\\
23051773 & I25118;D62;D696;R350;5A1221Z\\
23618067 & Z23\\
27840655 & I5033;I110\\
22528733 & N814\\
23106502 & I082;I671;H409;R6884;Z87891;Y92009\\
25920183 & A549;02HV33Z\\
26094695 & Z66;R627;Z6821\\
27356906 & J9600;Z781;K7290;Z87891;06L34CZ\\
21222006 & Z22321\\
21318772 & I2109;Z23;E861;R740;B1920\\
21386441 & F17210\\
22257486 & G4733\\
22733522 & G4733;E669;Z6833;Z87891\\
23354056 & D696;E860\\
23694175 & K648;R29700\\
24345583 & F17210\\
24852593 & F0390;Z86711\\
25307585 & E881;A630;L732;Z87891\\
25334768 & A419;E46;I272\\
25510774 & F1021\\
26911900 & Z9119\\
27567712 & T859XXA;T81.4XXA;Y92129;Z85820;0FPGX0Z;0F2GX0Z;02HV33Z;0JD80ZZ;0DHA8UZ;3E0H76Z\\
28716988 & Z781\\
28831703 & T8172XA;I808\\
29520101 & Z87891\\
22032290 & K55029;R6521;D62;F05;Z66;F17210\\
22114206 & Z87891\\
22556702 & Z006;K219\\
22983901 & D6832;D6959;Z87891;T45515A\\
23383624 & M25561;Z87820\\
23869666 & E46;Z6829\\
24309140 & B9562;I452;Z8673;Z86711;F17210;0W383ZZ;02H633Z\\
24352758 & D684;N179;Z87891;D6959;Z781;0BC68ZZ;0BC48ZZ;3E0G76Z\\
24378932 & N179;N182\\
24672299 & Z87891\\
24974242 & F17210\\
24991332 & I871;05JY3ZZ\\
26852604 & Z87891\\
27705753 & Z87891;0U20KZ\\
28206965 & I509;F05;BT11YZZ\\
28731328 & Z87891;R0682;F40240\\
28756843 & I959;Z87891\\
29060004 & Z87891\\
29402217 & Z22322\\
\hline
\end{tabular}
}
\label{tab:no_rationale}
\end{table*}

\paragraph{Details of Data Processing}

When evaluating plausibility, the original rationale annotations are preprocessed to align with the input formats of ICD coding models. We apply the same preprocessing procedures used in \citet{edin2023automated}’s medical coding reproducibility study. Specifically, CAML and LAAT annotations are cleaned by lowercasing, removing special characters and stray numbers, and trimming extra spaces. For PLM‑ICD, this cleaning step is followed by tokenization using the RoBERTa‑base‑PM‑M3‑Voc‑distill‑align‑hf tokenizer from Hugging Face. These procedures ensure that the rationales extracted by each model are accurately comparable to our processed gold-standard annotations.
\section{Code Overlap Analysis}

MDACE annotates a subset of MIMIC-III clinical notes with ICD-10 codes independently, then identifies their corresponding rationales. These coding results are then automatically mapped to ICD-9 codes using the General Equivalence Mappings (GEMs). This workflow introduces inconsistencies with the original ICD-9 code distribution in the MIMIC-III dataset commonly used for ICD coding tasks. We analyze the overlap between the two datasets under both the \textit{Full} code setting and the \textit{Top-50} code setting, with average overlaps of 37.00\% and 14.59\%, respectively. Figure \ref{overlap} presents the distribution of overlap ratios in terms of number of documents and cumulative proportion. The results indicate that, for approximately 80\% of the documents, the code overlap is below 60\%. In the Top-50 setting, none of the documents exhibit more than 60\% overlap. Figure \ref{frequency} presents the frequency distribution of the Top-50 ICD-9 codes in both MDACE and the original MIMIC-III ICD-9 dataset. Notably, 6 of the Top-50 codes are entirely missing in MDACE (and 725 codes are missing in the Full-code setting, which comprises 1,281 codes in total). This substantial inconsistency poses challenges in accurately evaluating the explainability of ICD coding models.

In contrast, the annotation workflow of our dataset, RD-IV-10, is conducted by providing annotators with the discharge summaries and their corresponding labels from the MIMIC-IV ICD-10 dataset. The code distribution of our dataset closely matches that of MIMIC-IV ICD-10. As shown in Figure \ref{overlap_RD}, most samples exhibit nearly 100\% overlap, indicating a high consistency with the MIMIC-IV ICD-10 distribution. Specifically, the average overlaps reach 93.15\% and 83.88\% under the Full and Top-50 settings, respectively. Figure \ref{frequency_RD} further demonstrates the consistency in distribution. The reason they do not reach 100\% is that both annotators observed that not all labels have supporting rationales in the text. Details of the codes without supporting rationales are provided in Appendix C.

Additionally, we investigate code switching by evaluating PLM-ICD on both the MDACE dataset and a filtered MIMIC-III ICD-9 test set sharing the same HADM IDs. MDACE samples come from two categories: Inpatient and Profee. The Inpatient category includes 302 samples, while Profee contains 52 samples corresponding to the same discharge summary but assigned with different codes. We combined the codes for samples sharing the same HADM ID. For comparison, we filtered the original MIMIC-III ICD-9 test set to include only samples with HADM IDs matching those in MDACE. We trained PLM-ICD on an extended label space that includes the 40 new ICD-9 codes introduced by MDACE. The testing results differ substantially from the original test set, showing a Precision@8 of 55.34\% on MDACE versus 78.02\% on the filtered MIMIC-III ICD-9 set, as shown in Table~\ref{testing_two_sets}.

\begin{table*}[h!]
\centering
\caption{Results on MDACE and filtered MIMIC-III ICD-9 test set (same HADM IDs).}
\resizebox{2\columnwidth}{!}{%
\begin{tabular}{lccccc}
\hline
\textbf{Test Set} & \textbf{F1 Macro} & \textbf{F1 Micro} & \textbf{AUC Macro} & \textbf{AUC Micro} & \textbf{Precision@8} \\
\hline
MDACE & 3.39 & 50.38 & 90.82 & 96.95 & 55.34 \\
Filtered MIMIC-III ICD-9 Test Set & 5.01 & 59.74 & 94.78 & 99.01 & 78.02 \\
\hline
\end{tabular}
}
\label{testing_two_sets}
\end{table*}

\setcounter{figure}{5}
\begin{figure}[h!]
\centering
\includegraphics[width=0.47\textwidth]{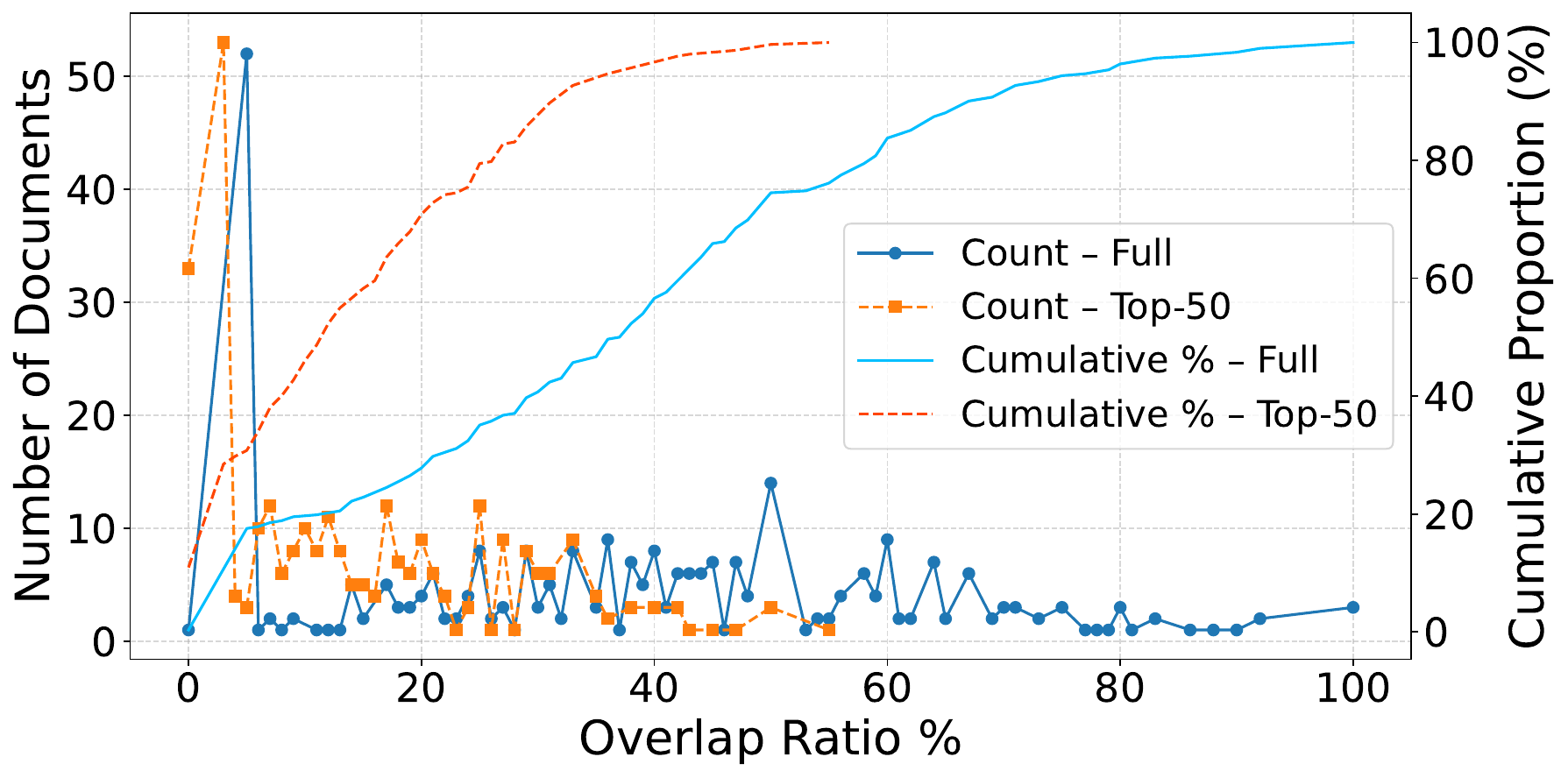} % Reduce the figure size so that it is slightly narrower than the column.
\caption{The statistics of overlap between MIMIC-III ICD-9 code set and mapped ICD-9 code set in MDACE.}
\label{overlap}
\end{figure}
\setcounter{figure}{5}
\begin{figure}[h!]
\centering
\includegraphics[width=0.47\textwidth]{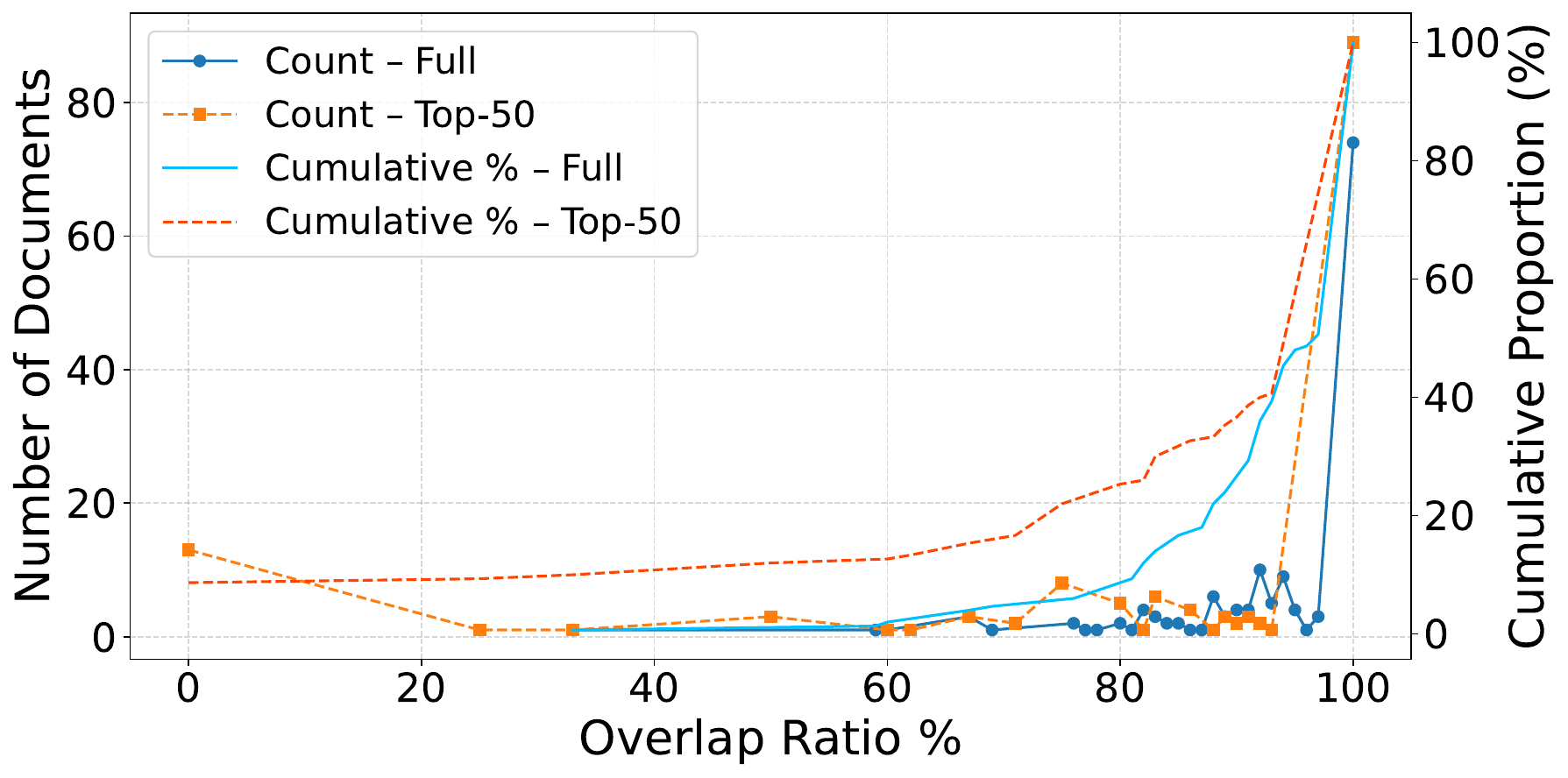} % Reduce the figure size so that it is slightly narrower than the column.
\caption{The statistics of overlap between MIMIC-IV ICD-10 code set and ICD-10 code set in RD-IV-10.}
\label{overlap_RD}
\end{figure}

\begin{figure}[h!]
\centering
\includegraphics[width=0.47\textwidth]{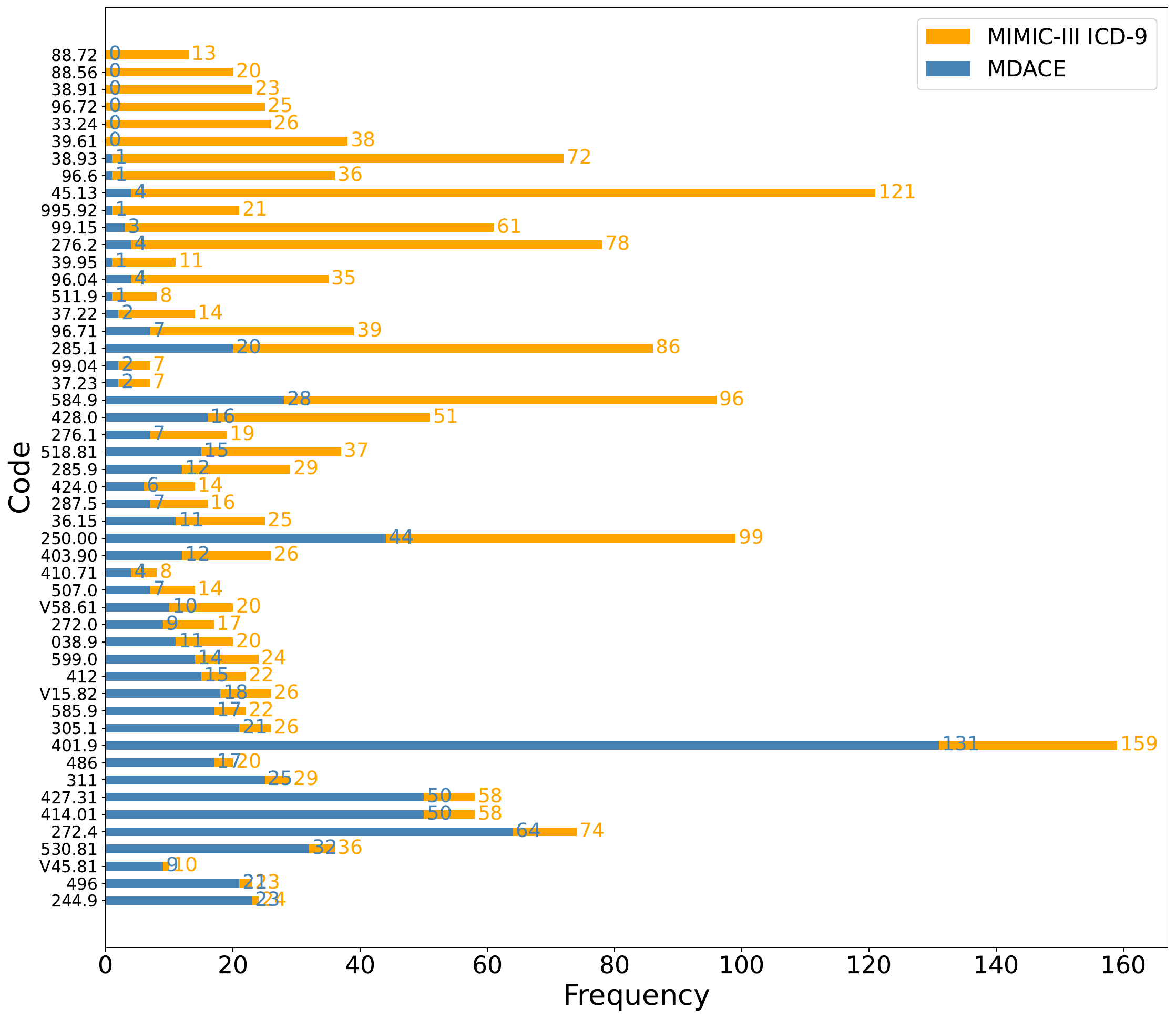} % Reduce the figure size so that it is slightly narrower than the column.
\caption{Code distributions in MDACE and MIMIC-III ICD-9. The analysis is based on the Top-50 codes in MIMIC-III ICD-9.}
\label{frequency}
\end{figure}
\begin{figure}[h!]
\centering
\includegraphics[width=0.47\textwidth]{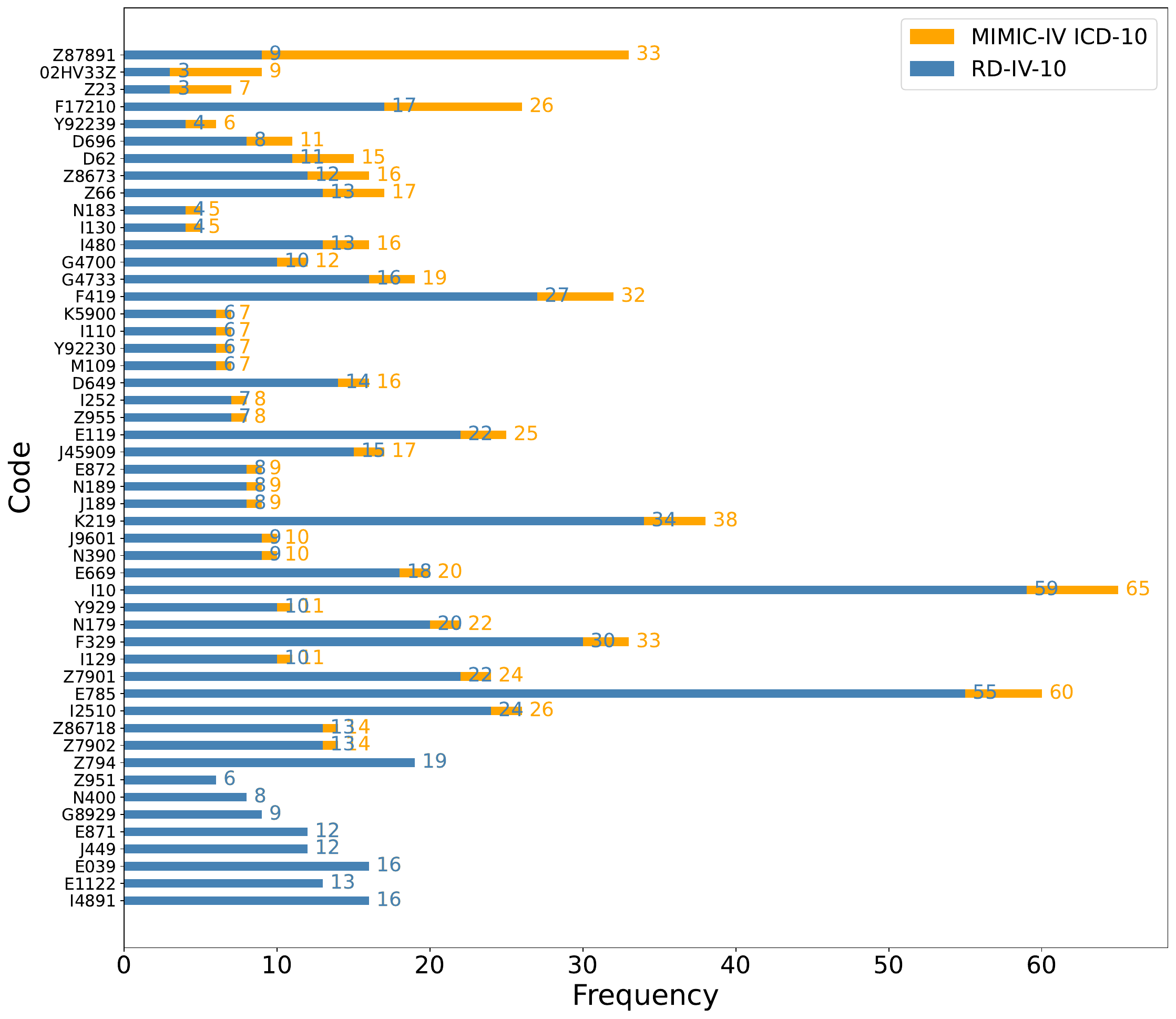} % Reduce the figure size so that it is slightly narrower than the column.
\caption{Code distributions in RD-IV-10 and MIMIC-IV ICD-10. The analysis is based on the Top-50 codes in MIMIC-IV ICD-10.}
\label{frequency_RD}
\end{figure}
\section{A Case Study: Comparing Annotation Quality in RD-IV-10 and MDACE}

\begin{figure*}[!htbp]
\centering
\includegraphics[width=0.95\textwidth]{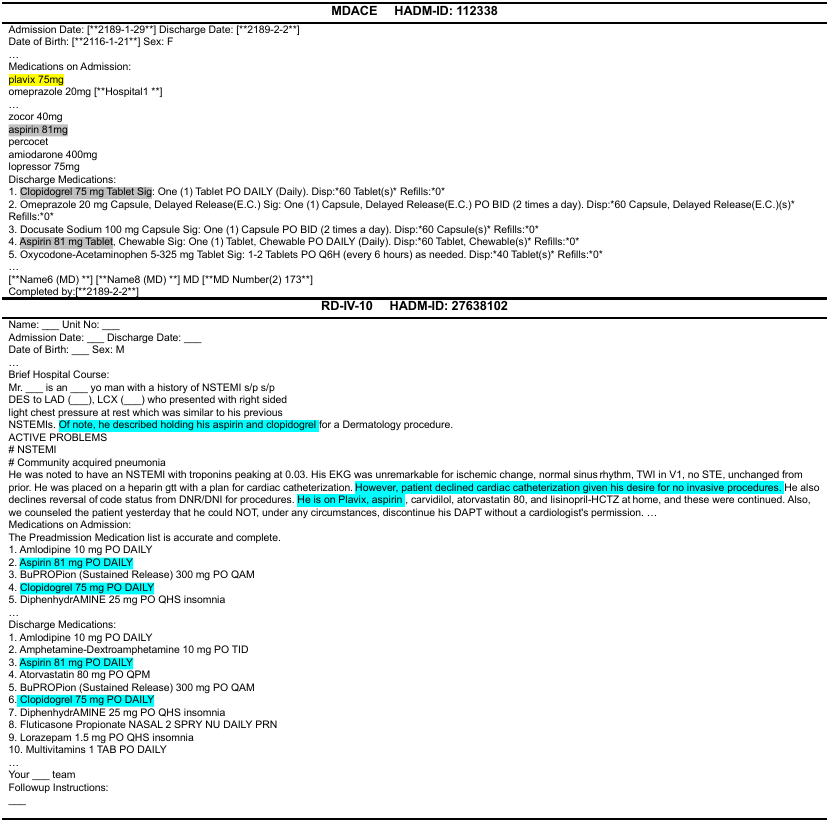} % Reduce the figure size so that it is slightly narrower than the column.
\caption{A case study of the annotation quality of RD-IV-10 and MDACE.}
\label{case_quality_two_datasets}
\end{figure*}

Table 1 indicates that, on average, each ICD code in MDACE is associated with roughly one supporting rationale, given the ratio of 13.83 evidence spans to 11.53 labels per document. In contrast, RD-IV-10 provides substantially richer annotations, averaging 35.94 rationales for 14.82 labels per document. To illustrate this discrepancy, we present a case study shown in Figure \ref{case_quality_two_datasets} comparing both datasets. We randomly select one document annotated with the ICD‑10 code \textbf{Z79.02 - Long‑term use of antithrombotics/antiplatelets} from two datasets. In MDACE, the sole highlighted rationale is `\textit{plavix 75 mg}'. RD-IV-10, however, highlights not only `\textit{Plavix}' but also other directly relevant medications such as `\textit{aspirin}' and `\textit{clopidogrel}', found in multiple locations that MDACE fails to capture (highlighted in grey in the visualization). Additionally, our dataset annotates indirect supportive evidence `\textit{However, patient declined cardiac catheterization given his desire for no invasive procedures}'. This statement reinforces the patient’s complex cardiac history and justifies ongoing antiplatelet therapy, despite declining further interventions.

\section{Complete Results of Faithfulness and Plausibility of CAML, LAAT and PLM-ICD}

\paragraph{Results of Faithfulness}

%Tables \ref{tab:faithfulness_caml}, \ref{tab:faithfulness_laat}, and \ref{tab:faithfulness_plmicd} 
Tables 14, 15, and 16 summarize the faithfulness results of CAML, LAAT, and PLM‑ICD, respectively, using two rationale selection strategies across four MIMIC datasets.

\paragraph{Results of Plausibility}

Tables \ref{tab:plausibility_caml}, \ref{tab:plausibility_laat}, and \ref{tab:plausibility_plmicd} summarize the plausibility results of CAML, LAAT, and PLM‑ICD, respectively, using two rationale selection strategies across all threshold settings.

\paragraph{Model-generated Rationales: \textit{Do Tokens with Higher Attention Weights Make Sense?}}

\begin{figure}[t]
\centering
\includegraphics[width=0.48\textwidth]{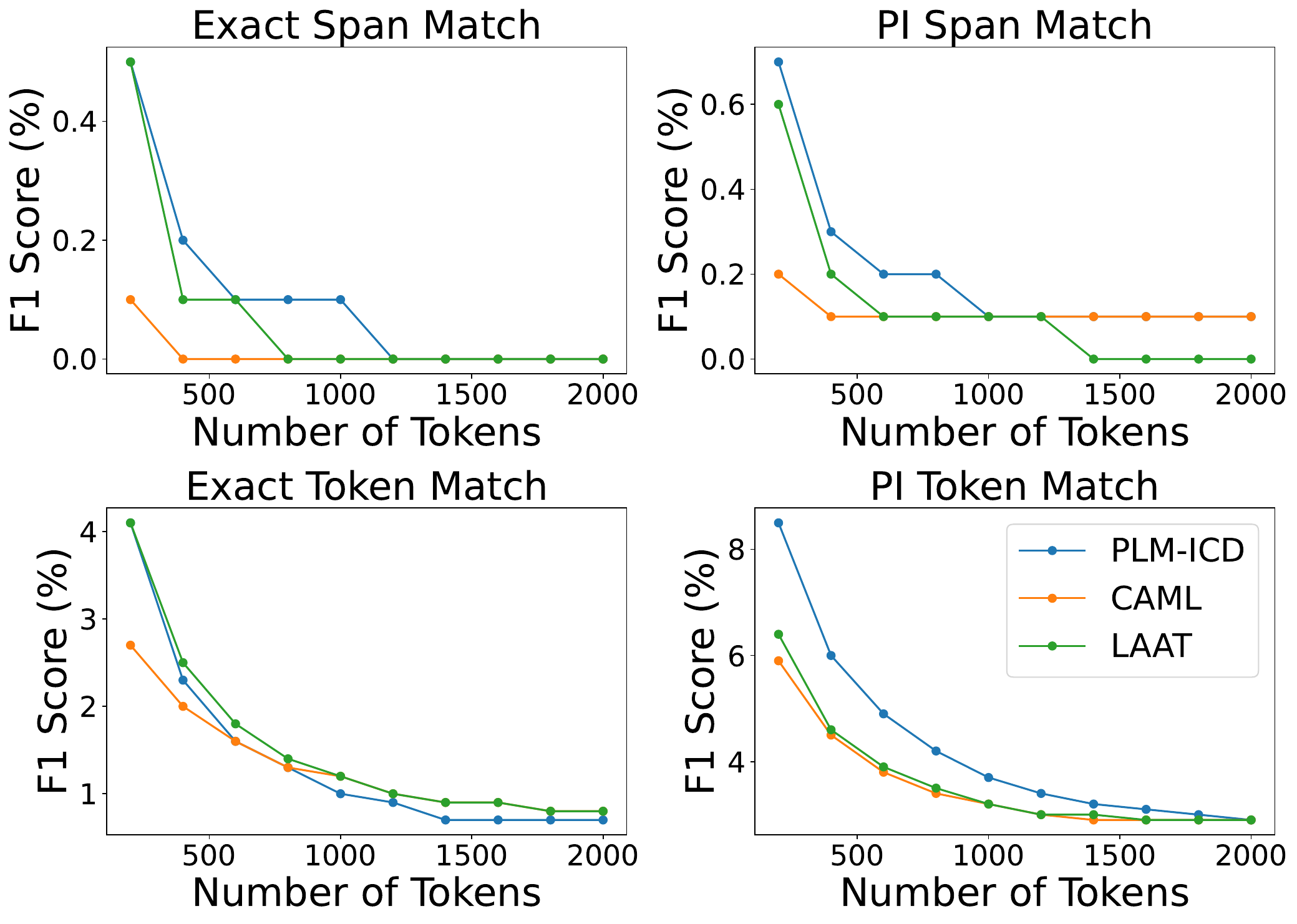} % Reduce the figure size so that it is slightly narrower than the column.
\caption{Plausibility results of ICD coding models across all thresholds settings.}
\label{fig:plausibility_icd}
\end{figure}

Figure \ref{fig:plausibility_icd} reports the matching results of \textit{Top N tokens} across all thresholds. Overall, the performances of the three models are comparable. Specifically, PLM-ICD demonstrates better plausibility than LAAT, which in turn outperforms CAML. Span-level matches remain close to zero across all thresholds. Models achieve higher scores at lower thresholds, because shorter spans are selected, which better align with human annotations. For token-level matches, the absolute number of true positives increases at higher thresholds due to the inclusion of more tokens. However, the F1 scores decline because the total number of predicted tokens increases more substantially than the number of true positives. In conclusion, the rationales generated by ICD coding models do not align with human explanations.

\section{Complete Results of Plausibility}

\paragraph{Plausibility Results of Three Types of Rationales}
Table \ref{tab:plausibility} presents the plausibility results of naive entity-level rationales, strong LLM-generated rationales, model-generated rationales.

\begin{table*}[!htbp]
\centering
\caption{Plausibility results of \textcolor{orange}{\uline{naive entity-level rationales}}, \textcolor{blue!60!black}{\dashuline{strong LLM-generated rationales}}, \textcolor{green!30!black}{\uuline{model-generated rationales}}. Prediction refers to the number of spans or tokens generated by the model, while Accurate denotes the number of spans or tokens matching the human-annotated gold standard. This evaluation is based on 64 documents to enable a comparison of entity linking.}
\resizebox{\textwidth}{!}{
\begin{tabular}{cccccccccc}
\hline
\textbf{Metric} & \textbf{Model/Dataset}  & \textbf{Prdiction} & \textbf{Accurate} & \textbf{TP} & \textbf{FP} & \textbf{FN} & \textbf{Precision (\%)} & \textbf{Recall (\%)} & \textbf{F1 (\%)} \\
\hline
\multirow{7}{*}{Exact SM} & \textcolor{orange}{\uline{Entity-Linking}} & 3546 & 2260 & 298 & 3248  & 1962 & 8.4 & 13.2 & 10.3 \\
& \textcolor{blue!60!black}{\dashuline{Gemini 2-Flash}} & 4726 & 2260 & 754 & 3972 & 1506 & 16.0 & 33.4 & 21.6  \\
& \textcolor{blue!60!black}{\dashuline{Gemini1.5-Pro}} & 11184 & 2260 & 907 & 10277 & 1353 & 8.1 & 40.1 & 13.5 \\
& \textcolor{blue!60!black}{\dashuline{LLaMA-3.3 Ins}} & 5789 & 2260 & 747 & 5042 & 1513 & 12.9 & 33.1 & 18.6 \\
& \textcolor{blue!60!black}{\dashuline{LLaMA-3.3 AWQ}} &  6428  & 2260 & 761 & 5667 & 1499  & 11.8 &33.7 &17.5\\
& \textcolor{green!30!black}{\uuline{CAML}}  & 84520 & 2269 & 55 & 84465 & 2214 & 0.1 & 2.4 & 0.1 \\
& \textcolor{green!30!black}{\uuline{LAAT}}  & 91482 & 2269 & 330 & 91152 & 1939 & 0.4 & 14.5 & 0.7 \\
& \textcolor{green!30!black}{\uuline{PLM-ICD}}  & 65639 & 2269 & 172 & 65467 & 2097 & 0.3 & 7.6 & 0.5 \\
\hline
\multirow{7}{*}{PI SM} & \textcolor{orange}{\uline{Entity-Linking}} & 2751 & 1762 & 207 & 2544  & 1555 & 7.5 & 11.7  & 9.2 \\
& \textcolor{blue!60!black}{\dashuline{Gemini-2flash}} & 4664 & 1762 & 773 &  3891 & 989 & 16.6  & 43.9  & 24.1 \\
& \textcolor{blue!60!black}{\dashuline{Gemini-1.5pro}} & 10996 & 1762 &  932 & 10064 &  830 & 8.5 & 52.9 & 14.6 \\
& \textcolor{blue!60!black}{\dashuline{LLaMA-3.3 Ins}} & 5726 &  1762 &  805  &  4921 & 957 & 14.1 & 45.7 & 21.5  \\
& \textcolor{blue!60!black}{\dashuline{LLaMA-3.3 AWQ}} &  6364 & 1762 & 816 & 5548 & 946  &12.8  & 46.3 & 20.1 \\ 
& \textcolor{green!30!black}{\uuline{CAML}}  & 77970 & 2021 & 88 & 77882 & 1933 & 0.1 & 4.4 & 0.2 \\
& \textcolor{green!30!black}{\uuline{LAAT}}  & 82542 & 2021 & 342 & 82200 & 1679 & 0.4 & 16.9 & 0.8 \\
& \textcolor{green!30!black}{\uuline{PLM-ICD}}  & 60851 & 2041 & 228 & 60623 & 1813 & 0.4 & 11.2 & 0.7 \\
\hline
\multirow{7}{*}{Exact TM} & \textcolor{orange}{\uline{Entity-Linking}} &  5629 & 11422 & 540 & 5089 & 10882 & 9.6 & 4.7 & 6.3 \\
& \textcolor{blue!60!black}{\dashuline{Gemini-2flash}} & 27639 & 11422 & 5881 & 21758 &  5541  & 21.3  & 51.5  & 30.1 \\
& \textcolor{blue!60!black}{\dashuline{Gemini-1.5pro}} & 57708 & 11422 &  7109 &  50599 &  4313 &  12.3 &  62.2  &  20.6  \\
& \textcolor{blue!60!black}{\dashuline{LLaMA-3.3 Ins}} & 28800 & 11422 & 5588 & 23212 & 5834 & 19.4  & 48.9  & 27.8  \\
& \textcolor{blue!60!black}{\dashuline{LLaMA-3.3 AWQ}} & 32693 & 11422 & 6008 & 26685 & 5414 &18.4  & 52.6 & 27.2 \\ 
& \textcolor{green!30!black}{\uuline{CAML}}  & 150968 & 11428 & 2493 & 148475 & 8935 & 1.7 & 21.8 & 3.1 \\
& \textcolor{green!30!black}{\uuline{LAAT}}  & 160732 & 11428 & 4266 & 156466 & 7162 & 2.7 & 37.3 & 5.0 \\
& \textcolor{green!30!black}{\uuline{PLM-ICD}}  & 173633 & 12517 & 3975 & 169658 & 8542 & 2.3 & 31.8 & 4.3 \\
\hline
\multirow{7}{*}{PI TM} & \textcolor{orange}{\uline{Entity-Linking}} & 4292 & 8160 & 381 & 3911  & 7779 &  8.9 & 4.7 & 6.1 \\
& \textcolor{blue!60!black}{\dashuline{Gemini-2flash}} & 18935 &  8160 & 5047  & 13888 &  3113 & 26.7 & 61.9  & 37.3 \\
& \textcolor{blue!60!black}{\dashuline{Gemini-1.5pro}} & 37197 & 8160 & 5898 & 31299 & 2262 & 15.9 & 72.3  & 26.0  \\
& \textcolor{blue!60!black}{\dashuline{LLaMA-3.3 Ins}} & 20708 &  8160 & 5056  & 15652 & 3104 & 24.4 & 62.0  & 35.0  \\
& \textcolor{blue!60!black}{\dashuline{LLaMA-3.3 AWQ}} & 23302  &  8160  &   5361 &  17941  &  2799 &  23.0  & 65.7  &  34.1 \\
& \textcolor{green!30!black}{\uuline{CAML}} & 114766 & 9367 & 4029 & 110737 & 5338 & 3.5 & 43.0 & 6.5 \\
& \textcolor{green!30!black}{\uuline{LAAT}}  & 121292 & 9367 & 4720 & 116572 & 4647 & 3.9 & 50.4 & 7.2 \\
& \textcolor{green!30!black}{\uuline{PLM-ICD}} & 120777 & 10269 & 5794 & 114983 & 4475 & 4.8 & 56.4 & 8.8 \\
\hline
\end{tabular}
}
\label{tab:plausibility}
\end{table*}

\paragraph{Plausibility Results of Top 5 Codes (Rationales Generated by Gemini 2-Flash)}
Table \ref{code_plausibility_gemini_reformatted} summarizes the plausibility results of top 5 codes with and without incorporating few-shot human-annotated examples in the prompts. Incorporating few-shot examples substantially improves performance across all five codes and metrics. Notably, the token-level matches for I10 and the span-level matches for I4891 nearly double compared to those generated without few-shot examples.

\begin{table*}[!htbp]
\centering
\caption{Plausibility results of top 5 codes. Rationales are generated by Gemini 2-Flash. “w/o” and “w/” indicate the absence or inclusion of few-shot examples in prompts.}
\resizebox{\textwidth}{!}{
\begin{tabular}{c c c r r r r r r r r r}
\toprule
\textbf{Code} & \textbf{Metric} & \textbf{Setting} & \textbf{Prdiction} & \textbf{Accturate} & \textbf{TP} & \textbf{FP} & \textbf{FN} & \textbf{Precision (\%)} & \textbf{Recall (\%)} & \textbf{F1 (\%)} & \textbf{$\Delta$F1} \\
\midrule
\multirow{8}{*}{I10} 
& \multirow{2}{*}{Exact SM} & w/o   & 147 & 151 & 75 & 72 & 76 & 51.0 & 49.7 & 50.3 & \multirow{2}{*}{+10.2} \\
&                           & w/    & 107 & 141 & 75 & 32 & 66 & 70.1 & 53.2 & 60.5 &   \\
& \multirow{2}{*}{PI SM}    & w/o   & 140 & 82  & 70 & 70 & 12 & 50.0 & 85.4 & 63.1 & \multirow{2}{*}{+15.1} \\
&                           & w/    & 100 & 79  & 70 & 30 & 9  & 70.0 & 88.6 & 78.2 &   \\
& \multirow{2}{*}{Exact TM} & w/o   & 526 & 155 & 92 & 434 & 63 & 17.5 & 59.4 & 27.0 & \multirow{2}{*}{+26.8} \\
&                           & w/    & 182 & 145 & 88 & 94 & 57 & 48.4 & 60.7 & 53.8 &   \\
& \multirow{2}{*}{PI TM}    & w/o   & 411 & 86  & 80 & 331 & 6  & 19.5 & 93.0 & 32.2 & \multirow{2}{*}{+34.2} \\
&                           & w/    & 152 & 83  & 78 & 74 & 5  & 51.3 & 94.0 & 66.4 &   \\
\hline
\multirow{8}{*}{E785} 
& \multirow{2}{*}{Exact SM} & w/o   & 79  & 97  & 55 & 24 & 42 & 69.6 & 56.7 & 62.5 & \multirow{2}{*}{+10.7} \\
&                           & w/    & 69  & 95  & 60 & 9  & 35 & 87.0 & 63.2 & 73.2 &   \\
& \multirow{2}{*}{PI SM}    & w/o   & 72  & 56  & 49 & 23 & 7  & 68.1 & 87.5 & 76.6 & \multirow{2}{*}{+12.9} \\
&                           & w/    & 60  & 54  & 51 & 9  & 3  & 85.0 & 94.4 & 89.5 &   \\
& \multirow{2}{*}{Exact TM} & w/o   & 140 & 107 & 62 & 78 & 45 & 44.3 & 57.9 & 50.2 & \multirow{2}{*}{+16.5} \\
&                           & w/    & 84  & 99  & 61 & 23 & 38 & 72.6 & 61.6 & 66.7 &   \\
& \multirow{2}{*}{PI TM}    & w/o   & 125 & 66  & 57 & 68 & 9  & 45.6 & 86.4 & 59.7 & \multirow{2}{*}{+18.5} \\
&                           & w/    & 75  & 58  & 52 & 23 & 6  & 69.3 & 89.7 & 78.2 &   \\
\hline
\multirow{8}{*}{Z7901} 
& \multirow{2}{*}{Exact SM} & w/o   & 83  & 50  & 6  & 77 & 44 & 7.2  & 12.0 & 9.0  & \multirow{2}{*}{+4.0} \\
&                           & w/    & 102 & 52  & 10 & 92 & 42 & 9.8  & 19.2 & 13.0 &   \\
& \multirow{2}{*}{PI SM}    & w/o   & 82  & 42  & 6  & 76 & 36 & 7.3  & 14.3 & 9.7  & \multirow{2}{*}{+7.1} \\
&                           & w/    & 100 & 43  & 12 & 88 & 31 & 12.0 & 27.9 & 16.8 &   \\
& \multirow{2}{*}{Exact TM} & w/o   & 524 & 387 & 163& 361& 224& 31.1 & 42.1 & 35.8 & \multirow{2}{*}{+10.0} \\
&                           & w/    & 645 & 389 & 237& 408& 152& 36.7 & 60.9 & 45.8 &   \\
& \multirow{2}{*}{PI TM}    & w/o   & 354 & 265 & 134& 220& 131& 37.9 & 50.6 & 43.3 & \multirow{2}{*}{+11.5} \\
&                           & w/    & 413 & 266 & 186& 227& 80 & 45.0 & 69.9 & 54.8 &   \\
\hline
\multirow{8}{*}{I4891} 
& \multirow{2}{*}{Exact SM} & w/o   & 47  & 55  & 6  & 41 & 49 & 12.8 & 10.9 & 11.8 & \multirow{2}{*}{+12.9} \\
&                           & w/    & 42  & 55  & 12 & 30 & 43 & 28.6 & 21.8 & 24.7 &   \\
& \multirow{2}{*}{PI SM}    & w/o   & 45  & 27  & 6  & 39 & 21 & 13.3 & 22.2 & 16.7 & \multirow{2}{*}{+20.2} \\
&                           & w/    & 38  & 27  & 12 & 26 & 15 & 31.6 & 44.4 & 36.9 &   \\
& \multirow{2}{*}{Exact TM} & w/o   & 221 & 93  & 44 & 177& 49 & 19.9 & 47.3 & 28.0 & \multirow{2}{*}{+19.9} \\
&                           & w/    & 124 & 93  & 52 & 72 & 41 & 41.9 & 55.9 & 47.9 &   \\
& \multirow{2}{*}{PI TM}    & w/o   & 164 & 37  & 35 & 129& 2  & 21.3 & 94.6 & 34.8 & \multirow{2}{*}{+21.4} \\
&                           & w/    & 84  & 37  & 34 & 50 & 3  & 40.5 & 91.9 & 56.2 &   \\
\hline
\multirow{8}{*}{E119} 
& \multirow{2}{*}{Exact SM} & w/o   & 53  & 50  & 21 & 32 & 29 & 39.6 & 42.0 & 40.8 & \multirow{2}{*}{+3.4} \\
&                           & w/    & 54  & 50  & 23 & 31 & 27 & 42.6 & 46.0 & 44.2 &   \\
& \multirow{2}{*}{PI SM}    & w/o   & 53  & 33  & 21 & 32 & 12 & 39.6 & 63.6 & 48.8 & \multirow{2}{*}{+4.1} \\
&                           & w/    & 54  & 33  & 23 & 31 & 10 & 42.6 & 69.7 & 52.9 &   \\
& \multirow{2}{*}{Exact TM} & w/o   & 225 & 79  & 55 & 170& 24 & 24.4 & 69.6 & 36.2 & \multirow{2}{*}{+7.4} \\
&                           & w/    & 187 & 79  & 58 & 129& 21 & 31.0 & 73.4 & 43.6 &   \\
& \multirow{2}{*}{PI TM}    & w/o   & 186 & 49  & 44 & 142& 5  & 23.7 & 89.8 & 37.4 & \multirow{2}{*}{+7.4} \\
&                           & w/    & 152 & 49  & 45 & 107& 4  & 29.6 & 91.8 & 44.8 &   \\
\bottomrule
\end{tabular}
}
\label{code_plausibility_gemini_reformatted}
\end{table*}

\paragraph{Plausibility Results of LLMs-guided Rationale Learning Approaches}

Table \ref{multi} presents the plausibility results of the multi-objective model and its base model, PLM‑ICD. We compare their performance across different numbers of selected tokens, ranging from 50 to 200. The multi-objective model consistently outperforms the base model across all settings and metrics, with particularly notable gains in span-level matches. Interestingly, the improvements are more pronounced at lower thresholds, indicating that the model generates more concise rationales, whereas at higher thresholds the performance gap narrows as more tokens are incorporated.

Table \ref{NER} presents the plausibility results of the NER model trained on a small dataset of 5,000 randomly selected samples. The results show that the NER model achieves the highest span-level plausibility, even surpassing its teacher model, Gemini 2‑Flash. In contrast, PLM‑ICD performs the worst. Interestingly, training on the smaller dataset improves plausibility compared to the results in Table \ref{multi}. We attribute this finding to the influence of rationale proportions on the model’s ability to generate plausible rationales.

\begin{table*}[!htbp]
\centering
\caption{Plausibility results of multi-objective learning approach and the baseline model PLM‑ICD. All results are reported as percentages. The experiments are conducted under the Top-50 code settings.}
\label{tab:ds_plausibility}
\resizebox{\textwidth}{!}{
\begin{tabular}{c|c|ccc|ccc|ccc|ccc}
\hline
\multirow{2}{*}{Setting} & \multirow{2}{*}{Model} & \multicolumn{3}{c|}{Exact Span} & \multicolumn{3}{c|}{PI Span} & \multicolumn{3}{c|}{Exact Token} & \multicolumn{3}{c}{PI Token} \\
 & & Pr & Rc & F1 & Pr & Rc & F1 & Pr & Rc & F1 & Pr & Rc & F1 \\
\hline
\multirow{2}{*}{50} 
 & PLM-ICD & 1.5 & 14.2 & 2.7 & 1.7 & 17.2 & 3.0 & 5.5 & 35.1 & 9.5 & 7.1 & 43.8 & 12.2 \\
 & Multi-objective & 2.1 & 20.9 & 3.9 & 2.2 & 23.5 & 4.1 & 6.1 & 39.1 & 10.6 & 7.6 & 47.4 & 13.2 \\
\hdashline
\multirow{2}{*}{100} 
 & PLM-ICD & 0.6 & 9.3 & 1.1 & 0.7 & 11.8 & 1.3 & 2.9 & 36.9 & 5.5 & 4.5 & 51.7 & 8.3 \\
 & Multi-objective & 1.0 & 16.6 & 1.9 & 1.1 & 18.8 & 2.0 & 3.2 & 40.2 & 5.9 & 4.7 & 53.5 & 8.6 \\
\hdashline
\multirow{2}{*}{150} 
 & PLM-ICD & 0.3 & 7.1 & 0.6 & 0.4 & 10.0 & 0.8 & 2.0 & 37.4 & 3.8 & 3.5 & 56.9 & 6.6 \\
 & Multi-objective & 0.7 & 14.0 & 1.3 & 0.8 & 16.9 & 1.4 & 2.1 & 38.2 & 3.9 & 3.5 & 55.2 & 6.5 \\
\hdashline
\multirow{2}{*}{200} 
 & PLM-ICD & 0.2 & 6.6 & 0.5 & 0.3 & 9.7 & 0.7 & 1.5 & 36.2 & 2.8 & 2.9 & 59.9 & 5.6 \\
 & Multi-objective & 0.5 & 12.4 & 1.0 & 0.6 & 14.9 & 1.1 & 1.6 & 38.4 & 3.0 & 2.9 & 58.8 & 5.5 \\
\hline
\end{tabular}
}
\label{multi}
\end{table*}

\begin{table*}[!htbp]
\centering
\caption{Plausibility results for the NER model, the teacher model Gemini 2‑Flash, and the baseline model PLM‑ICD. All results are reported as percentages. The experiments are conducted under the Top-50 code settings.}
\label{tab:ner_plausibility}
\resizebox{\textwidth}{!}{
\begin{tabular}{c|c|ccc|ccc|ccc|ccc}
\hline
\multirow{2}{*}{Setting} & \multirow{2}{*}{Model} & \multicolumn{3}{c|}{Exact Span} & \multicolumn{3}{c|}{PI Span} & \multicolumn{3}{c|}{Exact Token} & \multicolumn{3}{c}{PI Token} \\
 & & Pr & Rc & F1 & Pr & Rc & F1 & Pr & Rc & F1 & Pr & Rc & F1 \\
\hline
\multirow{1}{*}{-} 
 & NER Model & 29.3 & 24.3 & 26.5 & 29.0 & 32.5 & 30.6 & 35.1 & 15.8 & 21.8 & 37.1 & 21.2 & 27.0 \\
\hdashline
\multirow{1}{*}{-} 
 & Gemini 2-Flash & 16.4 & 20.5 & 18.2 & 18.2 & 31.2 & 23.0 & 23.0 & 40.6 & 29.4 & 23.7 & 46.4 & 31.4 \\
\hdashline
\multirow{1}{*}{50} 
 & PLM-ICD & 2.5 & 12.8 & 4.1 & 2.5 & 14.8 & 4.3 & 4.7 & 28.2 & 8.1 & 7.0 & 41.6 & 12.0 \\
\multirow{1}{*}{100} 
 & PLM-ICD & 1.5 & 11.9 & 2.6 & 1.6 & 14.3 & 2.8 & 2.7 & 31.5 & 4.9 & 4.7 & 51.2 & 8.7 \\
\multirow{1}{*}{150} 
 & PLM-ICD & 0.9 & 9.7 & 1.7 & 1.0 & 12.6 & 1.9 & 1.9 & 33.0 & 3.5 & 3.7 & 56.4 & 6.9 \\
\multirow{1}{*}{200} 
 & PLM-ICD & 0.6 & 7.5 & 1.1 & 0.7 & 10.0 & 1.3 & 1.5 & 35.1 & 2.9 & 3.1 & 60.5 & 6.0 \\
\hline
\end{tabular}
}
\label{NER}
\end{table*}

\paragraph{Plausibility Results of Top 5 Codes (Rationales Generated by NER with Supervision of Rationales Labels Generated by Gemini 2-Flash)}

Table \ref{tab:code_NER} presents the plausibility results for the top 5 codes. The rationales are generated using NER models trained on rationales produced by Gemini 2‑Flash, with and without the inclusion of few‑shot human‑annotated examples in the prompts. Models incorporating few‑shot examples demonstrate improved performance across most codes and metrics. Furthermore, the supervised model outperforms the unsupervised model for codes I10, E785, and E119, due to the higher frequency of corresponding rationales in the documents.

\begin{table*}[h!]
\centering
\caption{Plausibility results of top 5 codes. Rationales are generated by NER models. “w/o” and “w/” indicate the absence or inclusion of few-shot examples in prompts.}
\resizebox{\textwidth}{!}{
\begin{tabular}{c c c r r r r r r r r r}
\hline
\textbf{Code} & \textbf{Metric} & \textbf{Setting} & \textbf{Prdiction} & \textbf{Accturate} & \textbf{TP} & \textbf{FP} & \textbf{FN} & \textbf{Precision (\%)} & \textbf{Recall (\%)} & \textbf{F1 (\%)}  & \textbf{$\Delta$F1} \\
\hline
\multirow{8}{*}{I4891} 
& \multirow{2}{*}{Exact SM} & w/o   & 44  & 55  & 11  & 33  & 44  & 25.0 & 20.0 & 22.2 &  \multirow{2}{*}{+3.9} \\
&                           & w/    & 37  & 55  & 12  & 25  & 43  & 32.4 & 21.8 & 26.1 &   \\
& \multirow{2}{*}{PI SM}    & w/o   & 42  & 27  & 13  & 29  & 14  & 31.0 & 48.1 & 37.7 &  \multirow{2}{*}{+5.6} \\
&                           & w/    & 33  & 27  & 13  & 20  & 14  & 39.4 & 48.1 & 43.3 &   \\
& \multirow{2}{*}{Exact TM} & w/o   & 115 & 93  & 40  & 75  & 53  & 34.8 & 43.0 & 38.5 &  \multirow{2}{*}{+2.2} \\
&                           & w/    & 89  & 93  & 37  & 52  & 56  & 41.6 & 39.8 & 40.7 &   \\
& \multirow{2}{*}{PI TM}    & w/o   & 86  & 37  & 30  & 56  & 7   & 34.9 & 81.1 & 48.8 &  \multirow{2}{*}{+5.7} \\
&                           & w/    & 62  & 37  & 27  & 35  & 10  & 43.5 & 73.0 & 54.5 &   \\
\hline
\multirow{8}{*}{I10} 
& \multirow{2}{*}{Exact SM} & w/o   & 80  & 144 & 62  & 18  & 82  & 77.5 & 43.1 & 55.4 &  \multirow{2}{*}{+7.1} \\
&                           & w/    & 80  & 144 & 70  & 10  & 74  & 87.5 & 48.6 & 62.5 &   \\
& \multirow{2}{*}{PI SM}    & w/o   & 74  & 82  & 60  & 14  & 22  & 81.1 & 73.2 & 76.9 &  \multirow{2}{*}{+8.3} \\
&                           & w/    & 73  & 82  & 66  & 7   & 16  & 90.4 & 80.5 & 85.2 &   \\
& \multirow{2}{*}{Exact TM} & w/o   & 93  & 147 & 67  & 26  & 80  & 72.0 & 45.6 & 55.8 &  \multirow{2}{*}{+9.1} \\
&                           & w/    & 84  & 147 & 75  & 9   & 72  & 89.3 & 51.0 & 64.9 &   \\
& \multirow{2}{*}{PI TM}    & w/o   & 85  & 85  & 64  & 21  & 21  & 75.3 & 75.3 & 75.3 &  \multirow{2}{*}{+10.9} \\
&                           & w/    & 75  & 85  & 69  & 6   & 16  & 92.0 & 81.2 & 86.2 &   \\
\hline
\multirow{8}{*}{E785} 
& \multirow{2}{*}{Exact SM} & w/o   & 62  & 103 & 56  & 6   & 47  & 90.3 & 54.4 & 67.9 &  \multirow{2}{*}{+2.4} \\
&                           & w/    & 62  & 103 & 58  & 4   & 45  & 93.5 & 56.3 & 70.3 &   \\
& \multirow{2}{*}{PI SM}    & w/o   & 56  & 61  & 50  & 6   & 11  & 89.3 & 82.0 & 85.5 &  \multirow{2}{*}{+1.5} \\
&                           & w/    & 54  & 61  & 50  & 4   & 11  & 92.6 & 82.0 & 87.0 &   \\
& \multirow{2}{*}{Exact TM} & w/o   & 64  & 121 & 57  & 7   & 64  & 89.1 & 47.1 & 61.6 &  \multirow{2}{*}{+1.4} \\
&                           & w/    & 63  & 121 & 58  & 5   & 63  & 92.1 & 47.9 & 63.0 &   \\
& \multirow{2}{*}{PI TM}    & w/o   & 57  & 76  & 50  & 7   & 26  & 87.7 & 65.8 & 75.2 &  \multirow{2}{*}{+1.1} \\
&                           & w/    & 55  & 76  & 50  & 5   & 26  & 90.9 & 65.8 & 76.3 &   \\
\hline
\multirow{8}{*}{E119} 
& \multirow{2}{*}{Exact SM} & w/o   & 37  & 52  & 22  & 15  & 30  & 59.5 & 42.3 & 49.4 &  \multirow{2}{*}{-3.6} \\
&                           & w/    & 31  & 52  & 19  & 12  & 33  & 61.3 & 36.5 & 45.8 &   \\
& \multirow{2}{*}{PI SM}    & w/o   & 36  & 35  & 22  & 14  & 13  & 61.1 & 62.9 & 62.0 &  \multirow{2}{*}{-3.5} \\
&                           & w/    & 30  & 35  & 19  & 11  & 16  & 63.3 & 54.3 & 58.5 &   \\
& \multirow{2}{*}{Exact TM} & w/o   & 62  & 96  & 42  & 20  & 54  & 67.7 & 43.7 & 53.2 &  \multirow{2}{*}{-3.5} \\
&                           & w/    & 53  & 96  & 37  & 16  & 59  & 69.8 & 38.5 & 49.7 &   \\
& \multirow{2}{*}{PI TM}    & w/o   & 51  & 58  & 34  & 17  & 24  & 66.7 & 58.6 & 62.4 &  \multirow{2}{*}{-2.2} \\
&                           & w/    & 45  & 58  & 31  & 14  & 27  & 68.9 & 53.4 & 60.2 &   \\
\hline
\multirow{8}{*}{Z7901} 
& \multirow{2}{*}{Exact SM} & w/o   & 45  & 52  & 5   & 40  & 47  & 11.1 & 9.6  & 10.3 &  \multirow{2}{*}{+0.5} \\
&                           & w/    & 59  & 52  & 6   & 53  & 46  & 10.2 & 11.5 & 10.8 &   \\
& \multirow{2}{*}{PI SM}    & w/o   & 44  & 43  & 6   & 38  & 37  & 13.6 & 14.0 & 13.8 &  \multirow{2}{*}{-1.8} \\
&                           & w/    & 57  & 43  & 6   & 51  & 37  & 10.5 & 14.0 & 12.0 &   \\
& \multirow{2}{*}{Exact TM} & w/o   & 167 & 389 & 71  & 96  & 318 & 42.5 & 18.3 & 25.5 &  \multirow{2}{*}{-3.0} \\
&                           & w/    & 190 & 389 & 65  & 125 & 324 & 34.2 & 16.7 & 22.5 &   \\
& \multirow{2}{*}{PI TM}    & w/o   & 138 & 266 & 78  & 60  & 188 & 56.5 & 29.3 & 38.6 &  \multirow{2}{*}{+2.1} \\
&                           & w/    & 157 & 266 & 86  & 71  & 180 & 54.8 & 32.3 & 40.7 &   \\
\hline
\end{tabular}
}
\label{tab:code_NER}
\end{table*}

\section{A Case Study: Comparative Analysis of Rationales From Four Sources}

To examine the rationales produced by different sources in detail, we present a case study comparing Human Annotation, Entity-Linking, Gemini 2‑Flash, and PLM‑ICD. The focus is on rationales for ICD‑10 code N49.2 – Inflammatory disorders of scrotum. We randomly select the document HADM‑ID 29964986, to illustrate how each method justifies the assigned code.

\begin{figure*}[!htbp]
\centering
\includegraphics[width=0.95\textwidth]{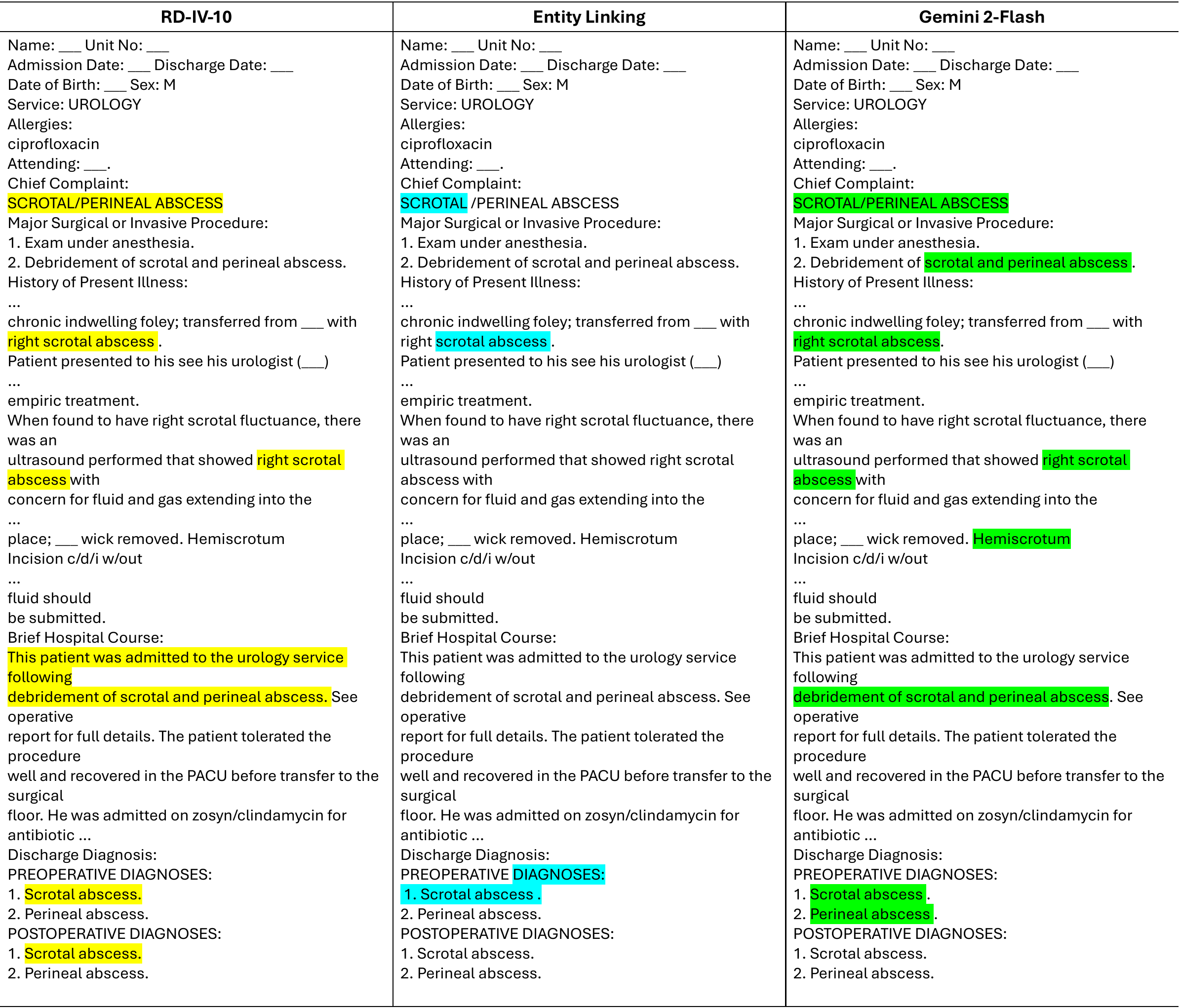} % Reduce the figure size so that it is slightly narrower than the column.
\caption{A case study of annotations of RD-IV-10, Entity Linking and Gemini 2-Flash. The highlights indicate their corresponding annotations.}
\label{case_three}
\end{figure*}

%Mingyang: Which table?
Table \ref{case_three} shows that Entity Linking and Human Annotation partially overlap. Entity Linking highlights only the direct mentions of the code, such as `SCROTAL' and `scrotal abscess', whereas Human Annotation additionally mark broader contextual information—the sentence: `This patient was admitted to the urology service following debridement of scrotal and perineal abscess.' Gemini 2‑Flash also performs well in this case. It highlights phrases overlooked by both human annotators and Entity Linking—such as `scrotal and perineal abscess' and `right scrotal fluctuance'—and even captures parts of that human-annotated sentence. However, it also returns some incorrect spans, including `Perineal abscess' and `Hemiscrotum'. Additionally, we observe that the rationales of PLM‑ICD are randomly distributed and lack meaningful explainability; therefore, we do not visualize them in this case study. This case study supports our earlier conclusion that Gemini 2‑Flash achieves the highest plausibility, while naive directly linked entities can also serve as rationales to a certain extent. In contrast, the rationales generated by PLM‑ICD exhibit the lowest plausibility.
%
%Mingyang: Can you summarise the finding in above paragraph with one sentence in the end?

%Mingyang: What is this about? 
%In PLM‑ICD, rationales are tokens with top 10\% highest attention weight. Its highlights appear random and lack clearly interpretable meaning.

\section{Trade-off Between ICD Coding Performance and Rationale Plausibility in an NER-Based Approach Across Different Training Data Sizes}

We also analyse the trade-off between coding performance (Table \ref{trade_off_coding}) and rationale plausibility (Table \ref{trade_off_plausibility}) under different training-data sizes. When the number of training samples is reduced by 80\% (retaining only 1000 samples), coding performance drops by about 10\% (F1 scores), while plausibility decreases by approximately 10–16\%, indicating comparable levels of degradation. Additionally, we observe that both coding performance and plausibility converge as the training size increases - the results with 3,000 and 5,000 samples are highly similar. This suggests that even relatively small training sets are sufficient to achieve high-level plausibility.

\begin{table*}[t]
\centering
\caption{Trade-off with Different Training-Data Sizes -- ICD Coding Performance (NER).}
\begin{tabular}{lcccccc}
\toprule
\textbf{Size} & \textbf{F1-Mac} & \textbf{F1-Mic} & \textbf{P-Mac} & \textbf{P-Mic} & \textbf{R-Mac} & \textbf{R-Mic} \\
\midrule
1000 & 49.14 & 60.88 & 41.69 & 50.60 & 62.95 & 76.39 \\
3000 & 54.00 & 68.22 & 48.76 & 60.30 & 64.13 & 78.54 \\
5000 & 53.46 & 67.75 & 49.21 & 60.93 & 61.79 & 76.30 \\
\bottomrule
\end{tabular}
\label{trade_off_coding}
\end{table*}

\begin{table*}[t]
\centering
\caption{Trade-off with Different Training-Data Sizes -- Rationale Plausibility (NER).}
\resizebox{0.97\textwidth}{!}{
\begin{tabular}{lccccccccc}
\toprule
\textbf{Metric} & \textbf{Size} & \textbf{Prediction} & \textbf{Accurate} & \textbf{TP} & \textbf{FP} & \textbf{FN} &
\textbf{Precision (\%)} & \textbf{Recall (\%)} & \textbf{F1 (\%)} \\
\midrule
& 1000 & 1551 & 1607 & 358 & 1193 & 1249 & 23.1 & 22.3 & 22.7 \\
\textbf{Exact SM} & 3000 & 1420 & 1607 & 416 & 1004 & 1191 & 29.3 & 25.9 & 27.5 \\
& 5000 & 1332 & 1607 & 390 & 942 & 1217 & 29.3 & 24.3 & 26.5 \\
\midrule
& 1000 & 1502 & 1145 & 343 & 1159 & 802 & 22.8 & 30.0 & 25.9 \\
\textbf{PI SM} & 3000 & 1368 & 1145 & 400 & 968 & 745 & 29.2 & 34.9 & 31.8 \\
& 5000 & 1283 & 1145 & 372 & 911 & 773 & 29.0 & 32.5 & 30.6 \\
\midrule
& 1000 & 2988 & 5531 & 803 & 2185 & 4728 & 26.9 & 14.5 & 18.9 \\
\textbf{Exact TM} & 3000 & 2713 & 5531 & 895 & 1818 & 4636 & 33.0 & 16.2 & 21.7 \\
& 5000 & 2493 & 5531 & 874 & 1619 & 4657 & 35.1 & 15.8 & 21.8 \\
\midrule
& 1000 & 2684 & 3891 & 799 & 1885 & 3092 & 29.8 & 20.5 & 24.3 \\
\textbf{PI TM} & 3000 & 2439 & 3891 & 852 & 1587 & 3039 & 34.9 & 21.9 & 26.9 \\
& 5000 & 2222 & 3891 & 825 & 1397 & 3066 & 37.1 & 21.2 & 27.0 \\
\bottomrule
\end{tabular}
}
\label{trade_off_plausibility}
\end{table*}

\section{Statistics of Gemini-2 Flash Generated Rationale Dataset Across Different Alignment Settings (Overlap)}

We use Gemini 2-Flash to generate rationale labels as weak supervision signals for all 122,004 samples in MIMIC-IV for rationale learning. The statistical analyses under different overlap-score thresholds for the alignments are presented in Table~\ref{distribution_alignment}. We observe that 93.35\% of the spans extracted by the LLM are nearly identical to the original text (an overlap score of 2 indicates an exact match).

\begin{table*}[ht]
\centering
\caption{Counts and ratios of rationales across different overlap scores.}
\begin{tabular}{lrr|lrr}
\hline
\textbf{Range} & \textbf{Count} & \textbf{Ratio (\%)} & \textbf{Range} & \textbf{Count} & \textbf{Ratio (\%)}  \\
\hline
0.0--0.1 & 174{,}730 & 2.82\% & 1.0--1.1 & 11{,}026  & 0.18\% \\
0.1--0.2 & 24{,}021  & 0.39\% & 1.1--1.2 & 2{,}985   & 0.05\% \\
0.2--0.3 & 64{,}776  & 1.05\% & 1.2--1.3 & 1{,}637   & 0.03\% \\
0.3--0.4 & 63{,}037  & 1.02\% & 1.3--1.4 & 10{,}390  & 0.17\% \\
0.4--0.5 & 48{,}663  & 0.79\% & 1.4--1.5 & 1{,}770   & 0.03\% \\
0.5--0.6 & 36{,}614  & 0.59\% & 1.5--1.6 & 9{,}493   & 0.15\% \\
0.6--0.7 & 19{,}274  & 0.31\% & 1.6--1.7 & 17{,}472  & 0.28\% \\
0.7--0.8 & 7{,}397   & 0.12\% & 1.7--1.8 & 17{,}392  & 0.28\% \\
0.8--0.9 & 7{,}648   & 0.12\% & 1.8--1.9 & 30{,}288  & 0.49\% \\
0.9--1.0 & 4{,}013   & 0.06\% & 1.9--2.0 & 5{,}780{,}713 & 93.35\% \\
\hline
 & & & \textbf{Total} & 6{,}191{,}350 & 100\% \\
\hline
\end{tabular}
\label{distribution_alignment}
\end{table*}

\section{Plausibility Across Different Alignment Settings (Overlap)}

We also compare the plausibility results for Gemini 2-Flash across different alignment settings, where the rationales are generated in a separate round. From Table \ref{alignment}, we observe that higher alignment quality consistently yields higher scores across all four metrics. More specifically, including spans with lower overlap scores substantially degrades token-level matching performance, while having only a minor impact on span-level metrics. This is because target spans often cover a wide range of tokens, so adding misaligned spans introduces many irrelevant tokens at the token level, whereas at the span level the misalignment affects the counts of TP, FP, and FN by only ±1.

\begin{table*}[!htbp]
\centering
\caption{Plausibility results for Gemini 2-Flash across different alignment settings (Overlap).}
\resizebox{\textwidth}{!}{
\begin{tabular}{cccccccccc}
\hline
\textbf{Metric} & \textbf{Overlap}  & \textbf{Prdiction} & \textbf{Accurate} & \textbf{TP} & \textbf{FP} & \textbf{FN} & \textbf{Precision (\%)} & \textbf{Recall (\%)} & \textbf{F1 (\%)} \\
\hline
\multirow{5}{*}{Exact SM} & 0 & 3418 & 2260 & 711 & 2707 & 1549 & 20.8 & 31.5 & 25.0 \\
& 0.5 & 3303 & 2260 & 711 & 2592 & 1549 & 21.5 & 31.5 & 25.6 \\
& 1 & 3180 & 2260 & 710 & 2470 & 1550 & 22.3 & 31.4 & 26.1 \\
& 1.5 & 3157 & 2260 & 704 & 2453 & 1556 & 22.3 & 31.2 & 26.0 \\
& 2 & 3105 & 2260 & 700 & 2405 & 1560 & 22.5 & 31.0 & 26.1 \\
\hline
\multirow{5}{*}{PI SM} & 0 & 3365 & 1762 & 730 & 2635 & 1032 & 21.7 & 41.4 & 28.5 \\
& 0.5 & 3250 & 1762 & 730 & 2520 & 1032 & 22.5 & 41.4 & 29.1 \\
& 1 & 3127 & 1762 & 729 & 2398 & 1033 & 23.3 & 41.4 & 29.8 \\
& 1.5 & 3105 & 1762 & 724 & 2381 & 1038 & 23.3 & 41.1 & 29.8 \\
& 2 & 3053 & 1762 & 720 & 2333 & 1042 & 23.6 & 40.9 & 29.9 \\
\hline
\multirow{5}{*}{Exact TM} & 0 & 64247 & 11422 & 5293 & 58954 & 6129 & 8.2 & 46.3 & 14.0 \\
& 0.5 & 53179 & 11422 & 5189 & 47990 & 6233 & 9.8 & 45.4 & 16.1 \\
& 1 & 20494 & 11422 & 4976 & 15518 & 6446 & 24.3 & 43.6 & 31.2 \\
& 1.5 & 18151 & 11422 & 4911 & 13240 & 6511 & 27.1 & 43.0 & 33.2 \\
& 2 & 17723 & 11422 & 4867 & 12856 & 6555 & 27.5 & 42.6 & 33.4 \\
\hline
\multirow{5}{*}{PI TM} & 0 & 36499 & 8160 & 4739 & 31760 & 3421 & 13.0 & 58.1 & 21.2 \\
& 0.5 & 29972 & 8160 & 4617 & 25355 & 3543 & 15.4 & 56.6 & 24.2 \\
& 1 & 14343 & 8160 & 4383 & 9960 & 3777 & 30.6 & 53.7 & 39.0 \\
& 1.5 & 13086 & 8160 & 4329 & 8757 & 3831 & 33.1 & 53.1 & 40.8 \\
& 2 & 12843 & 8160 & 4306 & 8537 & 3854 & 33.5 & 52.8 & 41.0 \\
\hline
\end{tabular}
}
\label{alignment}
\end{table*}

\section{Impact of Label Selection under Different Span Alignment Settings on Rationale Learning}

In Tables \ref{ner_supervised_coding_overlap} and \ref{ner_plausibility_overlap}, we report the downstream learning outcomes under different span alignment settings. We train the NER model using all spans (threshold = 0). We find that lower alignment does not substantially affect ICD coding performance or plausibility scores, because the proportion of training data with low alignment is small, with spans in the range 0.0–1.9 accounting for less than 7\% of the dataset.

\begin{table*}[t]
        \centering
        \captionof{table}{Effect of Span Alignment on Rationale Learning (NER) - ICD Coding Performance.}
\begin{tabular}{ccccccc}
\hline
\textbf{Threshold} & \textbf{F1-Mac} & \textbf{F1-Mic} & \textbf{P-Mac} & \textbf{P-Mic} & \textbf{R-Mac} & \textbf{R-Mic} \\
\hline
0 & 53.35 & 68.65 & 49.36 & 61.12 & 61.76 & 78.30\\
%1 & 52.92 & 65.98 & 45.79 & 56.50 & 64.99 & 79.28 \\
%2 & 54.31 & 68.24 & 54.97 & 64.08 & 58.77 & 72.97 \\
1.7 & 53.46	& 67.75 & 49.21 & 60.93 & 61.79 & 76.30 \\
\hline
\label{ner_supervised_coding_overlap}
\end{tabular}
\end{table*}

\begin{table*}[!htbp]
\centering
\caption{Effect of Span Alignment on Rationale Learning (NER) - Plausibility.}
\resizebox{\textwidth}{!}{
\begin{tabular}{c|ccc|ccc|ccc|ccc}
\hline
\multirow{2}{*}{Threshold} & \multicolumn{3}{c|}{Exact Span} & \multicolumn{3}{c|}{PI Span} & \multicolumn{3}{c|}{Exact Token} & \multicolumn{3}{c}{PI Token} \\
 & Pr & Rc & F1 & Pr & Rc & F1 & Pr & Rc & F1 & Pr & Rc & F1 \\
\hline
 0 & 29.5 & 25.1 & 27.1 & 29.8 & 34.3 & 31.9 & 32.2 & 15.6 & 21.1 & 35.3 & 21.7 & 26.9 \\
 1.7 & 29.3 & 24.3 & 26.5 & 29.0 & 32.5 & 30.6 & 35.1 & 15.8 & 21.8 & 37.1 & 21.2 & 27.0 \\
\hline
\end{tabular}
}
\label{ner_plausibility_overlap}
\end{table*}

\begin{table*}[!htbp]
\centering
\begin{adjustbox}{angle=270}
\begin{tabular}{@{} m{0.90\textheight} m{3.3cm} @{}}
    \captionof{table}{Faithfulness across Top-P and Top-N thresholds for CAML.}
\resizebox{23cm}{6cm}{%
\begin{tabular}{c|cccccc|cccccc|cccccc|cccccc}
\hline
\textbf{Dataset} & \multicolumn{6}{c}{\textbf{MIMIC-IV ICD10 Full}} & \multicolumn{6}{c}{\textbf{MIMIC-IV ICD10 Top-50}} & \multicolumn{6}{c}{\textbf{MIMIC-III ICD9 Full}} &  \multicolumn{6}{c}{\textbf{MIMIC-III ICD9 Top-50}}\\
\hline
\textbf{Threshold} & \textbf{F1-Mac} & \textbf{F1-Mic} & \textbf{AUC-Mac} & \textbf{AUC-Mic} & \textbf{P@8} & \textbf{Retention} & \textbf{F1-Mac} & \textbf{F1-Mic} & \textbf{AUC-Mac} & \textbf{AUC-Mic} & \textbf{P@5} & \textbf{Retention} & \textbf{F1-Mac} & \textbf{F1-Mic} & \textbf{AUC-Mac} & \textbf{AUC-Mic} & \textbf{P@8} & \textbf{Retention} & \textbf{F1-Mac} & \textbf{F1-Mic} & \textbf{AUC-Mac} & \textbf{AUC-Mic} & \textbf{P@5} & \textbf{Retention}\\
\hline
- & 15.82 & 54.87 & 91.01 & 98.52 & 66.09 & 100.00 & 65.78 & 71.23 & 91.95 & 94.58 & 63.53 & 100.00 & 9.93 & 50.84 & 81.87 & 97.46 & 70.81 & 100.00 & 55.62 & 63.98 & 89.77 & 92.56 & 63.32 & 100.00\\
\hline
\multicolumn{25}{c}{\textbf{Top-p Sufficiency}} \\
\hdashline
10\% & 7.72 & 30.00 & 79.09 & 93.61 & 43.63 & 12.05 & 56.90 & 62.76 & 86.23 & 89.17 & 46.09 & 15.18 & 5.05 & 17.82 & 67.95 & 91.84 & 41.54 & 10.99 & 33.45 & 35.69 & 80.55 & 81.50 & 40.45 & 15.26\\
20\% & 9.15 & 35.42 & 83.97 & 96.15 & 48.90 & 21.83  & 58.97 & 64.85 & 88.49 & 91.28 & 47.03 & 24.61   & 6.03 & 22.82 & 71.79 & 93.81 & 49.18 & 20.88  & 37.31 & 40.27 & 83.23 & 85.25 & 42.66 & 24.67\\
30\% & 10.12 & 38.88 & 86.22 & 97.11 & 52.09 & 31.60  & 60.10 & 65.96 & 89.57 & 92.30 & 47.63 & 34.03   & 6.61 & 26.53 & 74.28 & 94.95 & 53.59 & 30.77  & 40.55 & 44.19 & 85.06 & 87.68 & 44.26 & 34.09\\
40\% & 10.83 & 41.33 & 87.76 & 97.61 & 54.35 & 41.37  & 60.87 & 66.70 & 90.16 & 92.92 & 48.07 & 43.45   & 7.02 & 29.41 & 75.92 & 95.66 & 56.65 & 40.66  & 43.01 & 47.22 & 86.31 & 89.15 & 45.41 & 43.51\\
50\% & 11.36 & 43.16 & 88.76 & 97.91 & 55.99 & 51.14  & 61.48 & 67.28 & 90.60 & 93.36 & 48.42 & 52.88   & 7.33 & 31.81 & 76.87 & 96.18 & 58.81 & 50.55  & 44.88 & 49.53 & 87.17 & 90.09 & 46.27 & 52.92\\
60\% & 11.81 & 44.61 & 89.49 & 98.13 & 57.29 & 60.91 & 61.97 & 67.74 & 90.94 & 93.69 & 48.71 & 62.30 & 7.56 & 33.76 & 78.02 & 96.55 & 60.42 & 60.44  & 46.24 & 51.28 & 87.82 & 90.77 & 46.91 & 62.34\\
70\% & 12.19 & 45.78 & 90.04 & 98.27 & 58.33 & 70.68 & 62.38 & 68.13 & 91.20 & 93.95 & 48.94 & 71.73 & 7.74 & 35.34 & 78.43 & 96.73 & 61.67 & 70.33  & 47.26 & 52.63 & 88.33 & 91.29 & 47.39 & 71.75\\
80\% & 15.14 & 53.93 & 90.47 & 98.38 & 65.24 & 80.46 & 62.73 & 68.45 & 91.46 & 94.17 & 49.14 & 81.15  & 7.88 & 36.65 & 78.76 & 96.85 & 62.68 & 80.22  & 48.03 & 53.68 & 88.71 & 91.65 & 47.78 & 81.17\\
90\% & 15.39 & 54.21 & 90.78 & 98.46 & 65.47 & 90.23  & 63.02 & 68.73 & 91.71 & 94.37 & 49.33 & 90.58   & 8.01 & 37.77 & 79.06 & 96.96 & 63.51 & 90.11  & 48.62 & 54.51 & 89.23 & 92.11 & 48.11 & 90.58\\
\hline
\multicolumn{25}{c}{\textbf{Top-p Comprehensiveness}} \\
\hdashline
10\% & 12.62 & 46.54 & 89.17 & 97.92 & 57.84 & 90.23  & 13.72 & 15.39 & 71.68 & 71.27 & 22.78 & 90.58  & 2.80 & 13.54 & 75.15 & 95.37 & 35.25 & 90.11  & 8.62 & 9.66 & 75.71 & 74.72 & 29.31 & 90.58\\
20\% & 11.97 & 43.43 & 87.35 & 97.17 & 54.37 & 80.46  & 12.73 & 14.58 & 67.88 & 67.80 & 21.55 & 81.15  & 2.59 & 12.32 & 72.10 & 93.60 & 31.17 & 80.22  & 6.90 & 7.75 & 66.85 & 65.97 & 26.49 & 81.17\\
30\% & 11.33 & 40.59 & 85.54 & 96.34 & 51.17 & 70.68  & 11.97 & 14.03 & 65.01 & 65.33 & 20.63 & 71.73  & 2.45 & 11.31 & 69.62 & 92.06 & 28.28 & 70.33  & 6.20 & 6.98 & 60.43 & 61.35 & 24.80 & 71.75\\
40\% & 10.73 & 37.82 & 84.00 & 95.44 & 48.03 & 60.91  & 11.29 & 13.55 & 62.51 & 63.21 & 19.82 & 62.30  & 2.36 & 10.56 & 67.50 & 90.71 & 26.04 & 60.44  & 5.84 & 6.59 & 56.08 & 58.58 & 23.64 & 62.34\\
50\% & 10.11 & 34.98 & 82.40 & 94.31 & 44.93 & 51.14  & 10.74 & 13.18 & 60.01 & 61.34 & 19.10 & 52.88  & 2.29 & 9.84  & 65.22 & 89.41 & 24.06 & 50.55  & 5.77 & 6.49 & 53.10 & 56.58 & 22.77 & 52.92\\
60\% & 9.46 & 31.89 & 80.53 & 92.84 & 41.79 & 41.37  & 10.29 & 12.86 & 57.61 & 59.61 & 18.43 & 43.45  & 2.20 & 9.07  & 63.41 & 88.11 & 22.34 & 40.66  & 5.77 & 6.50 & 50.71 & 54.94 & 22.07 & 43.51\\
70\% & 8.77 & 28.38 & 78.43 & 90.86 & 38.64 & 31.60  & 9.87  & 12.55 & 55.22 & 58.08 & 17.85 & 34.03  & 2.13 & 8.34  & 61.66 & 86.76 & 20.79 & 30.77  & 5.80 & 6.54 & 48.70 & 53.44 & 21.47 & 34.09\\
80\% & 4.16 & 7.99 & 75.30 & 87.81 & 12.98 & 21.83  & 9.52  & 12.28 & 53.15 & 56.62 & 17.31 & 24.61  & 2.00 & 7.30  & 59.62 & 84.65 & 19.27 & 20.88  & 5.90 & 6.62 & 47.25 & 51.91 & 20.91 & 24.67\\
90\% & 3.17 & 4.81 & 70.84 & 82.12 & 9.55 & 12.05  & 9.23  & 12.09 & 51.01 & 55.13 & 16.79 & 15.18  & 1.76 & 5.64  & 56.38 & 80.38 & 17.70 & 10.99  & 6.03 & 6.79 & 46.07 & 50.32 & 20.31 & 15.26\\
\hline
\hline
\multicolumn{25}{c}{\textbf{Top-N Sufficiency}} \\
\hdashline
200  & 9.13  & 34.84 & 81.37 & 94.70 & 48.39 & 16.61   & 58.68 & 64.53 & 87.02 & 89.94 & 46.77 & 19.11  & 5.78 & 22.00 & 70.01 & 92.67 & 45.67 & 14.24   & 35.27 & 37.86 & 81.21 & 82.40 & 41.27 & 18.07\\
400  & 10.49 & 39.54 & 85.53 & 96.75 & 52.93 & 30.71   & 60.24 & 66.10 & 88.97 & 91.81 & 47.62 & 32.37   & 6.70 & 27.69 & 73.45 & 94.47 & 52.63 & 27.31   & 39.15 & 42.49 & 83.81 & 85.96 & 43.44 & 30.24\\
600  & 11.40 & 42.44 & 87.53 & 97.52 & 55.60 & 44.48  & 61.19 & 67.01 & 89.91 & 92.70 & 48.17 & 45.39   & 7.26 & 31.74 & 75.62 & 95.55 & 56.68 & 40.06   & 42.15 & 46.12 & 85.34 & 88.03 & 44.92 & 42.10\\
800  & 12.03 & 44.48 & 88.84 & 97.91 & 57.44 & 57.43  & 61.87 & 67.65 & 90.47 & 93.26 & 48.57 & 57.80  & 7.61 & 34.56 & 77.10 & 96.14 & 59.25 & 52.18   & 44.37 & 48.85 & 86.42 & 89.31 & 45.98 & 53.37\\
1000 & 12.51 & 45.99 & 89.60 & 98.14 & 58.79 & 68.87  & 62.36 & 68.13 & 90.89 & 93.67 & 48.88 & 68.93  & 7.86 & 36.70 & 77.82 & 96.54 & 61.09 & 63.17  & 46.04 & 50.93 & 87.28 & 90.25 & 46.74 & 63.74\\
1200 & 12.90 & 47.16 & 90.12 & 98.28 & 59.82 & 78.15  & 62.76 & 68.51 & 91.23 & 93.97 & 49.13 & 78.06  & 8.06 & 38.36 & 78.55 & 96.79 & 62.45 & 72.61  & 47.28 & 52.51 & 87.87 & 90.88 & 47.29 & 72.84\\
1400 & 13.22 & 48.08 & 90.47 & 98.37 & 60.60 & 85.12  & 63.10 & 68.83 & 91.46 & 94.17 & 49.35 & 84.98  & 8.22 & 39.67 & 79.14 & 96.95 & 63.49 & 80.17  & 48.26 & 53.77 & 88.36 & 91.37 & 47.74 & 80.22\\
1600 & 13.48 & 48.82 & 90.66 & 98.42 & 61.24 & 90.08  & 63.38 & 69.09 & 91.63 & 94.31 & 49.53 & 89.96  & 8.35 & 40.72 & 79.77 & 97.08 & 64.32 & 85.94  & 49.01 & 54.77 & 88.80 & 91.74 & 48.10 & 85.89\\
1800 & 13.70 & 49.43 & 90.80 & 98.46 & 61.75 & 93.53  & 63.61 & 69.30 & 91.73 & 94.40 & 49.68 & 93.44  & 8.45 & 41.60 & 80.12 & 97.18 & 64.98 & 90.27  & 49.62 & 55.60 & 89.13 & 92.04 & 48.40 & 90.12\\
2000 & 13.88 & 49.93 & 90.91 & 98.48 & 62.16 & 95.83  & 63.81 & 69.48 & 91.83 & 94.47 & 49.81 & 95.78  & 8.54 & 42.34 & 80.71 & 97.27 & 65.53 & 93.42  & 50.13 & 56.29 & 89.31 & 92.22 & 48.66 & 93.19\\
\hline
\multicolumn{25}{c}{\textbf{Top-N Comprehensiveness}} \\
\hdashline
200  & 11.54 & 43.65 & 88.26 & 97.59 & 54.84 & 85.74   & 13.28 & 15.04 & 70.03 & 69.81 & 21.82 & 86.64   & 2.61  & 12.73 & 73.94 & 94.75 & 32.95 & 86.85  & 8.94  & 9.69  & 73.65 & 72.90 & 28.46 & 87.77\\
400  & 10.36 & 38.18 & 85.78 & 96.38 & 50.36 & 71.64  & 12.20 & 14.23 & 65.25 & 65.56 & 20.45 & 73.38  & 2.23  & 9.66  & 69.99 & 92.10 & 28.98 & 73.79  & 7.12  & 7.84  & 64.24 & 64.55 & 25.90 & 75.60\\
600  & 9.01  & 30.66 & 83.31 & 94.65 & 46.05 & 57.88   & 11.31 & 13.56 & 61.39 & 62.41 & 19.33 & 60.37   & 1.72  & 5.81  & 65.97 & 88.96 & 26.00 & 61.04   & 6.47  & 7.24  & 57.57 & 59.46 & 24.28 & 63.74\\
800  & 7.68  & 22.08 & 80.22 & 91.89 & 41.97 & 44.93  & 10.65 & 13.11 & 57.83 & 59.62 & 18.42 & 47.97  & 1.33  & 3.46  & 62.26 & 85.06 & 23.73 & 48.93  & 6.44  & 7.24  & 53.60 & 56.45 & 23.17 & 52.48\\
1000 & 6.57  & 15.14 & 75.95 & 87.71 & 38.13 & 33.49  & 10.23 & 12.83 & 54.67 & 57.22 & 17.64 & 36.84  & 1.07  & 2.13  & 58.04 & 78.98 & 21.75 & 37.94  & 6.67  & 7.49  & 50.47 & 53.68 & 22.27 & 42.11\\
1200 & 5.64  & 10.44 & 71.41 & 82.43 & 34.66 & 24.20  & 10.05 & 12.74 & 52.30 & 55.32 & 16.98 & 27.72  & 0.90  & 1.48  & 54.52 & 71.60 & 19.98 & 28.51  & 7.13  & 7.97  & 48.31 & 51.47 & 21.46 & 33.02\\
1400 & 4.86  & 7.46  & 66.94 & 76.88 & 31.51 & 17.24  & 10.09 & 12.79 & 50.84 & 54.03 & 16.41 & 20.82  & 0.81  & 1.15  & 51.21 & 63.38 & 18.37 & 20.96  & 7.90  & 8.81  & 47.03 & 49.90 & 20.72 & 25.65\\
1600 & 4.25  & 5.59  & 63.48 & 71.81 & 28.69 & 12.27  & 10.28 & 12.99 & 50.20 & 53.37 & 15.95 & 15.84  & 0.76  & 0.99  & 49.37 & 56.65 & 16.90 & 15.20  & 8.81  & 9.77  & 46.88 & 49.36 & 20.05 & 19.99\\
1800 & 3.77  & 4.37  & 60.59 & 67.37 & 26.21 & 8.82   & 10.57 & 13.26 & 50.11 & 53.05 & 15.56 & 12.37  & 0.74  & 0.91  & 48.38 & 50.96 & 15.60 & 10.87  & 9.69  & 10.70 & 47.67 & 49.45 & 19.42 & 15.77\\
2000 & 3.38  & 3.54  & 58.38 & 63.85 & 24.04 & 6.52   & 10.86 & 13.54 & 50.05 & 52.83 & 15.24 & 10.04  & 0.74  & 0.86  & 47.53 & 45.64 & 14.42 & 7.73  & 10.46 & 11.51 & 48.06 & 49.45 & 18.89 & 12.70\\
\hline
\end{tabular}
}
\centering
  \end{tabular}
\end{adjustbox}
\label{tab:faithfulness_caml}
\end{table*}

\begin{table*}[!htbp]
\centering
\begin{adjustbox}{angle=270}
\begin{tabular}{@{} m{0.90\textheight} m{3.3cm} @{}}
    \captionof{table}{Faithfulness across Top-P and Top-N thresholds for LAAT.}
\resizebox{23cm}{6cm}{%
\begin{tabular}{c|cccccc|cccccc|cccccc|cccccc}
\hline
\textbf{Dataset} & \multicolumn{6}{c}{\textbf{MIMIC-IV ICD10 Full}} & \multicolumn{6}{c}{\textbf{MIMIC-IV ICD10 Top-50}} & \multicolumn{6}{c}{\textbf{MIMIC-III ICD9 Full}} &  \multicolumn{6}{c}{\textbf{MIMIC-III ICD9 Top-50}}\\
\hline
\textbf{Threshold} & \textbf{F1-Mac} & \textbf{F1-Mic} & \textbf{AUC-Mac} & \textbf{AUC-Mic} & \textbf{P@8} & \textbf{Retention} & \textbf{F1-Mac} & \textbf{F1-Mic} & \textbf{AUC-Mac} & \textbf{AUC-Mic} & \textbf{P@5} & \textbf{Retention} & \textbf{F1-Mac} & \textbf{F1-Mic} & \textbf{AUC-Mac} & \textbf{AUC-Mic} & \textbf{P@8} & \textbf{Retention} & \textbf{F1-Mac} & \textbf{F1-Mic} & \textbf{AUC-Mac} & \textbf{AUC-Mic} & \textbf{P@5} & \textbf{Retention}\\
\hline
- & 20.88 & 57.86 & 95.25 & 99.02 & 68.98 & 100.00 & 67.50 & 72.93 & 92.76 & 95.22 & 64.68 & 100.00 & 16.89 & 56.53 & 89.94 & 98.59 & 74.35 & 100.00 & 59.51 & 67.58 & 90.27 & 93.06 & 64.23 & 100.00\\
\hline
\multicolumn{25}{c}{\textbf{Top-p Sufficiency}} \\
\hdashline
10\% & 11.45 & 44.53 & 86.80 & 95.28 & 58.37 & 11.27 & 54.99 & 62.98 & 85.39 & 88.63 & 46.10 & 15.52 & 11.37 & 46.26 & 85.83 & 97.08 & 64.55 & 10.21 & 52.02 & 58.64 & 83.99 & 87.42 & 46.96 & 13.64\\
20\% & 13.25 & 47.59 & 91.60 & 97.41 & 60.88 & 21.13 & 56.29 & 64.03 & 87.06 & 90.26 & 46.77 & 24.90 & 12.50 & 48.08 & 88.33 & 98.03 & 66.49 & 20.19 & 52.83 & 60.24 & 86.34 & 89.58 & 48.04 & 23.24\\
30\% & 14.50 & 49.56 & 93.32 & 98.26 & 62.45 & 30.99 & 57.52 & 65.06 & 88.26 & 91.31 & 47.26 & 34.29 & 13.32 & 49.43 & 89.27 & 98.33 & 67.78 & 30.17 & 53.69 & 61.36 & 87.34 & 90.41 & 48.71 & 32.83\\
40\% & 15.41 & 50.97 & 94.22 & 98.67 & 63.57 & 40.85 & 58.62 & 65.98 & 89.18 & 92.10 & 47.68 & 43.68 & 13.93 & 50.48 & 89.71 & 98.46 & 68.73 & 40.14 & 54.43 & 62.26 & 87.90 & 90.93 & 49.18 & 42.43\\
50\% & 16.13 & 52.03 & 94.77 & 98.86 & 64.41 & 50.71 & 59.59 & 66.80 & 89.89 & 92.75 & 48.05 & 53.07 & 14.38 & 51.28 & 89.82 & 98.51 & 69.45 & 50.12 & 55.01 & 62.94 & 88.42 & 91.41 & 49.55 & 52.02\\
60\% & 19.97 & 56.98 & 95.00 & 98.94 & 68.21 & 60.56 & 64.36 & 70.85 & 90.47 & 93.27 & 50.09 & 62.45 & 14.76 & 51.96 & 89.88 & 98.55 & 70.07 & 60.09 & 55.61 & 63.56 & 88.72 & 91.67 & 49.83 & 61.62\\
70\% & 20.26 & 57.19 & 95.14 & 98.99 & 68.42 & 70.42 & 64.75 & 71.14 & 90.98 & 93.70 & 50.32 & 71.84 & 15.06 & 52.51 & 89.99 & 98.58 & 70.59 & 70.07 & 56.09 & 64.05 & 88.95 & 91.90 & 50.03 & 71.21\\
80\% & 20.42 & 57.35 & 95.21 & 99.01 & 68.56 & 80.28 & 65.12 & 71.40 & 91.42 & 94.07 & 50.50 & 81.23 & 15.31 & 52.98 & 89.97 & 98.59 & 71.02 & 80.05 & 56.48 & 64.46 & 89.16 & 92.10 & 50.20 & 80.81\\
90\% & 20.54 & 57.47 & 95.24 & 99.02 & 68.66 & 90.14 & 65.44 & 71.62 & 91.83 & 94.42 & 50.68 & 90.61 & 15.52 & 53.36 & 89.94 & 98.59 & 71.38 & 90.02 & 56.78 & 64.77 & 89.45 & 92.38 & 50.36 & 90.40\\
\hline
\multicolumn{25}{c}{\textbf{Top-p Comprehensiveness}} \\
\hdashline
10\% & 5.90 & 14.43 & 85.66 & 91.47 & 21.73 & 90.14 & 15.77 & 20.19 & 69.27 & 70.31 & 25.00 & 90.61 & 2.39 & 7.33 & 80.53 & 92.92 & 23.70 & 90.02 & 7.68 & 7.96 & 62.84 & 63.57 & 19.37 & 90.41\\
20\% & 4.34 & 10.77 & 77.83 & 83.95 & 17.50 & 80.28 & 15.22 & 19.52 & 66.17 & 68.42 & 24.11 & 81.23 & 1.88 & 5.67 & 73.83 & 87.92 & 19.30 & 80.05 & 8.41 & 8.97 & 59.67 & 61.86 & 18.67 & 80.81\\
30\% & 3.27 & 7.78  & 72.15 & 77.07 & 14.34 & 70.42 & 14.71 & 18.97 & 65.25 & 67.92 & 23.54 & 71.84 & 1.51 & 4.65 & 68.47 & 83.20 & 16.41 & 70.07 & 8.89 & 9.68 & 56.78 & 59.30 & 17.97 & 71.21\\
40\% & 2.56 & 5.65  & 67.33 & 71.21 & 11.99 & 60.56 & 14.24 & 18.45 & 63.84 & 66.99 & 23.01 & 62.45 & 1.25 & 3.89 & 64.77 & 78.88 & 14.28 & 60.09 & 9.18 & 10.32 & 54.80 & 57.49 & 17.38 & 61.62\\
50\% & 2.06 & 4.21  & 63.59 & 66.04 & 10.22 & 50.70 & 13.77 & 17.94 & 62.40 & 65.94 & 22.50 & 53.07 & 1.08 & 3.35 & 60.95 & 75.10 & 12.68 & 50.12 & 9.54 & 11.06 & 53.46 & 56.25 & 16.95 & 52.02\\
60\% & 1.71 & 3.23  & 60.37 & 61.71 & 8.88  & 40.85 & 10.87 & 15.05 & 61.12 & 64.87 & 19.57 & 43.68 & 0.95 & 2.91 & 58.32 & 72.20 & 11.46 & 40.14 & 9.97 & 11.90 & 52.96 & 55.64 & 16.67 & 42.43\\
70\% & 1.45 & 2.58  & 57.71 & 58.40 & 7.85  & 30.99 & 10.38 & 14.56 & 59.83 & 63.70 & 19.11 & 34.29 & 0.84 & 2.57 & 55.99 & 69.88 & 10.56 & 30.17 & 10.42 & 12.78 & 52.18 & 55.13 & 16.42 & 32.83\\
80\% & 1.25 & 2.12  & 55.03 & 55.81 & 7.05  & 21.13 & 9.92  & 14.14 & 58.43 & 62.36 & 18.59 & 24.90 & 0.76 & 2.32 & 54.76 & 68.51 & 9.91 & 20.19 & 11.03 & 13.84 & 52.02 & 55.36 & 16.28 & 23.24\\
90\% & 1.10 & 1.80  & 52.68 & 53.55 & 6.43  & 11.27 & 9.44  & 13.72 & 56.58 & 60.63 & 18.11 & 15.52 & 0.69 & 2.12 & 53.34 & 68.02 & 9.54 & 10.21 & 11.82 & 14.94 & 51.16 & 55.25 & 16.11 & 13.64\\
\hline
\hline
\multicolumn{25}{c}{\textbf{Top-N Sufficiency}} \\
\hdashline
200  & 12.84 & 46.95 & 88.98 & 96.19 & 60.47 & 15.89 & 55.77 & 63.64 & 86.06 & 89.36 & 46.58 & 19.41 & 11.75 & 46.78 & 86.56 & 97.33 & 65.59 & 13.54 & 52.11 & 59.00 & 84.54 & 87.84 & 46.93 & 16.57\\
400  & 14.54 & 49.51 & 92.65 & 97.88 & 62.66 & 30.15 & 57.42 & 65.03 & 87.69 & 90.96 & 47.31 & 32.54 & 13.01 & 48.65 & 88.77 & 98.15 & 67.36 & 26.75 & 53.35 & 60.86 & 86.52 & 89.81 & 48.23 & 28.97\\
600  & 15.64 & 51.16 & 93.94 & 98.51 & 64.01 & 44.07 & 58.77 & 66.18 & 88.87 & 92.00 & 47.87 & 45.44 & 13.77 & 49.93 & 89.43 & 98.38 & 68.50 & 39.64 & 54.18 & 61.93 & 87.33 & 90.57 & 48.93 & 41.06\\
800  & 16.44 & 52.31 & 94.58 & 98.78 & 64.92 & 57.14 & 59.90 & 67.12 & 89.85 & 92.84 & 48.35 & 57.75 & 14.30 & 50.89 & 89.81 & 98.49 & 69.41 & 51.86 & 54.89 & 62.76 & 87.86 & 91.08 & 49.39 & 52.55\\
1000 & 17.05 & 53.17 & 94.91 & 98.91 & 65.59 & 68.68 & 60.85 & 67.89 & 90.64 & 93.48 & 48.75 & 68.84 & 14.65 & 51.64 & 89.88 & 98.54 & 70.12 & 62.96 & 55.46 & 63.40 & 88.28 & 91.46 & 49.74 & 63.13\\
1200 & 17.52 & 53.83 & 95.08 & 98.96 & 66.08 & 78.02 & 61.65 & 68.51 & 91.22 & 93.98 & 49.11 & 77.97 & 14.97 & 52.25 & 89.98 & 98.57 & 70.66 & 72.47 & 55.94 & 63.91 & 88.74 & 91.79 & 50.02 & 72.37\\
1400 & 17.90 & 54.35 & 95.16 & 98.99 & 66.46 & 85.01 & 62.33 & 69.03 & 91.67 & 94.35 & 49.41 & 84.89 & 15.21 & 52.73 & 89.98 & 98.58 & 71.09 & 80.07 & 56.35 & 64.32 & 89.05 & 92.06 & 50.25 & 79.90\\
1600 & 18.21 & 54.75 & 95.21 & 99.01 & 66.76 & 90.01 & 62.90 & 69.45 & 92.01 & 94.62 & 49.67 & 89.89 & 15.41 & 53.13 & 89.97 & 98.59 & 71.46 & 85.87 & 56.68 & 64.66 & 89.36 & 92.33 & 50.44 & 85.69\\
1800 & 18.47 & 55.08 & 95.23 & 99.02 & 67.00 & 93.48 & 63.39 & 69.81 & 92.27 & 94.82 & 49.88 & 93.38 & 15.57 & 53.47 & 89.95 & 98.59 & 71.76 & 90.23 & 56.96 & 64.95 & 89.60 & 92.52 & 50.59 & 89.99\\
2000 & 18.68 & 55.36 & 95.24 & 99.02 & 67.20 & 95.79 & 63.80 & 70.11 & 92.44 & 94.96 & 50.07 & 95.73 & 15.70 & 53.76 & 89.98 & 98.59 & 72.01 & 93.39 & 57.20 & 65.19 & 89.83 & 92.69 & 50.73 & 93.11\\
\hline
\multicolumn{25}{c}{\textbf{Top-N Comprehensiveness}} \\
\hdashline
200  & 4.72 & 11.97 & 81.88 & 88.43 & 18.72 & 85.59 & 15.43 & 19.69 & 67.47 & 69.24 & 24.55 & 86.79 & 2.18 & 6.93 & 78.28 & 91.80 & 21.51 & 86.70 & 7.80  & 8.17  & 59.45 & 61.31 & 18.78 & 87.54\\
400  & 3.18 & 7.68  & 72.90 & 78.43 & 14.45 & 71.33 & 14.84 & 18.98 & 64.41 & 67.36 & 23.64 & 73.66 & 1.72 & 5.39 & 70.28 & 85.41 & 17.43 & 73.48 & 8.52  & 9.12  & 55.56 & 58.15 & 17.72 & 75.14\\
600  & 2.23 & 5.03  & 66.68 & 70.84 & 11.59 & 57.42 & 14.27 & 18.34 & 62.68 & 66.01 & 22.86 & 60.75 & 1.38 & 4.31 & 65.42 & 80.03 & 14.75 & 60.60 & 8.74  & 9.71  & 53.09 & 56.02 & 16.94 & 63.05\\
800  & 1.68 & 3.50  & 61.99 & 64.58 & 9.61  & 44.34 & 13.68 & 17.67 & 60.24 & 63.85 & 22.11 & 48.44 & 1.13 & 3.57 & 61.46 & 75.33 & 12.79 & 48.38 & 9.05  & 10.45 & 51.28 & 54.31 & 16.46 & 51.56\\
1000 & 1.32 & 2.56  & 58.09 & 59.34 & 8.21  & 32.80 & 13.20 & 17.15 & 57.50 & 61.26 & 21.41 & 37.36 & 0.94 & 2.97 & 57.55 & 70.94 & 11.42 & 37.29 & 9.65  & 11.40 & 50.56 & 53.52 & 16.23 & 40.99\\
1200 & 1.08 & 1.96  & 55.20 & 55.40 & 7.16  & 23.46 & 12.88 & 16.78 & 54.74 & 58.60 & 20.74 & 28.23 & 0.79 & 2.49 & 54.98 & 67.07 & 10.43 & 27.79 & 10.42 & 12.42 & 49.43 & 52.42 & 16.03 & 31.74\\
1400 & 0.92 & 1.58  & 53.29 & 52.91 & 6.37  & 16.47 & 12.73 & 16.62 & 52.94 & 56.87 & 20.15 & 21.30 & 0.69 & 2.14 & 52.21 & 64.16 & 9.76  & 20.20 & 11.30 & 13.45 & 48.94 & 51.89 & 15.85 & 24.21\\
1600 & 0.81 & 1.34  & 51.77 & 51.57 & 5.74  & 11.47 & 12.77 & 16.64 & 52.16 & 56.09 & 19.64 & 16.30 & 0.62 & 1.87 & 50.90 & 61.69 & 9.25  & 14.41 & 12.20 & 14.42 & 48.94 & 51.60 & 15.71 & 18.42\\
1800 & 0.72 & 1.16  & 51.17 & 50.82 & 5.24  & 8.01 & 12.90 & 16.77 & 51.97 & 55.86 & 19.20 & 12.81 & 0.56 & 1.67 & 50.13 & 60.20 & 8.88  & 10.06 & 13.03 & 15.25 & 49.16 & 51.61 & 15.56 & 14.12\\
2000 & 0.66 & 1.04  & 50.79 & 50.41 & 4.84  & 5.69 & 13.09 & 16.98 & 52.10 & 56.01 & 18.82 & 10.46 & 0.52 & 1.52 & 49.07 & 58.83 & 8.58  & 6.90 & 13.76 & 15.96 & 49.13 & 51.34 & 15.40 & 11.00\\
\hline
\end{tabular}
}
\centering
  \end{tabular}
\end{adjustbox}
\label{tab:faithfulness_laat}
\end{table*}

\begin{table*}[!htbp]
\centering
\begin{adjustbox}{angle=270}
\begin{tabular}{@{} m{0.90\textheight} m{3.3cm} @{}}
    \captionof{table}{Faithfulness across Top-P and Top-N thresholds for PLM-ICD.}
\resizebox{23cm}{6cm}{%
\begin{tabular}{c|cccccc|cccccc|cccccc|cccccc}
\hline
\textbf{Dataset} & \multicolumn{6}{c}{\textbf{MIMIC-IV ICD10 Full}} & \multicolumn{6}{c}{\textbf{MIMIC-IV ICD10 Top-50}} & \multicolumn{6}{c}{\textbf{MIMIC-III ICD9 Full}} &  \multicolumn{6}{c}{\textbf{MIMIC-III ICD9 Top-50}}\\
\hline
\textbf{Threshold} & \textbf{F1-Mac} & \textbf{F1-Mic} & \textbf{AUC-Mac} & \textbf{AUC-Mic} & \textbf{P@8} & \textbf{Retention} & \textbf{F1-Mac} & \textbf{F1-Mic} & \textbf{AUC-Mac} & \textbf{AUC-Mic} & \textbf{P@5} & \textbf{Retention} & \textbf{F1-Mac} & \textbf{F1-Mic} & \textbf{AUC-Mac} & \textbf{AUC-Mic} & \textbf{P@8} & \textbf{Retention} & \textbf{F1-Mac} & \textbf{F1-Mic} & \textbf{AUC-Mac} & \textbf{AUC-Mic} & \textbf{P@5} & \textbf{Retention}\\
\hline
- & 19.70 & 58.32 & 96.54 & 99.23 & 69.79 & 100.00 & 68.20 & 73.42 & 93.30 & 95.50 & 65.35 & 100.00 & 16.20 & 56.33 & 91.39 & 98.75 & 73.10 & 100.00 & 65.70 & 71.18 & 91.96 & 94.26 & 66.65 & 100.00\\
\hline
\multicolumn{25}{c}{\textbf{Top-p Sufficiency}} \\
\hdashline
10\%  & 6.44  & 47.05 & 85.04 & 95.21 & 62.86 & 11.35 & 63.49 & 70.12 & 89.09 & 91.99 & 49.64 & 14.97 & 12.34 & 46.90 & 79.79 & 95.24 & 66.51 & 10.37 & 58.83 & 66.03 & 86.02 & 88.14 & 49.45 & 12.13\\
20\%  & 7.63  & 48.99 & 90.07 & 96.94 & 63.98 & 21.20 & 64.37 & 70.62 & 91.10 & 93.70 & 50.18 & 24.42 & 13.74 & 47.72 & 84.76 & 96.75 & 67.03 & 20.33 & 61.03 & 67.52 & 88.34 & 90.80 & 50.66 & 21.89\\
30\%  & 8.44  & 50.22 & 92.80 & 97.79 & 64.69 & 31.05 & 65.07 & 71.06 & 92.16 & 94.56 & 50.60 & 33.87 & 14.47 & 48.70 & 87.65 & 97.73 & 67.62 & 30.29 & 61.95 & 68.18 & 89.24 & 91.75 & 51.23 & 31.66\\
40\%  & 8.99  & 51.10 & 94.25 & 98.27 & 65.24 & 40.90 & 65.56 & 71.40 & 92.67 & 94.97 & 50.89 & 43.31 & 15.00 & 49.58 & 89.16 & 98.19 & 68.11 & 40.25 & 62.59 & 68.62 & 90.13 & 92.55 & 51.66 & 41.42\\
50\%  & 10.98 & 54.48 & 95.11 & 98.55 & 67.60 & 50.75 & 67.42 & 72.76 & 92.91 & 95.19 & 51.99 & 52.76 & 15.36 & 50.34 & 89.79 & 98.42 & 68.56 & 50.21 & 63.05 & 68.92 & 90.62 & 93.16 & 52.01 & 51.19\\
60\%  & 11.40 & 54.85 & 95.70 & 98.76 & 67.90 & 60.60 & 67.54 & 72.87 & 93.09 & 95.33 & 52.08 & 62.21 & 15.54 & 51.02 & 90.56 & 98.59 & 69.04 & 60.17 & 63.37 & 69.17 & 90.95 & 93.51 & 52.27 & 60.95\\
70\%  & 11.90 & 55.27 & 96.06 & 98.92 & 68.19 & 70.45 & 67.66 & 72.98 & 93.20 & 95.42 & 52.13 & 71.66 & 15.66 & 51.60 & 90.87 & 98.67 & 69.48 & 70.12 & 63.62 & 69.36 & 91.33 & 93.84 & 52.48 & 70.71\\
80\%  & 12.62 & 55.70 & 96.31 & 99.05 & 68.45 & 80.30 & 67.76 & 73.06 & 93.24 & 95.46 & 52.18 & 81.10 & 15.76 & 52.07 & 91.19 & 98.72 & 69.85 & 80.08 & 63.85 & 69.56 & 91.54 & 93.95 & 52.65 & 80.48\\
90\%  & 13.60 & 56.14 & 96.45 & 99.16 & 68.68 & 90.15 & 67.84 & 73.13 & 93.27 & 95.48 & 52.22 & 90.55 & 15.82 & 52.48 & 91.27 & 98.73 & 70.19 & 90.04 & 64.01 & 69.69 & 91.85 & 94.15 & 52.79 & 90.24\\
\hline
\multicolumn{25}{c}{\textbf{Top-p Comprehensiveness}} \\
\hdashline
10\% & 7.30 & 24.37 & 91.34 & 96.93 & 37.00 & 90.15 & 15.25 & 16.14 & 73.90 & 73.50 & 24.12 & 90.55 & 10.40 & 32.43 & 87.87 & 97.42 & 46.89 & 90.04 & 38.46 & 41.69 & 83.08 & 84.77 & 40.01 & 90.24\\
20\% & 5.36 & 18.05 & 83.41 & 93.87 & 30.82 & 80.30 & 12.16 & 12.73 & 66.44 & 66.82 & 21.74 & 81.10 & 8.47  & 25.96 & 80.34 & 95.11 & 39.04 & 80.08 & 32.28 & 34.74 & 78.77 & 79.37 & 36.65 & 80.48\\
30\% & 4.26 & 14.22 & 76.66 & 91.42 & 27.09 & 70.45 & 10.16 & 10.54 & 61.77 & 63.63 & 20.44 & 71.66 & 7.01  & 21.37 & 73.10 & 93.16 & 33.48 & 70.12 & 27.32 & 29.38 & 75.10 & 74.74 & 33.91 & 70.71\\
40\% & 3.53 & 11.69 & 70.78 & 89.94 & 24.74 & 60.60 & 8.70 & 8.96 & 58.63 & 61.90 & 19.64 & 62.21 & 5.99  & 18.10 & 66.40 & 91.71 & 29.53 & 60.17 & 23.54 & 25.31 & 72.09 & 71.52 & 31.89 & 60.95\\
50\% & 3.02 & 9.90 & 65.45 & 89.12 & 23.24 & 50.75 & 7.58 & 7.78 & 57.07 & 61.26 & 19.15 & 52.76 & 5.23  & 15.69 & 62.50 & 89.90 & 26.54 & 50.21 & 20.60 & 22.17 & 69.27 & 68.95 & 30.31 & 51.19\\
60\% & 2.64 & 8.56 & 60.53 & 88.62 & 22.35 & 40.90 & 6.71 & 6.86 & 56.82 & 61.15 & 18.82 & 43.31 & 4.64  & 13.77 & 59.32 & 89.89 & 24.49 & 40.25 & 18.33 & 19.74 & 67.22 & 67.14 & 29.10 & 41.42\\
70\% & 2.35 & 7.52 & 56.61 & 88.29 & 21.83 & 31.05 & 6.01 & 6.13 & 57.30 & 61.09 & 18.55 & 33.87 & 4.18  & 12.25 & 58.10 & 89.60 & 23.03 & 30.29 & 16.48 & 17.75 & 65.25 & 65.53 & 28.12 & 31.66\\
80\% & 2.12 & 6.69 & 53.49 & 88.08 & 21.56 & 21.20 & 5.44 & 5.53 & 57.41 & 60.70 & 18.29 & 24.42 & 3.79  & 11.01 & 55.17 & 87.56 & 21.89 & 20.33 & 14.92 & 16.05 & 63.51 & 64.04 & 27.23 & 21.89\\
90\% & 1.94 & 6.02 & 51.72 & 87.96 & 21.39 & 11.35 & 4.97 & 5.04 & 56.85 & 60.06 & 18.02 & 14.97 & 3.47  & 9.99  & 55.20 & 86.69 & 20.98 & 10.37 & 13.60 & 14.62 & 60.28 & 62.37 & 26.39 & 12.13\\
\hline
\hline
\multicolumn{25}{c}{\textbf{Top-N Sufficiency}} \\
\hdashline
200  & 6.68  & 46.56 & 82.04 & 94.81 & 63.10 & 13.21 & 63.51 & 70.14 & 88.90 & 91.86 & 49.70 & 16.38 & 12.20 & 47.59 & 80.72 & 95.53 & 66.93 & 11.42 & 57.96 & 65.62 & 84.96 & 87.67 & 49.35 & 12.92\\
400  & 7.93  & 48.63 & 87.60 & 96.50 & 64.23 & 24.86 & 64.36 & 70.64 & 91.01 & 93.62 & 50.22 & 27.21 & 13.81 & 48.75 & 85.80 & 97.19 & 67.70 & 22.39 & 60.40 & 67.27 & 88.02 & 90.56 & 50.46 & 23.43\\
600  & 8.69  & 49.90 & 90.89 & 97.38 & 64.97 & 36.39 & 64.99 & 71.05 & 92.05 & 94.46 & 50.61 & 37.96 & 14.54 & 49.46 & 87.83 & 97.70 & 68.10 & 33.25 & 61.59 & 68.05 & 89.17 & 91.68 & 51.15 & 33.89\\
800  & 9.39  & 50.82 & 92.64 & 97.92 & 65.54 & 47.62 & 65.45 & 71.35 & 92.56 & 94.88 & 50.89 & 48.52 & 15.00 & 50.17 & 88.95 & 98.15 & 68.54 & 43.83 & 62.20 & 68.43 & 89.49 & 92.17 & 51.53 & 44.03\\
1000 & 10.07 & 51.57 & 93.58 & 98.26 & 66.03 & 58.23 & 65.82 & 71.61 & 92.87 & 95.14 & 51.11 & 58.61 & 15.34 & 50.78 & 89.56 & 98.34 & 68.92 & 53.95 & 62.67 & 68.76 & 90.26 & 92.90 & 51.87 & 53.78\\
1200 & 10.77 & 52.19 & 94.08 & 98.46 & 66.43 & 67.72 & 66.12 & 71.83 & 93.04 & 95.28 & 51.27 & 67.79 & 15.59 & 51.35 & 90.18 & 98.52 & 69.32 & 63.27 & 63.00 & 68.97 & 90.60 & 93.26 & 52.13 & 62.87\\
1400 & 11.48 & 52.73 & 94.46 & 98.63 & 66.77 & 75.87 & 66.35 & 72.00 & 93.12 & 95.36 & 51.40 & 75.77 & 15.76 & 51.83 & 90.58 & 98.60 & 69.69 & 71.59 & 63.30 & 69.18 & 90.85 & 93.47 & 52.36 & 71.06\\
1600 & 12.17 & 53.20 & 94.77 & 98.75 & 67.06 & 82.28 & 66.55 & 72.14 & 93.18 & 95.41 & 51.51 & 82.10 & 15.88 & 52.24 & 90.84 & 98.66 & 70.01 & 78.51 & 63.53 & 69.36 & 91.12 & 93.71 & 52.53 & 78.05\\
1800 & 12.81 & 53.62 & 95.07 & 98.86 & 67.30 & 87.34 & 66.72 & 72.27 & 93.24 & 95.45 & 51.60 & 87.17 & 15.97 & 52.61 & 91.05 & 98.70 & 70.29 & 84.22 & 63.73 & 69.53 & 91.29 & 93.85 & 52.69 & 83.80\\
2000 & 13.40 & 54.00 & 95.35 & 98.94 & 67.52 & 91.04 & 66.85 & 72.37 & 93.27 & 95.48 & 51.67 & 90.90 & 16.03 & 52.92 & 91.17 & 98.72 & 70.54 & 88.72 & 63.91 & 69.68 & 91.49 & 93.95 & 52.82 & 88.40\\
\hline
\multicolumn{25}{c}{\textbf{Top-N Comprehensiveness}} \\
\hdashline
200  & 7.62 & 25.40 & 86.84 & 96.09 & 36.33 & 88.29 & 15.97 & 16.63 & 73.93 & 73.84 & 24.28 & 89.14 & 10.08 & 32.48 & 85.62 & 96.80 & 45.67 & 88.99 & 40.28 & 43.28 & 83.21 & 84.77 & 40.15 & 89.45 \\
400  & 5.78 & 19.40 & 77.83 & 93.40 & 30.79 & 76.64 & 12.82 & 13.18 & 66.56 & 67.30 & 22.02 & 78.31 & 8.43  & 26.82 & 77.48 & 94.37 & 38.55 & 78.02 & 33.70 & 36.13 & 79.05 & 79.66 & 36.90 & 78.94 \\
600  & 4.63 & 15.58 & 71.39 & 91.60 & 27.45 & 65.12 & 10.76 & 10.96 & 61.95 & 64.05 & 20.75 & 67.56 & 7.14  & 22.64 & 71.13 & 92.77 & 33.69 & 67.16 & 28.81 & 30.85 & 75.42 & 75.48 & 34.27 & 68.48 \\
800  & 3.86 & 12.96 & 65.87 & 90.45 & 25.38 & 53.89 & 9.29  & 9.38  & 58.48 & 62.18 & 19.93 & 57.01 & 6.19  & 19.50 & 66.61 & 91.73 & 30.19 & 56.59 & 25.24 & 26.91 & 72.29 & 72.17 & 32.33 & 58.34 \\
1000 & 3.31 & 11.06 & 62.46 & 89.71 & 24.04 & 43.29 & 8.16  & 8.20  & 56.39 & 61.22 & 19.37 & 46.91 & 5.44  & 17.09 & 63.23 & 91.21 & 27.64 & 46.48 & 22.32 & 23.82 & 69.51 & 69.70 & 30.77 & 48.59 \\
1200 & 2.90 & 9.63  & 59.59 & 89.23 & 23.18 & 33.79 & 7.29  & 7.30  & 55.73 & 60.91 & 18.97 & 37.74 & 4.88  & 15.17 & 61.40 & 90.65 & 25.64 & 37.15 & 20.03 & 21.40 & 67.39 & 67.87 & 29.59 & 39.50 \\
1400 & 2.59 & 8.51  & 57.53 & 88.87 & 22.59 & 25.66 & 6.57  & 6.56  & 55.54 & 60.65 & 18.64 & 29.78 & 4.40  & 13.61 & 59.70 & 90.39 & 24.08 & 28.85 & 18.15 & 19.40 & 64.78 & 66.07 & 28.57 & 31.31 \\
1600 & 2.34 & 7.61  & 56.37 & 88.60 & 22.19 & 19.26 & 5.98  & 5.96  & 56.16 & 60.63 & 18.37 & 23.44 & 4.01  & 12.32 & 58.11 & 89.83 & 22.84 & 21.93 & 16.54 & 17.70 & 62.89 & 64.86 & 27.67 & 24.33 \\
1800 & 2.13 & 6.88  & 55.84 & 88.40 & 21.92 & 14.20 & 5.49  & 5.46  & 56.70 & 60.48 & 18.14 & 18.39 & 3.68  & 11.23 & 56.84 & 89.46 & 21.86 & 16.23 & 15.20 & 16.28 & 61.31 & 63.78 & 26.92 & 18.57 \\
2000 & 1.97 & 6.27  & 54.88 & 88.26 & 21.72 & 10.5 & 5.08  & 5.04  & 57.10 & 60.41 & 17.93 & 14.66 & 3.40  & 10.30 & 56.25 & 89.02 & 21.01 & 11.73 & 14.05 & 15.04 & 59.85 & 63.07 & 26.27 & 13.97 \\
\hline
\end{tabular}
}
\centering
  \end{tabular}
\end{adjustbox}
\label{tab:faithfulness_plmicd}
\end{table*}

\begin{table*}[!htbp]
\centering
\hspace*{-7cm}
\begin{adjustbox}{angle=270,center}
\begin{minipage}{1.0\textheight}  % Use textheight since we're rotating
\centering
\caption{Plausibility results of CAML across all thresholds.}
\label{tab:plausibility_caml}

\begin{minipage}[t][0.8\textheight][t]{0.48\textwidth}
\centering
\subcaption{Results using the Top‑p selection strategy.}
\scriptsize
\begin{tabular}{lrrrrrrrrr}
\toprule
Metric & Threshold & \#Prd & \#Act & TP & FP & FN & Pr & Rc & F1 \\
\midrule
\multirow{9}{*}{Exact SM} & 0.1 & 84520 & 2269 & 55 & 84465 & 2214 & 0.1\% & 2.4\% & 0.1\% \\
& 0.2 & 146091 & 2269 & 44 & 146047 & 2225 & 0.0\% & 1.9\% & 0.1\% \\
& 0.3 & 189001 & 2269 & 43 & 188958 & 2226 & 0.0\% & 1.9\% & 0.0\% \\
& 0.4 & 215458 & 2269 & 32 & 215426 & 2237 & 0.0\% & 1.4\% & 0.0\% \\
& 0.5 & 225795 & 2269 & 29 & 225766 & 2240 & 0.0\% & 1.3\% & 0.0\% \\
& 0.6 & 220015 & 2269 & 24 & 219991 & 2245 & 0.0\% & 1.1\% & 0.0\% \\
& 0.7 & 195599 & 2269 & 14 & 195585 & 2255 & 0.0\% & 0.6\% & 0.0\% \\
& 0.8 & 154368 & 2269 & 8 & 154360 & 2261 & 0.0\% & 0.4\% & 0.0\% \\
& 0.9 & 93383 & 2269 & 3 & 93380 & 2266 & 0.0\% & 0.1\% & 0.0\% \\
\hline
\multirow{9}{*}{PI SM} & 0.1 & 77970 & 2021 & 88 & 77882 & 1933 & 0.1\% & 4.4\% & 0.2\% \\
& 0.2 & 133104 & 2021 & 91 & 133013 & 1930 & 0.1\% & 4.5\% & 0.1\% \\
& 0.3 & 172526 & 2021 & 96 & 172430 & 1925 & 0.1\% & 4.8\% & 0.1\% \\
& 0.4 & 198268 & 2021 & 85 & 198183 & 1936 & 0.0\% & 4.2\% & 0.1\% \\
& 0.5 & 210205 & 2021 & 79 & 210126 & 1942 & 0.0\% & 3.9\% & 0.1\% \\
& 0.6 & 207520 & 2021 & 68 & 207452 & 1953 & 0.0\% & 3.4\% & 0.1\% \\
& 0.7 & 187097 & 2021 & 54 & 187043 & 1967 & 0.0\% & 2.7\% & 0.1\% \\
& 0.8 & 149936 & 2021 & 38 & 149898 & 1983 & 0.0\% & 1.9\% & 0.1\% \\
& 0.9 & 91959 & 2021 & 19 & 91940 & 2002 & 0.0\% & 0.9\% & 0.0\% \\
\hline
\multirow{9}{*}{Exact TM} & 0.1 & 150968 & 11428 & 2493 & 148475 & 8935 & 1.7\% & 21.8\% & 3.1\% \\
& 0.2 & 299351 & 11428 & 3522 & 295829 & 7906 & 1.2\% & 30.8\% & 2.3\% \\
& 0.3 & 445550 & 11428 & 4216 & 441334 & 7212 & 0.9\% & 36.9\% & 1.8\% \\
& 0.4 & 590911 & 11428 & 4660 & 586251 & 6768 & 0.8\% & 40.8\% & 1.5\% \\
& 0.5 & 735001 & 11428 & 5010 & 729991 & 6418 & 0.7\% & 43.8\% & 1.3\% \\
& 0.6 & 876874 & 11428 & 5347 & 871527 & 6081 & 0.6\% & 46.8\% & 1.2\% \\
& 0.7 & 1016760 & 11428 & 5619 & 1011141 & 5809 & 0.6\% & 49.2\% & 1.1\% \\
& 0.8 & 1155761 & 11428 & 6073 & 1149688 & 5355 & 0.5\% & 53.1\% & 1.0\% \\
& 0.9 & 1293460 & 11428 & 6065 & 1287395 & 5363 & 0.5\% & 53.1\% & 0.9\% \\
\hline
\multirow{9}{*}{PI TM} & 0.1 & 114766 & 9367 & 4029 & 110737 & 5338 & 3.5\% & 43.0\% & 6.5\% \\
& 0.2 & 200255 & 9367 & 5190 & 195065 & 4177 & 2.6\% & 55.4\% & 5.0\% \\
& 0.3 & 271137 & 9367 & 5931 & 265206 & 3436 & 2.2\% & 63.3\% & 4.2\% \\
& 0.4 & 331052 & 9367 & 6331 & 324721 & 3036 & 1.9\% & 67.6\% & 3.7\% \\
& 0.5 & 382967 & 9367 & 6692 & 376275 & 2675 & 1.7\% & 71.4\% & 3.4\% \\
& 0.6 & 426343 & 9367 & 6998 & 419345 & 2369 & 1.6\% & 74.7\% & 3.2\% \\
& 0.7 & 462182 & 9367 & 7197 & 454985 & 2170 & 1.6\% & 76.8\% & 3.1\% \\
& 0.8 & 490358 & 9367 & 7423 & 482935 & 1944 & 1.5\% & 79.2\% & 3.0\% \\
& 0.9 & 514078 & 9367 & 7481 & 506597 & 1886 & 1.5\% & 79.9\% & 2.9\% \\
\bottomrule
\end{tabular}
\end{minipage}
\hfill
\begin{minipage}[t][0.8\textheight][t]{0.48\textwidth}
\centering
\subcaption{Results using the Top‑N selection strategy.}
\scriptsize
\begin{tabular}{lrrrrrrrrr}
\toprule
Metric & Threshold & \#Prd & \#Act & TP & FP & FN & Pr & Rc & F1 \\
\midrule
\multirow{10}{*}{Exact SM} & 200 & 103366 & 2269 & 52 & 103314 & 2217 & 0.1\% & 2.3\% & 0.1\% \\
& 400 & 169040 & 2269 & 38 & 169002 & 2231 & 0.0\% & 1.7\% & 0.0\% \\
& 600 & 204444 & 2269 & 31 & 204413 & 2238 & 0.0\% & 1.4\% & 0.0\% \\
& 800 & 209526 & 2269 & 25 & 209501 & 2244 & 0.0\% & 1.1\% & 0.0\% \\
& 1000 & 185153 & 2269 & 22 & 185131 & 2247 & 0.0\% & 1.0\% & 0.0\% \\
& 1200 & 143952 & 2269 & 14 & 143938 & 2255 & 0.0\% & 0.6\% & 0.0\% \\
& 1400 & 102270 & 2269 & 8 & 102262 & 2261 & 0.0\% & 0.4\% & 0.0\% \\
& 1600 & 67889 & 2269 & 6 & 67883 & 2263 & 0.0\% & 0.3\% & 0.0\% \\
& 1800 & 38832 & 2269 & 5 & 38827 & 2264 & 0.0\% & 0.2\% & 0.0\% \\
& 2000 & 21988 & 2269 & 1 & 21987 & 2268 & 0.0\% & 0.0\% & 0.0\% \\
\hline
\multirow{10}{*}{PI SM} & 200 & 95197 & 2021 & 84 & 95113 & 1937 & 0.1\% & 4.2\% & 0.2\% \\
& 400 & 154719 & 2021 & 89 & 154630 & 1932 & 0.1\% & 4.4\% & 0.1\% \\
& 600 & 188430 & 2021 & 85 & 188345 & 1936 & 0.0\% & 4.2\% & 0.1\% \\
& 800 & 195157 & 2021 & 71 & 195086 & 1950 & 0.0\% & 3.5\% & 0.1\% \\
& 1000 & 174043 & 2021 & 59 & 173984 & 1962 & 0.0\% & 2.9\% & 0.1\% \\
& 1200 & 136261 & 2021 & 43 & 136218 & 1978 & 0.0\% & 2.1\% & 0.1\% \\
& 1400 & 97396 & 2021 & 28 & 97368 & 1993 & 0.0\% & 1.4\% & 0.1\% \\
& 1600 & 65016 & 2021 & 22 & 64994 & 1999 & 0.0\% & 1.1\% & 0.1\% \\
& 1800 & 37116 & 2021 & 14 & 37102 & 2007 & 0.0\% & 0.7\% & 0.1\% \\
& 2000 & 20846 & 2021 & 10 & 20836 & 2011 & 0.0\% & 0.5\% & 0.1\% \\
\hline
\multirow{10}{*}{Exact TM} & 200 & 194791 & 11428 & 2820 & 191971 & 8608 & 1.4\% & 24.7\% & 2.7\% \\
& 400 & 385838 & 11428 & 3929 & 381909 & 7499 & 1.0\% & 34.4\% & 2.0\% \\
& 600 & 573507 & 11428 & 4628 & 568879 & 6800 & 0.8\% & 40.5\% & 1.6\% \\
& 800 & 758757 & 11428 & 5183 & 753574 & 6245 & 0.7\% & 45.4\% & 1.3\% \\
& 1000 & 938428 & 11428 & 5623 & 932805 & 5805 & 0.6\% & 49.2\% & 1.2\% \\
& 1200 & 1088084 & 11428 & 5560 & 1082524 & 5868 & 0.5\% & 48.7\% & 1.0\% \\
& 1400 & 1206408 & 11428 & 5652 & 1200756 & 5776 & 0.5\% & 49.5\% & 0.9\% \\
& 1600 & 1284292 & 11428 & 5953 & 1278339 & 5475 & 0.5\% & 52.1\% & 0.9\% \\
& 1800 & 1337091 & 11428 & 5577 & 1331514 & 5851 & 0.4\% & 48.8\% & 0.8\% \\
& 2000 & 1365475 & 11428 & 5404 & 1360071 & 6024 & 0.4\% & 47.3\% & 0.8\% \\
\hline
\multirow{10}{*}{PI TM} & 200 & 142740 & 9367 & 4470 & 138270 & 4897 & 3.1\% & 47.7\% & 5.9\% \\
& 400 & 244099 & 9367 & 5692 & 238407 & 3675 & 2.3\% & 60.8\% & 4.5\% \\
& 600 & 324089 & 9367 & 6374 & 317715 & 2993 & 2.0\% & 68.0\% & 3.8\% \\
& 800 & 388125 & 9367 & 6818 & 381307 & 2549 & 1.8\% & 72.8\% & 3.4\% \\
& 1000 & 439775 & 9367 & 7095 & 432680 & 2272 & 1.6\% & 75.7\% & 3.2\% \\
& 1200 & 481563 & 9367 & 7432 & 474131 & 1935 & 1.5\% & 79.3\% & 3.0\% \\
& 1400 & 509233 & 9367 & 7615 & 501618 & 1752 & 1.5\% & 81.3\% & 2.9\% \\
& 1600 & 530370 & 9367 & 7885 & 522485 & 1482 & 1.5\% & 84.2\% & 2.9\% \\
& 1800 & 543222 & 9367 & 7988 & 535234 & 1379 & 1.5\% & 85.3\% & 2.9\% \\
& 2000 & 551266 & 9367 & 8080 & 543186 & 1287 & 1.5\% & 86.3\% & 2.9\% \\
\bottomrule
\end{tabular}
\end{minipage}
\end{minipage}
\end{adjustbox}
\end{table*}

\begin{table*}[!htbp]
\centering
\hspace*{-7cm}
\begin{adjustbox}{angle=270,center}
\begin{minipage}{1.0\textheight}  % Use textheight since we're rotating
\centering
\caption{Plausibility results of LAAT across all thresholds.}
\label{tab:plausibility_laat}

\begin{minipage}[t][0.8\textheight][t]{0.48\textwidth}  % Fixed height
\centering
\subcaption{Results using the Top‑p selection strategy.}
\scriptsize
\begin{tabular}{lrrrrrrrrr}
\toprule
Metric & Threshold & \#Prd & \#Act & TP & FP & FN & Pr & Rc & F1 \\
\midrule
\multirow{9}{*}{Exact SM} & 0.1 & 91482 & 2269 & 330 & 91152 & 1939 & 0.4\% & 14.5\% & 0.7\% \\
& 0.2 & 152195 & 2269 & 183 & 152012 & 2086 & 0.1\% & 8.1\% & 0.2\% \\
& 0.3 & 190875 & 2269 & 107 & 190768 & 2162 & 0.1\% & 4.7\% & 0.1\% \\
& 0.4 & 211973 & 2269 & 62 & 211911 & 2207 & 0.0\% & 2.7\% & 0.1\% \\
& 0.5 & 217293 & 2269 & 40 & 217253 & 2229 & 0.0\% & 1.8\% & 0.0\% \\
& 0.6 & 207522 & 2269 & 29 & 207493 & 2240 & 0.0\% & 1.3\% & 0.0\% \\
& 0.7 & 182928 & 2269 & 15 & 182913 & 2254 & 0.0\% & 0.7\% & 0.0\% \\
& 0.8 & 143185 & 2269 & 3 & 143182 & 2266 & 0.0\% & 0.1\% & 0.0\% \\
& 0.9 & 87373 & 2269 & 1 & 87372 & 2268 & 0.0\% & 0.0\% & 0.0\% \\
\hline
\multirow{9}{*}{PI SM} & 0.1 & 82542 & 2021 & 342 & 82200 & 1679 & 0.4\% & 16.9\% & 0.8\% \\
& 0.2 & 138074 & 2021 & 223 & 137851 & 1798 & 0.2\% & 11.0\% & 0.3\% \\
& 0.3 & 174708 & 2021 & 145 & 174563 & 1876 & 0.1\% & 7.2\% & 0.2\% \\
& 0.4 & 196066 & 2021 & 102 & 195964 & 1919 & 0.1\% & 5.0\% & 0.1\% \\
& 0.5 & 203071 & 2021 & 74 & 202997 & 1947 & 0.0\% & 3.7\% & 0.1\% \\
& 0.6 & 195899 & 2021 & 55 & 195844 & 1966 & 0.0\% & 2.7\% & 0.1\% \\
& 0.7 & 174663 & 2021 & 41 & 174622 & 1980 & 0.0\% & 2.0\% & 0.0\% \\
& 0.8 & 138486 & 2021 & 24 & 138462 & 1997 & 0.0\% & 1.2\% & 0.0\% \\
& 0.9 & 85688 & 2021 & 11 & 85677 & 2010 & 0.0\% & 0.5\% & 0.0\% \\
\hline
\multirow{9}{*}{Exact TM} & 0.1 & 160732 & 11428 & 4266 & 156466 & 7162 & 2.7\% & 37.3\% & 5.0\% \\
& 0.2 & 316121 & 11428 & 4840 & 311281 & 6588 & 1.5\% & 42.4\% & 3.0\% \\
& 0.3 & 467215 & 11428 & 5258 & 461957 & 6170 & 1.1\% & 46.0\% & 2.2\% \\
& 0.4 & 615384 & 11428 & 5468 & 609916 & 5960 & 0.9\% & 47.8\% & 1.7\% \\
& 0.5 & 759735 & 11428 & 5539 & 754196 & 5889 & 0.7\% & 48.5\% & 1.4\% \\
& 0.6 & 901596 & 11428 & 5961 & 895635 & 5467 & 0.7\% & 52.2\% & 1.3\% \\
& 0.7 & 1040440 & 11428 & 6142 & 1034298 & 5286 & 0.6\% & 53.7\% & 1.2\% \\
& 0.8 & 1175547 & 11428 & 6022 & 1169525 & 5406 & 0.5\% & 52.7\% & 1.0\% \\
& 0.9 & 1303770 & 11428 & 5812 & 1297958 & 5616 & 0.4\% & 50.9\% & 0.9\% \\
\hline
\multirow{9}{*}{PI TM} & 0.1 & 121292 & 9367 & 4720 & 116572 & 4647 & 3.9\% & 50.4\% & 7.2\% \\
& 0.2 & 212105 & 9367 & 5732 & 206373 & 3635 & 2.7\% & 61.2\% & 5.2\% \\
& 0.3 & 284284 & 9367 & 6234 & 278050 & 3133 & 2.2\% & 66.6\% & 4.2\% \\
& 0.4 & 344568 & 9367 & 6705 & 337863 & 2662 & 1.9\% & 71.6\% & 3.8\% \\
& 0.5 & 394117 & 9367 & 6979 & 387138 & 2388 & 1.8\% & 74.5\% & 3.5\% \\
& 0.6 & 436327 & 9367 & 7249 & 429078 & 2118 & 1.7\% & 77.4\% & 3.3\% \\
& 0.7 & 470428 & 9367 & 7420 & 463008 & 1947 & 1.6\% & 79.2\% & 3.1\% \\
& 0.8 & 498098 & 9367 & 7545 & 490553 & 1822 & 1.5\% & 80.5\% & 3.0\% \\
& 0.9 & 523255 & 9367 & 7637 & 515618 & 1730 & 1.5\% & 81.5\% & 2.9\% \\
\bottomrule
\end{tabular}
\end{minipage}
\hfill
\begin{minipage}[t][0.8\textheight][t]{0.48\textwidth}  % Fixed height
\centering
\subcaption{Results using the Top‑N selection strategy.}
\scriptsize
\begin{tabular}{lrrrrrrrrr}
\toprule
Metric & Threshold & \#Prd & \#Act & TP & FP & FN & Pr & Rc & F1 \\
\midrule
\multirow{10}{*}{Exact SM} & 200 & 109729 & 2269 & 291 & 109438 & 1978 & 0.3\% & 12.8\% & 0.5\% \\
& 400 & 171537 & 2269 & 124 & 171413 & 2145 & 0.1\% & 5.5\% & 0.1\% \\
& 600 & 200746 & 2269 & 75 & 200671 & 2194 & 0.0\% & 3.3\% & 0.1\% \\
& 800 & 202268 & 2269 & 38 & 202230 & 2231 & 0.0\% & 1.7\% & 0.0\% \\
& 1000 & 178051 & 2269 & 19 & 178032 & 2250 & 0.0\% & 0.8\% & 0.0\% \\
& 1200 & 138551 & 2269 & 12 & 138539 & 2257 & 0.0\% & 0.5\% & 0.0\% \\
& 1400 & 97825 & 2269 & 4 & 97821 & 2265 & 0.0\% & 0.2\% & 0.0\% \\
& 1600 & 64925 & 2269 & 3 & 64922 & 2266 & 0.0\% & 0.1\% & 0.0\% \\
& 1800 & 37745 & 2269 & 2 & 37743 & 2267 & 0.0\% & 0.1\% & 0.0\% \\
& 2000 & 21659 & 2269 & 3 & 21656 & 2266 & 0.0\% & 0.1\% & 0.0\% \\
\hline
\multirow{10}{*}{PI SM} & 200 & 99384 & 2021 & 305 & 99079 & 1716 & 0.3\% & 15.1\% & 0.6\% \\
& 400 & 156766 & 2021 & 165 & 156601 & 1856 & 0.1\% & 8.2\% & 0.2\% \\
& 600 & 185712 & 2021 & 120 & 185592 & 1901 & 0.1\% & 5.9\% & 0.1\% \\
& 800 & 189014 & 2021 & 72 & 188942 & 1949 & 0.0\% & 3.6\% & 0.1\% \\
& 1000 & 167363 & 2021 & 51 & 167312 & 1970 & 0.0\% & 2.5\% & 0.1\% \\
& 1200 & 130794 & 2021 & 38 & 130756 & 1983 & 0.0\% & 1.9\% & 0.1\% \\
& 1400 & 93001 & 2021 & 14 & 92987 & 2007 & 0.0\% & 0.7\% & 0.0\% \\
& 1600 & 61976 & 2021 & 7 & 61969 & 2014 & 0.0\% & 0.3\% & 0.0\% \\
& 1800 & 35988 & 2021 & 5 & 35983 & 2016 & 0.0\% & 0.2\% & 0.0\% \\
& 2000 & 20502 & 2021 & 5 & 20497 & 2016 & 0.0\% & 0.2\% & 0.0\% \\
\hline
\multirow{10}{*}{Exact TM} & 200 & 206362 & 11428 & 4482 & 201880 & 6946 & 2.2\% & 39.2\% & 4.1\% \\
& 400 & 404155 & 11428 & 5093 & 399062 & 6335 & 1.3\% & 44.6\% & 2.5\% \\
& 600 & 596166 & 11428 & 5414 & 590752 & 6014 & 0.9\% & 47.4\% & 1.8\% \\
& 800 & 782134 & 11428 & 5637 & 776497 & 5791 & 0.7\% & 49.3\% & 1.4\% \\
& 1000 & 959318 & 11428 & 5701 & 953617 & 5727 & 0.6\% & 49.9\% & 1.2\% \\
& 1200 & 1104095 & 11428 & 5811 & 1098284 & 5617 & 0.5\% & 50.8\% & 1.0\% \\
& 1400 & 1217029 & 11428 & 5727 & 1211302 & 5701 & 0.5\% & 50.1\% & 0.9\% \\
& 1600 & 1294041 & 11428 & 5943 & 1288098 & 5485 & 0.5\% & 52.0\% & 0.9\% \\
& 1800 & 1343294 & 11428 & 5736 & 1337558 & 5692 & 0.4\% & 50.2\% & 0.8\% \\
& 2000 & 1368335 & 11428 & 5417 & 1362918 & 6011 & 0.4\% & 47.4\% & 0.8\% \\
\hline
\multirow{10}{*}{PI TM} & 200 & 150423 & 9367 & 5104 & 145319 & 4263 & 3.4\% & 54.5\% & 6.4\% \\
& 400 & 255097 & 9367 & 6039 & 249058 & 3328 & 2.4\% & 64.5\% & 4.6\% \\
& 600 & 336224 & 9367 & 6659 & 329565 & 2708 & 2.0\% & 71.1\% & 3.9\% \\
& 800 & 400263 & 9367 & 7100 & 393163 & 2267 & 1.8\% & 75.8\% & 3.5\% \\
& 1000 & 452497 & 9367 & 7337 & 445160 & 2030 & 1.6\% & 78.3\% & 3.2\% \\
& 1200 & 490632 & 9367 & 7612 & 483020 & 1755 & 1.6\% & 81.3\% & 3.0\% \\
& 1400 & 515576 & 9367 & 7784 & 507792 & 1583 & 1.5\% & 83.1\% & 3.0\% \\
& 1600 & 535308 & 9367 & 7933 & 527375 & 1434 & 1.5\% & 84.7\% & 2.9\% \\
& 1800 & 547300 & 9367 & 8051 & 539249 & 1316 & 1.5\% & 86.0\% & 2.9\% \\
& 2000 & 552280 & 9367 & 8042 & 544238 & 1325 & 1.5\% & 85.9\% & 2.9\% \\
\bottomrule
\end{tabular}
\end{minipage}
\end{minipage}
\end{adjustbox}
\end{table*}

\begin{table*}[!htbp]
\centering
\hspace*{-7cm}
\begin{adjustbox}{angle=270,center}
\begin{minipage}{1.0\textheight}  % Use textheight since we're rotating
\centering
\caption{Plausibility results of PLM-ICD across all thresholds.}
\label{tab:plausibility_plmicd}

\begin{minipage}[t][0.8\textheight][t]{0.48\textwidth}  % Fixed height for alignment
\centering
\subcaption{Results using the Top‑p selection strategy.}
\scriptsize
\begin{tabular}{lrrrrrrrrr}
\toprule
Metric & Threshold & \#Prd & \#Act & TP & FP & FN & Pr & Rc & F1 \\
\midrule
\multirow{9}{*}{Exact SM} & 10\% & 65639 & 2269 & 172 & 65467 & 2097 & 0.3\% & 7.6\% & 0.5\% \\
& 20\% & 99489 & 2269 & 112 & 99377 & 2157 & 0.1\% & 4.9\% & 0.2\% \\
& 30\% & 117733 & 2269 & 68 & 117665 & 2201 & 0.1\% & 3.0\% & 0.1\% \\
& 40\% & 123897 & 2269 & 48 & 123849 & 2221 & 0.0\% & 2.1\% & 0.1\% \\
& 50\% & 123055 & 2269 & 38 & 123017 & 2231 & 0.0\% & 1.7\% & 0.1\% \\
& 60\% & 112511 & 2269 & 23 & 112488 & 2246 & 0.0\% & 1.0\% & 0.0\% \\
& 70\% & 96512 & 2269 & 14 & 96498 & 2255 & 0.0\% & 0.6\% & 0.0\% \\
& 80\% & 72706 & 2269 & 8 & 72698 & 2261 & 0.0\% & 0.4\% & 0.0\% \\
& 90\% & 44799 & 2269 & 2 & 44797 & 2267 & 0.0\% & 0.1\% & 0.0\% \\
\hline
\multirow{9}{*}{PI SM} & 10\% & 60851 & 2041 & 228 & 60623 & 1813 & 0.4\% & 11.2\% & 0.7\% \\
& 20\% & 91601 & 2041 & 173 & 91428 & 1868 & 0.2\% & 8.5\% & 0.4\% \\
& 30\% & 108523 & 2041 & 125 & 108398 & 1916 & 0.1\% & 6.1\% & 0.2\% \\
& 40\% & 114615 & 2041 & 109 & 114506 & 1932 & 0.1\% & 5.3\% & 0.2\% \\
& 50\% & 113966 & 2041 & 89 & 113877 & 1952 & 0.1\% & 4.4\% & 0.2\% \\
& 60\% & 104917 & 2041 & 63 & 104854 & 1978 & 0.1\% & 3.1\% & 0.1\% \\
& 70\% & 90569 & 2041 & 46 & 90523 & 1995 & 0.1\% & 2.3\% & 0.1\% \\
& 80\% & 68893 & 2041 & 29 & 68864 & 2012 & 0.0\% & 1.4\% & 0.1\% \\
& 90\% & 42952 & 2041 & 13 & 42939 & 2028 & 0.0\% & 0.6\% & 0.1\% \\
\hline
\multirow{9}{*}{Exact TM} & 10\% & 173633 & 12517 & 3975 & 169658 & 8542 & 2.3\% & 31.8\% & 4.3\% \\
& 20\% & 332771 & 12517 & 4194 & 328577 & 8323 & 1.3\% & 33.5\% & 2.4\% \\
& 30\% & 486349 & 12517 & 4251 & 482098 & 8266 & 0.9\% & 34.0\% & 1.7\% \\
& 40\% & 641050 & 12517 & 4256 & 636794 & 8261 & 0.7\% & 34.0\% & 1.3\% \\
& 50\% & 795367 & 12517 & 4168 & 791199 & 8349 & 0.5\% & 33.3\% & 1.0\% \\
& 60\% & 949630 & 12517 & 4015 & 945615 & 8502 & 0.4\% & 32.1\% & 0.8\% \\
& 70\% & 1100798 & 12517 & 4189 & 1096609 & 8328 & 0.4\% & 33.5\% & 0.8\% \\
& 80\% & 1246480 & 12517 & 4106 & 1242374 & 8411 & 0.3\% & 32.8\% & 0.7\% \\
& 90\% & 1381492 & 12517 & 4208 & 1377284 & 8309 & 0.3\% & 33.6\% & 0.6\% \\
\hline
\multirow{9}{*}{PI TM} & 10\% & 120777 & 10269 & 5794 & 114983 & 4475 & 4.8\% & 56.4\% & 8.8\% \\
& 20\% & 203259 & 10269 & 6548 & 196711 & 3721 & 3.2\% & 63.8\% & 6.1\% \\
& 30\% & 270926 & 10269 & 7060 & 263866 & 3209 & 2.6\% & 68.8\% & 5.0\% \\
& 40\% & 332829 & 10269 & 7461 & 325368 & 2808 & 2.2\% & 72.7\% & 4.3\% \\
& 50\% & 388647 & 10269 & 7665 & 380982 & 2604 & 2.0\% & 74.6\% & 3.8\% \\
& 60\% & 439645 & 10269 & 7835 & 431810 & 2434 & 1.8\% & 76.3\% & 3.5\% \\
& 70\% & 485691 & 10269 & 8082 & 477609 & 2187 & 1.7\% & 78.7\% & 3.3\% \\
& 80\% & 526864 & 10269 & 8145 & 518719 & 2124 & 1.5\% & 79.3\% & 3.0\% \\
& 90\% & 560856 & 10269 & 8233 & 552623 & 2036 & 1.5\% & 80.2\% & 2.9\% \\
\bottomrule
\end{tabular}
\end{minipage}
\hfill
\begin{minipage}[t][0.8\textheight][t]{0.48\textwidth}  % Same height for alignment
\centering
\subcaption{Results using the Top‑N selection strategy.}
\scriptsize
\begin{tabular}{lrrrrrrrrr}
\toprule
Metric & Threshold & \#Prd & \#Act & TP & FP & FN & Pr & Rc & F1 \\
\midrule
\multirow{10}{*}{Exact SM} & 200 & 67058 & 2269 & 170 & 66888 & 2099 & 0.3\% & 7.5\% & 0.5\% \\
& 400 & 99646 & 2269 & 98 & 99548 & 2171 & 0.1\% & 4.3\% & 0.2\% \\
& 600 & 115216 & 2269 & 68 & 115148 & 2201 & 0.1\% & 3.0\% & 0.1\% \\
& 800 & 118029 & 2269 & 45 & 117984 & 2224 & 0.0\% & 2.0\% & 0.1\% \\
& 1000 & 112128 & 2269 & 31 & 112097 & 2238 & 0.0\% & 1.4\% & 0.1\% \\
& 1200 & 99278 & 2269 & 18 & 99260 & 2251 & 0.0\% & 0.8\% & 0.0\% \\
& 1400 & 81722 & 2269 & 13 & 81709 & 2256 & 0.0\% & 0.6\% & 0.0\% \\
& 1600 & 64482 & 2269 & 5 & 64477 & 2264 & 0.0\% & 0.2\% & 0.0\% \\
& 1800 & 43416 & 2269 & 4 & 43412 & 2265 & 0.0\% & 0.2\% & 0.0\% \\
& 2000 & 30369 & 2269 & 1 & 30368 & 2268 & 0.0\% & 0.0\% & 0.0\% \\
\hline
\multirow{10}{*}{PI SM} & 200 & 62197 & 2041 & 225 & 61972 & 1816 & 0.4\% & 11.0\% & 0.7\% \\
& 400 & 92102 & 2041 & 161 & 91941 & 1880 & 0.2\% & 7.9\% & 0.3\% \\
& 600 & 106563 & 2041 & 125 & 106438 & 1916 & 0.1\% & 6.1\% & 0.2\% \\
& 800 & 109437 & 2041 & 93 & 109344 & 1948 & 0.1\% & 4.6\% & 0.2\% \\
& 1000 & 104052 & 2041 & 79 & 103973 & 1962 & 0.1\% & 3.9\% & 0.1\% \\
& 1200 & 92471 & 2041 & 55 & 92416 & 1986 & 0.1\% & 2.7\% & 0.1\% \\
& 1400 & 76305 & 2041 & 40 & 76265 & 2001 & 0.1\% & 2.0\% & 0.1\% \\
& 1600 & 60330 & 2041 & 25 & 60305 & 2016 & 0.0\% & 1.2\% & 0.1\% \\
& 1800 & 40620 & 2041 & 22 & 40598 & 2019 & 0.1\% & 1.1\% & 0.1\% \\
& 2000 & 28513 & 2041 & 14 & 28499 & 2027 & 0.0\% & 0.7\% & 0.1\% \\
\hline
\multirow{10}{*}{Exact TM} & 200 & 182545 & 12517 & 3972 & 178573 & 8545 & 2.2\% & 31.7\% & 4.1\% \\
& 400 & 348662 & 12517 & 4175 & 344487 & 8342 & 1.2\% & 33.4\% & 2.3\% \\
& 600 & 509830 & 12517 & 4253 & 505577 & 8264 & 0.8\% & 34.0\% & 1.6\% \\
& 800 & 672002 & 12517 & 4313 & 667689 & 8204 & 0.6\% & 34.5\% & 1.3\% \\
& 1000 & 834402 & 12517 & 4207 & 830195 & 8310 & 0.5\% & 33.6\% & 1.0\% \\
& 1200 & 989036 & 12517 & 4262 & 984774 & 8255 & 0.4\% & 34.0\% & 0.9\% \\
& 1400 & 1121881 & 12517 & 4068 & 1117813 & 8449 & 0.4\% & 32.5\% & 0.7\% \\
& 1600 & 1218150 & 12517 & 4272 & 1213878 & 8245 & 0.4\% & 34.1\% & 0.7\% \\
& 1800 & 1285721 & 12517 & 4576 & 1281145 & 7941 & 0.4\% & 36.6\% & 0.7\% \\
& 2000 & 1335649 & 12517 & 4828 & 1330821 & 7689 & 0.4\% & 38.6\% & 0.7\% \\
\hline
\multirow{10}{*}{PI TM} & 200 & 126983 & 10269 & 5801 & 121182 & 4468 & 4.6\% & 56.5\% & 8.5\% \\
& 400 & 212631 & 10269 & 6635 & 205996 & 3634 & 3.1\% & 64.6\% & 6.0\% \\
& 600 & 283642 & 10269 & 7157 & 276485 & 3112 & 2.5\% & 69.7\% & 4.9\% \\
& 800 & 346990 & 10269 & 7527 & 339463 & 2742 & 2.2\% & 73.3\% & 4.2\% \\
& 1000 & 404689 & 10269 & 7725 & 396964 & 2544 & 1.9\% & 75.2\% & 3.7\% \\
& 1200 & 453426 & 10269 & 7908 & 445518 & 2361 & 1.7\% & 77.0\% & 3.4\% \\
& 1400 & 491786 & 10269 & 7968 & 483818 & 2301 & 1.6\% & 77.6\% & 3.2\% \\
& 1600 & 518079 & 10269 & 8153 & 509926 & 2116 & 1.6\% & 79.4\% & 3.1\% \\
& 1800 & 533761 & 10269 & 8111 & 525650 & 2158 & 1.5\% & 79.0\% & 3.0\% \\
& 2000 & 547686 & 10269 & 8052 & 539634 & 2217 & 1.5\% & 78.4\% & 2.9\% \\
\bottomrule
\end{tabular}
\end{minipage}
\end{minipage}
\end{adjustbox}
\end{table*}

\end{document}